\documentclass{article}

\usepackage[final,nonatbib]{neurips_2021}
\usepackage[utf8]{inputenc} 
\usepackage[T1]{fontenc}    
\usepackage{hyperref}       
\usepackage{url}            
\usepackage{amsfonts}       
\usepackage{nicefrac}       
\usepackage{microtype}      
\usepackage[caption=false,font=footnotesize]{subfig}
\usepackage{url}
\usepackage{graphicx}
\usepackage{mathtools}
\usepackage{epstopdf}
\usepackage{float}
\usepackage{sidecap}
\usepackage{newpxmath}
\usepackage{newpxtext,newpxmath}
\usepackage{amsmath,amssymb,bm}
\usepackage{footnote}
\usepackage{enumerate}
\usepackage{physics}
\usepackage{enumitem}
\usepackage[english]{babel}
\usepackage{comment}
\usepackage{setspace}
\usepackage[ruled,vlined]{algorithm2e}

\usepackage{forloop}
\usepackage{chngpage}
\usepackage{nicefrac}       
\usepackage{microtype}      
\usepackage{xcolor, colortbl}
\definecolor{darkgreen}{rgb}{0, 0.5, 0}
\definecolor{red}{rgb}{1, 0, 0}
\definecolor{purple}{rgb}{0.5, 0, 0.5}
\usepackage{booktabs, multirow}
\usepackage{comment}
\usepackage[toc,page]{appendix}

\newcommand\ie{\textit{i.e.}}
\newcommand\eg{\textit{e.g.}}
\newcommand\st{\textit{s.t.}}
\newcommand\vs{\textit{vs}}
\newcommand\wrt{\textit{w.r.t.}}
\newcommand\aka{\textit{a.k.a.}}
\newcommand\etc{\textit{etc.}}

\newcommand\doubleR{\mathbb{R}}

\newcommand\scriptM{\mathcal{M}}
\newcommand\scriptQ{\mathcal{Q}}
\newcommand\scriptS{\mathcal{S}}
\newcommand\scriptO{\mathcal{O}}
\newcommand\scriptA{\mathcal{A}}

\newcommand\scriptL{\mathcal{L}}

\usepackage{titlecaps}
\newcommand\compression{selection}
\newcommand\decompression{integration}

\newcommand\compress{select}
\newcommand\decompress{integrate}

\newcommand\compressor{selector}
\newcommand\decompressor{integrator}

\newcommand{\llIf}[2]{{\let\par\relax\lIf{#1}{#2}}}
\newcommand{\llElse}[1]{{\let\par\relax\lElse{#1}}}

\title{A Consciousness-Inspired Planning Agent for Model-Based Reinforcement Learning}

\author{
Mingde Zhao$^{1,4,*}$, Zhen Liu$^{2,4,*}$, Sitao Luan$^{1,4,*}$, Shuyuan Zhang$^{1,4,*}$\\
\textbf{
Doina Precup$^{1,3,4,5\dagger}$, Yoshua Bengio$^{2,4,5\dagger}$
}\\
$^1$McGill University; $^2$Universit\'{e} de Montr\'eal; $^3$DeepMind; $^4$Mila; $^5$ CIFAR AI Chair\\
\textbf{$^{*}$}: Equal Contribution, \textbf{$^{\dagger}$}: Equal Supervision
}

\begin{document}

\maketitle

\begin{abstract}
We present an end-to-end, model-based deep reinforcement learning agent which dynamically attends to relevant parts of its state during planning. The agent uses a bottleneck mechanism over a set-based representation to force the number of entities to which the agent attends at each planning step to be small. In experiments, we investigate the bottleneck mechanism with several sets of customized environments featuring different challenges. We consistently observe that the design allows the planning agents to generalize their learned task-solving abilities in compatible unseen environments by attending to the relevant objects, leading to better out-of-distribution generalization performance.
\end{abstract}

\section{Introduction}

Whether when planning our paths home from the office or from a hotel to an airport in an unfamiliar city, we typically focus on a small subset of relevant variables, \eg{} the change in position or the presence of traffic. An interesting hypothesis of how this path planning skill generalizes across scenarios is that it is due to computation associated with the conscious processing of information \cite{baars1993cognitive,baars2002conscious,dehane2017consciousness}. Conscious attention focuses on a few necessary environment elements, with the help of an internal abstract representation of the world~\cite{vangulick2004consciousness,dehane2017consciousness}. This pattern, also known as consciousness in the first sense (C1)~\cite{dehane2017consciousness}, has been theorized to enable humans' exceptional adaptability and learning efficiency \cite{baars1993cognitive,baars2002conscious,dehane2017consciousness,vangulick2004consciousness,bengio2017consciousness,goyal2020inductive}. A central characterization of conscious processing is that it involves a \textit{bottleneck}, which forces one to handle dependencies between very few environmental characteristics at a time \cite{dehane2017consciousness,bengio2017consciousness,goyal2020inductive}. Though focusing on a subset of the available information may seem limiting, it facilitates Out-Of-Distribution (OOD) and systematic generalization to other situations where the ignored variables are different and yet still irrelevant~\cite{bengio2017consciousness,goyal2020inductive}.

In this paper, we encode some of these ideas into reinforcement learning agents. Reinforcement learning (RL) is an approach for learning behaviors from agent-environment interactions \cite{sutton2018reinforcement}. However, most of the big successes of RL have been obtained by deep, model-free agents~\cite{mnih2015human,silver2016mastering,silver2017mastering}. While Model-Based RL (MBRL) has generated significant research due to the potentials of using an extra model \cite{moerland2020model}, its empirical performance has typically lagged behind, with some recent notable exceptions \cite{schrittwieser2020mastering,kaiser2019model,hafner2020discrete}.

Our proposal is to take inspiration from human consciousness to build an architecture which learns a useful state space and in which attention can be focused on a small set of variables at any time, where the aspect of ``partial planning''\footnote{Partial planning is interpreted in different ways. For example, concurrent work \cite{khetarpal2021temporally} focuses on modelling ``affordable'' temporally extended actions, \st{} an ``intent'' could be achieved more efficiently.} is enabled by modern deep RL techniques \cite{talvitie2008simple,khetarpal2021temporally}. Specifically, we propose an end-to-end latent-space MBRL agent which does not require reconstructing the observations, as in most existing works, and uses Model Predictive Control (MPC) framework for decision-time planning~\cite{rao2009survey,richards2005robust}. From an observation, the agent encodes a set of objects as a state, with a selective attention bottleneck mechanism to plan over selected subsets of the state (Sec.~\ref{sec:CP}). Our experiments show that the inductive biases improve a specific form of OOD generalization, where consistent dynamics are preserved across seemingly different environment settings (Sec.~\ref{sec:experiments}).

\section{Background \& Context}
\label{sec:setting}

We consider an agent interacting with its environment at discrete timesteps. At time $t$, the agent receives observation $o_t$ and takes action $a_t$, receiving a reward $r_{t+1}$ and new observation $o_{t+1}$. The interaction is episodic. The agent is also building a latent-space transition model, $\scriptM$, which can be used to sample 
a next state, $\hat{s}_{t+1}$, a reward $\hat{r}_{t+1}$ and a binary signal $\hat{\omega}_{t+1}$ which indicates if the model predicts termination after the transition. We will now compare and contrast our approach with some existing methods from the MBRL literature, explaining the rationale for our design choices. 

{\bf Observation Level Planning and Reconstruction \vs{} Latent Space Planning}\\
Many MBRL methods plan in the observation space or rely on reconstruction-based losses to obtain state representations \cite{kaiser2019model,schrittwieser2020mastering,hafner2020discrete,wang2018leap}. Appropriate as these methods may be for some robotic tasks with few sensory inputs, \eg{} continuous control with joint states, they are arguably difficult with high-dimensional inputs like images, since they may focus on predictable yet useless aspects of the raw observations \cite{moerland2020model}. Besides suffering from the need to reconstruct noise or irrelevant parts of the signal, it is not clear if representations built by a reconstruction loss (\eg{} $L_2$ in the observation space) are effective for an MBRL agent to plan or predict the desired signals \cite{silver2016predictron,hafner2020discrete,hamrick2020role}, \eg{} values (in the RL sense), rewards, \etc{}. In this work, we use an approach similar to those in \cite{silver2016predictron,schrittwieser2020mastering,hafner2020discrete}, building a latent space representation that is jointly shaped by all the relevant RL signals (to serve value estimation and planning) without using reconstruction. 

{\bf Staged Training \vs{} End-to-End Training}\\
Some MBRL agents based on a world model~\cite{ha2018world,kaiser2019model,moerland2020model} use two explicit stages of training: (1) an inner representation of the world is trained using exploration (usually with random trajectories); (2) the representation is fixed and used for planning and MBRL. Despite the advantages of being more stable and easier to train, this procedure relies on having an environment where the initial exploration provides transitions that are sufficiently similar to those observed under improved policies, which is not the case in many environments.
Furthermore, the learned representation may not be effective for value estimation, if these transitions do not contain reward information that can be used to update the input-to-representation encoder. End-to-end MBRL agents, \eg{} \cite{silver2016predictron,schrittwieser2020mastering}, are able to learn the representation online, simultaneously with the value function, hence adapting better to non-stationarity in the transition distribution and  rewards.

{\bf Type of planning}\\
MBRL agents can use the model in different ways.  Dyna~\cite{sutton1991dyna} learns a model to generate ``imaginary'' transitions, which contribute to the training of the value estimator \cite{sutton1991dyna}, in addition to the real observations, thus boosting sample efficiency. However, if the model is inaccurate, the transitions it generates may be ``delusional", which may alter the value estimator and negatively impact performance. Moreover, Dyna is typically used to generate extra transitions from the states visited in a trajectory, and updates the model based on the observed transitions as well. This means Dyna is focused on the data distribution encountered by the agent and may have trouble generalizing OOD. In contrast, simulation-based model-predictive control (MPC) and its variants \cite{rao2009survey,richards2005robust,hamrick2020role} only update the value estimator based on real data, using the model simply to perform lookahead at decision-time. Hence, model inaccuracies impact less, with more favorable OOD generalization capabilities. Hence, MPC is adopted in our approach.

\begin{figure*}[htbp]
\centering
\captionsetup{justification = centering}
\includegraphics[width=1.0\textwidth]{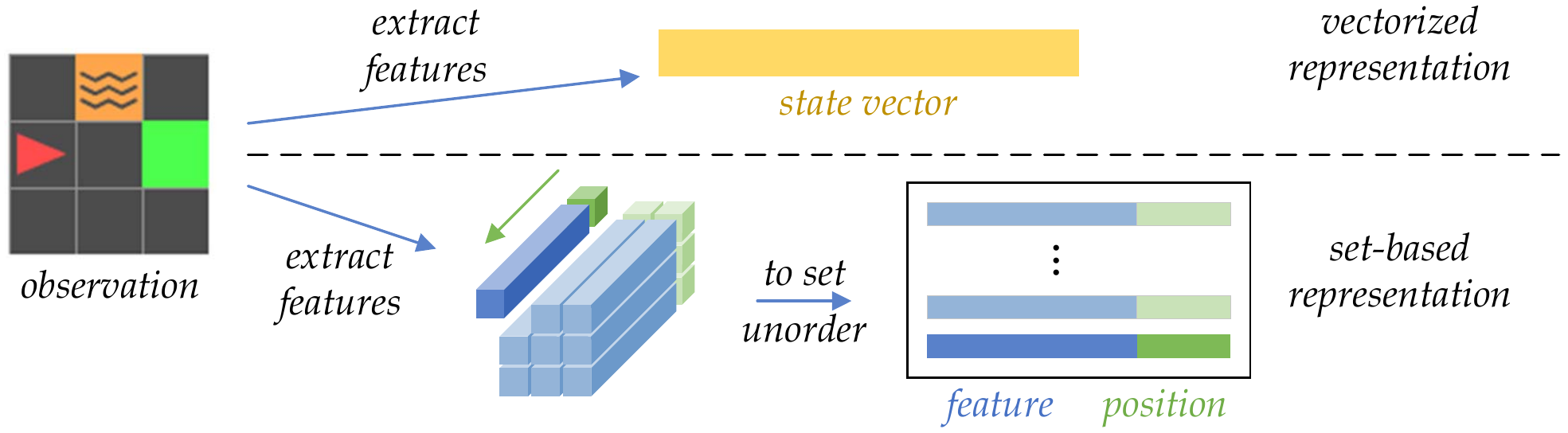}
\caption{\small \textbf{Set-based state encoder} compared to classical vectorized state encoders: the feature map extracted by some feature extractor, \eg{} a CNN, is ``chopped'' into feature vectors and concatenated with positional information. All of the resulting concatenations are treated as \textit{objects} in a set, capturing the features of observed entities. The permutation-invariance of set computations forces the learner to be robust to small changes in the set (\eg{} one of the elements being different or missing).}
\label{fig:encoder}
\end{figure*}

{\bf Vectorized \vs{} Set Representations for RL}\\
Most Deep Reinforcement Learning (DRL) work focus on learning vectorized state representations, where the agents' observation is transformed into a feature vector of fixed dimensionality~\cite{mnih2015human,hessel2017rainbow}. Instead, set-based encoders, \aka{} object-oriented architectures, are designed to extract a set of unordered vectors from which to predict the desired signals via permutation-invariant computations \cite{zaheer2017sets}, as illustrated in Fig.~\ref{fig:encoder}. Recent works in RL have shown the promise of set-based representations in capturing environmental states, in terms of generalization, as well as their similarities to human perception \cite{davidson2020investigating,wang2018nervenet,mu2020refactoring,vinyals2019grandmaster,lowe2020contrasting}. Additionally in this work, we utilize the compositionality of set representations to enable the discovery of sparse interactions among objects, \ie{} underlying dynamics, as well as to facilitate the bottleneck mechanism, analogous to C1 selection. The set-based representation coupled with the bottleneck provides an inductive bias consistent with selecting only the relevant aspects of a situation on-the-fly through an attention mechanism. The small size of the working memory bottleneck also enforces sparsity of the dependencies~\cite{bengio2017consciousness,goyal2020inductive} captured by the learned dynamics model: each transition can only relate a few objects together, no more than the size of the bottleneck.

\section{MBRL with Set Representations}
\label{sec:UP}

We present an end-to-end baseline MBRL agent that uses a set-based representation and carries out latent space planning, but \textbf{without} a consciousness-inspired small bottleneck. This agent serves as a baseline to investigate the OOD generalization capabilities brought by the bottleneck, which is to be introduced later in Sec.~\ref{sec:CP}.

The mapping from observations to values is a combination of an \textit{encoder} and a \textit{value estimator}. The encoder maps an observation vector to a set of objects, which constitutes the latent state. The value estimator is a permutation-invariant set-to-vector architecture that maps the latent state to a value estimate. Note that the same state set is used for all the agents' predictions, including future states, rewards \etc{}, as we will discuss later.

\textbf{Encoder.} For image-based observations, we use the features at each position of the CNN output feature map to characterize the feature of an object, similar to \cite{carion2020end}, as shown in Figure \ref{fig:encoder}. To recover positional information lost during the process, we concatenate each object feature vector with a positional embedding to form a complete object embedding. Such approach is different from the common practice of mixing positional information by addition \cite{vaswani2017attention}. This is for the compatibility with our dynamics model training procedure, discussed below. 

\begin{figure*}[htbp]
\centering
\captionsetup{justification = centering}
\includegraphics[width=1.0\textwidth]{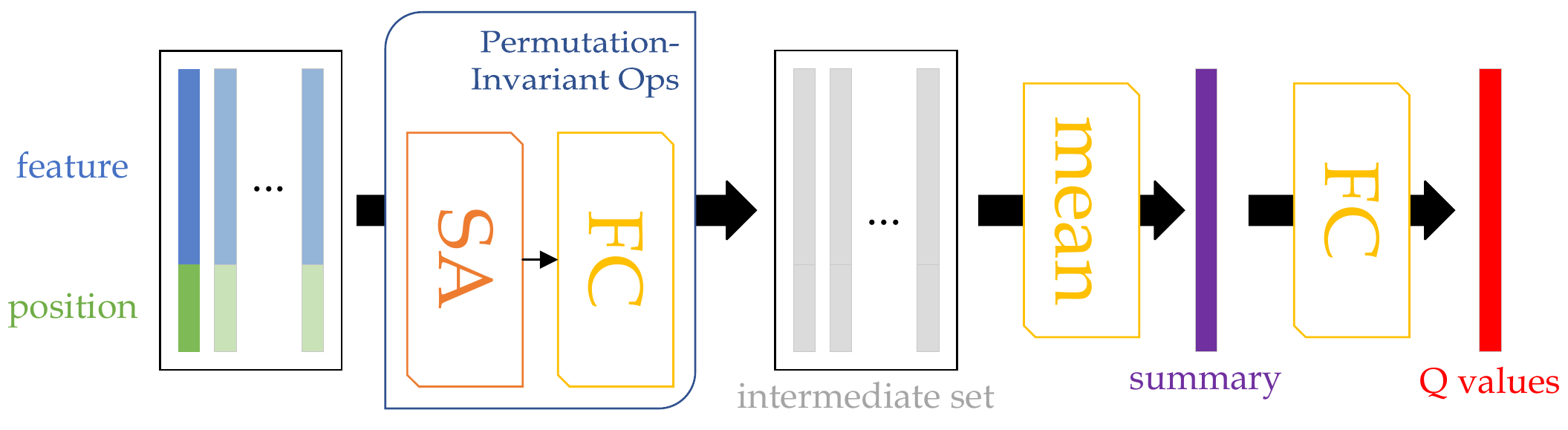}
\caption{\small \textbf{Value estimator} $Q$ and \textbf{a generic set-to-vector architecture}: we modify the design of DeepSets \cite{zaheer2017sets} by replacing the MLP before pooling with transformer layers (multi-head Self-Attention (SA) + object-wise Fully Connected (FC)) \cite{vaswani2017attention}. We found this change to be helpful for performance. After applying the transformer layers, the intermediate set (colored gray) entangles features and positions. Please check the Appendix for more details on the self-attention operations involved.}
\label{fig:value_estimator}
\end{figure*}

\textbf{(State-Action) Value Estimator} takes the form $Q: \scriptS \to \doubleR^{|\scriptA|}$, where $\scriptS$ is the learned state space by the set-based encoder (hoping to capture the real underlying state space of the MDP) and $\scriptA$ is a discrete action set. We use an improved architecture upon DeepSets \cite{zaheer2017sets}, depicted in Figure \ref{fig:value_estimator}. The architecture performs reasoning on a set of encoded objects, resembling pervasive usage in natural language processing, where the objects are typically word tokens \cite{porada2021modeling}.

\textbf{Transition Model.}
The transition model maps from $s_t, a_t$ to $\hat{s}_{t+1}$, $\hat{r}_t$ and $\hat{\omega}_{t+1}$. We separate this into: 1) the \textbf{dynamics model}, in charge of simulating how the state would change with the input of $a_t$ and 2) the \textbf{reward-termination estimator} which maps $s_t, a_t$ to $\hat{r}_t$ and $\hat{\omega}_{t+1}$.

\begin{figure*}[htbp]
\centering
\captionsetup{justification = centering}
\includegraphics[width=1.0\textwidth]{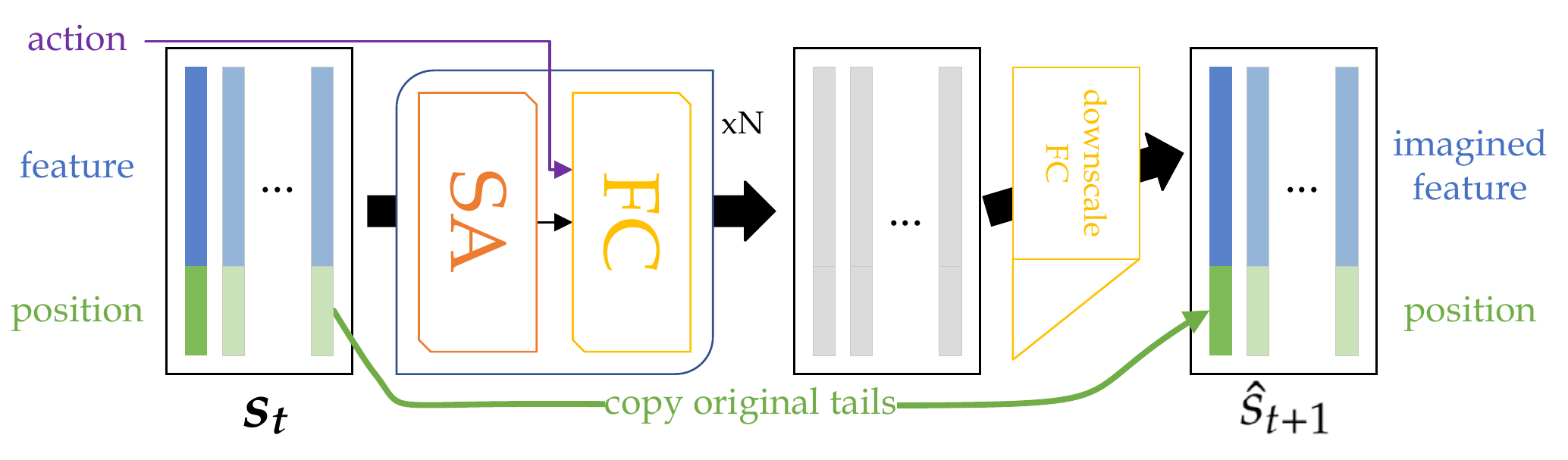}
\caption{\small \textbf{Dynamics model}: for FC sub-layers of the transformer layers, we inject an action embedding \st{} the transformer computations are now action conditioned. After getting the intermediate set, we downscale each of the objects, leaving the positions untouched and directly copied from the input $s_t$. FC downscale is a linear transformation which downscales the dimensionality of the intermediate objects to that of the features part of objects (before the layernorm). In this way, after concatenating the positional tails the objects have consistent dimensionality. Intuitively, each object slot recovers its positional tail at the output. Though the objects in the sets (input-intermediate-output) are aligned, within each set they are still unordered, \ie{} permutation-invariant.}
\label{fig:dynamics_UP}
\end{figure*}

While designing reward-termination estimator is straightforward (a two-headed augmented architecture similar to the value estimator), the dynamics model requires regression on \textit{unordered} sets of objects (set-to-set). A common approach is to use matching methods, \eg{} Chamfer matching or Hausdorff distance, However, they are computationally demanding and subject to local optima \cite{barrow1977parametric,borgefors1988hierarchical,kosiorek2020conditional}. Targeting this, our feature-position separated set encoding not only makes the permutation-invariant computations position-aware, but also allows simple end-to-end training over the dynamics. By forcing the positional tails to be \textit{immutable} during the computational pass, we can use them to solve the matching trivially: objects ``labeled'' with the same positional tail in the prediction $\hat{s}_{t+1}$ (output of the dynamics model) and the training sample $s_{t+1}$ (state obtained from the next observation) are aligned, forming pairs of objects with changes \textit{only} in the feature, as shown in Figure \ref{fig:dynamics_UP}. 

\textbf{Tree Search MPC.} The agent employs a tree-search based behavior policy (with $\epsilon$-greedy exploration). 
During planning, each tree search call maintains a priority queue of branches to simulate with the model. When a designated budget (\eg{} number of steps of simulation) is spent, the agent greedily picks the immediate action that leads to the most promising path. We present the pseudocode of the Q-value based prioritized tree-search MPC in Appendix.


Equivalence could be drawn from this planning approach to Monte-Carlo Tree Search (MCTS) \cite{silver2016mastering,silver2017mastering}. While this method is far more simplistic and require fewer simulations for each planning call (see example in Appendix).

\textbf{Training.}
The proposed agent is trained from sampled transitions with the following losses:

\begin{itemize}[leftmargin=*]
\item 
Temporal Difference (TD) $\scriptL_{\text{TD}}$: regresses the current value estimate to the update target, \eg{} calculated according to DQN or Double DQN (DDQN)~\cite{mnih2015human,hasselt2015double}. In experiments, a distributional output is used for both value and reward estimation, making this loss a KL-divergence \cite{bellemare2017distributional}.
\item
Dynamics Consistency $\scriptL_{\text{dyn}}$: A $L_2$ penalty established between the aligned $\hat{s}_{t+1}$ and $s_{t+1}$, where $\hat{s}_{t+1}$ is the imagined next (latent) state given $o_t, a_t$ and $s_{t+1}$ is the true next (latent) state encoded from $o_{t+1}$.
\item
Reward Estimation $\scriptL_{r}$: the KL-divergence between the imagined reward $\hat{r}_{t+1}$ predicted by the model and the true reward $r_{t+1}$ of the observed transition.
\item
Termination Estimation $\scriptL_{\omega}$: the binary cross-entropy loss from the imagined termination $\hat{\omega}_{t+1}$ to the ground truth $\omega_{t+1}$, obtained from environment feedback.
\end{itemize}

The resulting total loss for end-to-end training of this set-based MBRL agent is thus\footnote{In our experiments, no re-weighting is used for each term of the total loss. This is possible for the fact that they are in similar magnitudes. In our experimental implementation, no recurrent mechanism is used however the same training procedure is naturally extendable.}:
\[
    \mathcal{L} = \scriptL_{\text{TD}} + \scriptL_{\text{dyn}} + \scriptL_{r} + \scriptL_{\omega}
\]
Jointly shaping the states avoids the representation collapsing to trivial solutions and makes the representation useful for all signal predictions of interest.

\section{Consciousness-Inspired Bottleneck}
\label{sec:CP}

\begin{figure*}[htbp]
\centering
\captionsetup{justification = centering}
\includegraphics[width=0.85\textwidth]{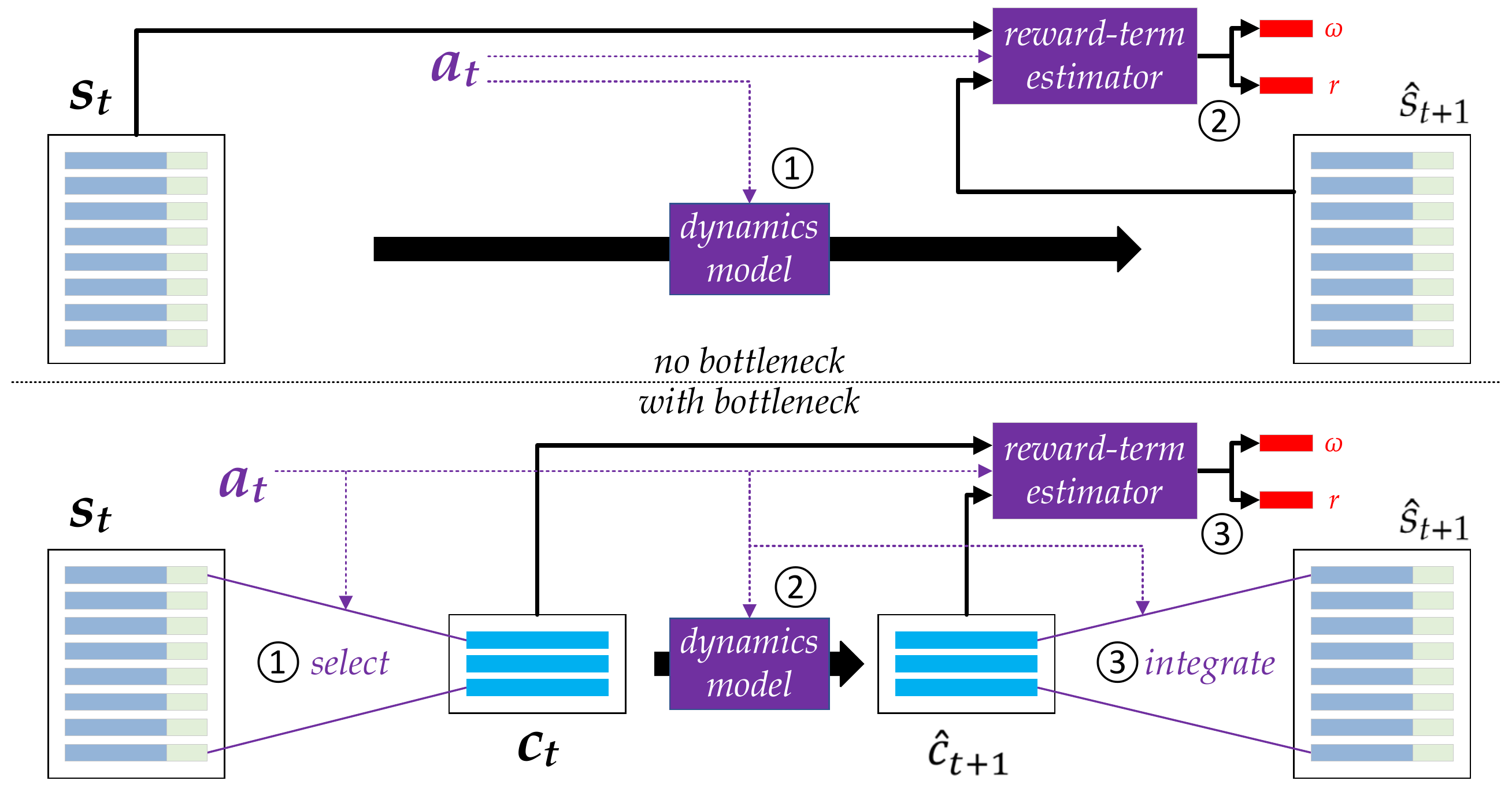}
\caption{\small \textbf{Bottleneck stages} (operations colored in \textcolor{purple}{purple} are conditioned on a chosen action): 1) a bottleneck set $c_t$ is soft-selected from the whole state (object set) $s_t$ through semi-hard multi-head attention; 2) dynamics are applied to the bottleneck set $c_t$ to form $\hat{c}_{t+1}$; 3) the reward and termination signals are predicted from $c_t$, $\hat{c}_{t+1}$ and $a_t$. Then, the changes introduced in $\hat{c}_{t+1}$ are \decompress{}d with $s_t$ to obtain $\hat{s}_{t+1}$, the imagined next state, with the help of attention. Note that the two computational flows in stage $3$ are naturally parallelizable.}
\label{fig:bottleneck}
\end{figure*}

In this section, we introduce an inductive bias which facilitates C1-capable planning. In a nutshell, the planning is expected to focus on the parts of the world that matter for the plan. Simulations and predictions are all expected to be performed on a (small) bottleneck set, which contains all the important transition-related information. As illustrated in Figure \ref{fig:bottleneck}, the model performs 1) \compression{} of the bottleneck set from the full state-set, 2) dynamics simulation on the bottleneck set and 3) \decompression{} of predicted bottleneck set to form the predicted next state.

\textbf{Conditional State \titlecap{\compression{}}} We \compress{} a bottleneck set $c_t$ of $n$ objects from the potentially large state set $s_t$ of $m \gg n$ objects. Then we only model the transition for the selected objects in $c_t$. To make this \compression{}, we use a key-query-value attention mechanism, where the key and the value for each object in $s_t$ are obtained from that object, and the query is a function of some learned dedicated set of vectors and of the action considered (see Appendix for details). Inspired by the work on self-attention for memory access \cite{ke2018sparse}, we use a semi-hard top-$k$ attention mechanism to facilitate the \compression{} of the bottleneck set. That is, after the query, the top-$k$ attention weights are kept, all others are set to $0$, and then the attention weights are renormalized. This semi-hard attention technique limits the influence of the ill-matched objects on the bottleneck set $c_t$ while allowing for a gradient to propagate on the assignment of relative weight to different objects. With purely soft attention, weights for irrelevant objects are never $0$ and learning to disentangle objects may be more difficult.

\textbf{Dynamics / Reward-Termination Prediction on Bottleneck Sets.} We use the same architecture as described in Sec.~\ref{sec:UP}, but taking the bottleneck objects as input rather than the full state set. Details of the architecture are in the Appendix.

\textbf{Change \titlecap{\decompression{}}.} An \decompression{} operation, intuitively the inverse operation of \compression{}, is implemented to `soft paste-back' the changes of the bottleneck state onto the state set $s_t$, yielding the imagined next state set $\hat{s}_{t+1}$. This is also achieved by attention operations, more specifically querying $\hat{c}_{t+1}$ with $s_t$, conditioned on the action $a_t$. Please check the Appendix for more details.

\textbf{Discussion.}
The bottleneck described in this section is a natural complement to the MBRL model with set representations discussed previously. In particular, planning and training are carried out the same way as discussed in Sec.~\ref{sec:UP}.

We expect the Conscious Planning (CP) agent to demonstrate the following advantages:
\begin{itemize}[leftmargin=*]
\item
Higher Quality Representation: the interplay between the set representation and the \compression{} / \decompression{} forces the representation to be more disentangled and more capable of capturing the locally sparse dynamics.
\item
More Effective Generalization: only essential objects for the purpose of planning participate in the transition, thus generalization should be improved both in-distribution and OOD, because the transition does not depend on the parts of the state ignored by the bottleneck.
\item
Lower Computational Complexity: directly employing transformers to simulate the full state dynamics results in a complexity of $\scriptO(|s_t|^2 d)$, where $d$ is the length of the objects, due to the use of Self-Attention (SA), while the bottleneck lowers it to $\scriptO(|s_t||c_t| d)$.
\end{itemize}

\begin{figure}[H]
\centering
\subfloat[In-dist, diff $0.35$]{
\captionsetup{justification = centering}
\includegraphics[width=0.185\textwidth]{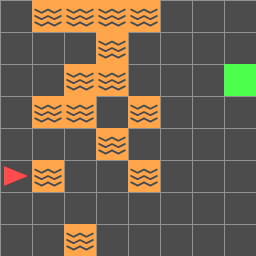}}
\hfill
\subfloat[OOD, diff $0.25$]{
\captionsetup{justification = centering}
\includegraphics[width=0.185\textwidth]{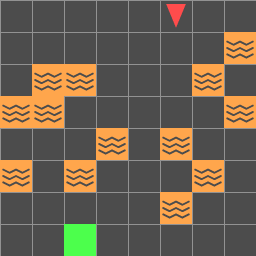}}
\hfill
\subfloat[OOD, diff $0.35$]{
\captionsetup{justification = centering}
\includegraphics[width=0.185\textwidth]{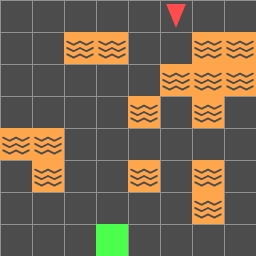}}
\hfill
\subfloat[OOD, diff $0.45$]{
\captionsetup{justification = centering}
\includegraphics[width=0.185\textwidth]{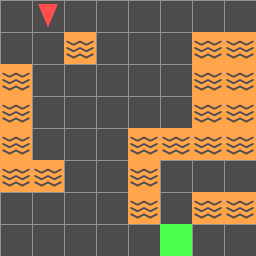}}
\hfill
\subfloat[OOD, diff $0.55$]{
\captionsetup{justification = centering}
\includegraphics[width=0.185\textwidth]{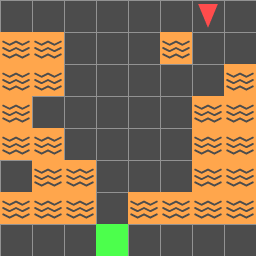}}
\caption{\small \textbf{Non-Static RL Setting, with in-distribution and OOD tasks}: (a) example of training environments (b - e) examples of OOD environments (rotated 90 degrees, changing the distribution of grid elements). For OOD testing, we evaluate different levels of difficulty (b - e). The agent (red triangle) points  in the forward movement direction. The goal is marked in green. For each episode (training or OOD), we randomly generate a new world from a sampling distribution. Note that the training environments and the OOD testing environments have no intersecting observations.}
\label{fig:distshift}
\end{figure}

\section{Experiments}
\label{sec:experiments}

We present our experimental settings and ablation studies of our CP agent against baselines to investigate the OOD generalization capabilities enabled by the C1-inspired bottleneck mechanism. To clarify, the OOD generalization we refer to specifically is \textit{the agents' ability to generalize its learned task skills across seemingly different tasks with common underlying dynamics}. Take the set of experiments in this section for example, we want the agent to be able to generalize its navigation skills in unseen environments.

\subsection{Environment / Task Description}


We use environments based on the MiniGrid-BabyAI framework \cite{chevalierboisvert2018minigrid,chevalier2018babyai,hui2020babyai}, which can be customized for generating OOD generalization tests with varying difficulties. To make sure we assess the agents as clearly as possible, the customized environments feature clear object definitions, with well-understood underlying dynamics based on object interactions. Furthermore, the environments are solvable by Dynamic Programming (DP) and can be easily tuned to generate OOD evaluation tasks. These characteristics are \textbf{crucial} for the experimental insights we are seeking.

In this section, the experiments are carried out on $8 \times 8$ gridworlds\footnote{We provide additional results for world sizes ranging from $6\times6$ to $10 \times 10$ in the Appendix. $8 \times 8$ is chosen as the demonstrative case.}, as shown in Figure \ref{fig:distshift}. The agent (red triangle) needs to navigate (by turning left, right or stepping forward) to the goal while dodging the lava cells along the way\footnote{In the Appendix, we provide additional test settings with different dynamics, which also demonstrates the agents' ability to work well despite cluttering distractions.}. If the agent steps into lava (orange square), the episode terminates immediately with no reward. If the agent successfully reaches the goal (green square), it receives a reward of $+1$ and the episode terminates. For better generalization, the agent needs to understand how to avoid lava in general (and not at specific locations, since their placement changes) and to reach the goal as quickly as possible\footnote{Please check the Appendix for extra sets of tasks with different agent actions and task objectives.}. The environments provide grid-based observations that are ready to be interpreted as set representations: each cell of the observation array is an object, thus resulting in a set of $64$ objects in $s_t$ for each observation.

For the agent to be able to \textit{understand} the environment dynamics instead of \textit{memorizing} specific task layouts, we generate a new environment for each training or evaluation episode. In each training episode, the agent starts at a random position on the leftmost or rightmost edge and the goal is placed randomly somewhere along the opposite edge. In between the two edges, the lava cells are randomly generated according to a \textit{difficulty} parameter which controls the probability of placing a lava cell at each valid position. The difficulty parameter controls partially how seemingly different the OOD evaluation tasks are to the in-distribution training tasks, though we know the underlying dynamics of all these tasks are the same. For training episodes, the difficulty is fixed to $0.35$. We note that most usual RL benchmarks contain fixed environments, where the agent is expected to acquire a specific optimal policy. These environments are ill-suited for our purpose.

For OOD evaluation, the agent is expected to adapt in new tasks with the \textbf{same} underlying dynamics in a $0$-shot fashion, \ie{} with the agent's parameters fixed. The OOD tasks are crafted to include changes both in the support (orientation) and in the distribution (difficulty): the agent is deployed in \textit{transposed} layouts\footnote{The agent starts at the top or bottom edge and the goal is respectively on the bottom or top edge, whereas a training environment has the agent and goal on the left or right edges} with varying levels of difficulty ($\{0.25, \textbf{0.35}, 0.45, 0.55\}$). The differences of in-distribution (training) and OOD (evaluation) environments are illustrated in Figure \ref{fig:distshift}.

\subsection{Agent Setting}
We build all the set-based MBRL agents included in the evaluation on a common model-free baseline: a set-based variant of Double-DQN (DDQN) \cite{hasselt2015double} with prioritized replay and distributional outputs. For more details, please check the Appendix.

We compare the proposed approach, labelled CP in the figures (for Conscious Planning) against the following methods:
\begin{itemize}[leftmargin=*]
\item
\textit{UP} (for Unconscious Planning): the agent proposed in Section \ref{sec:UP}, lacking the bottleneck.
\item
\textit{model-free}: the model-free set-based agent is the basis for the set-based model-based agents. It consists of only the encoder and the value estimator, sharing their architectures with CP and UP.
\item
\textit{Dyna}: the set-based MBRL agent which includes a model-free agent and an observation-level transition model, \ie{} a transition generator. For the model, we use the CP transition model (with the same hyperparameters as the best performing CP agent) on the original environment features without an encoder. We also use the same hyperparameters as in the CP model training. The agent essentially doubles the batch size of the model-free baseline by augmenting training batches with an equal number of generated transitions.
\item
\textit{Dyna*}: A Dyna baseline that uses the true environment model for transition generation. This is expected to demonstrate Dyna's performance limit.
\item
\textit{WM-CP}: A world model CP variant that differs by following a $2$-stage training procedure \cite{ha2018world}. First, the model (together with the encoder) is trained with $10^{6}$ random transitions. After this, the encoder and the model are fixed and RL begins.
\item
\textit{NOSET}: A UP-counterpart with vectorized representations and no bottleneck mechanism.
\end{itemize}

Particularly, for CP and UP agents, we also test the following variants:

\begin{itemize}[leftmargin=*]
\item
\textit{CP-noplan}: A CP agent that trains normally but does not plan in OOD evaluations, \ie{} carrying out model-free behavior. This baseline aims to demonstrate the impact of planning in the training process on the OOD capability of the value estimator.
\item
\textit{UP-noplan}: UP counterpart of CP-noplan.
\end{itemize}

Note that the compared methods share architectures as much as possible to ensure fair comparisons. Details of the compared methods, their design and hyperparameters are provided in the Appendix.

\subsection{Performance Evaluation}

\subsubsection{In-Distribution}
\begin{SCfigure}
\centering
\captionsetup{justification = centering}
\includegraphics[width=0.4\textwidth]{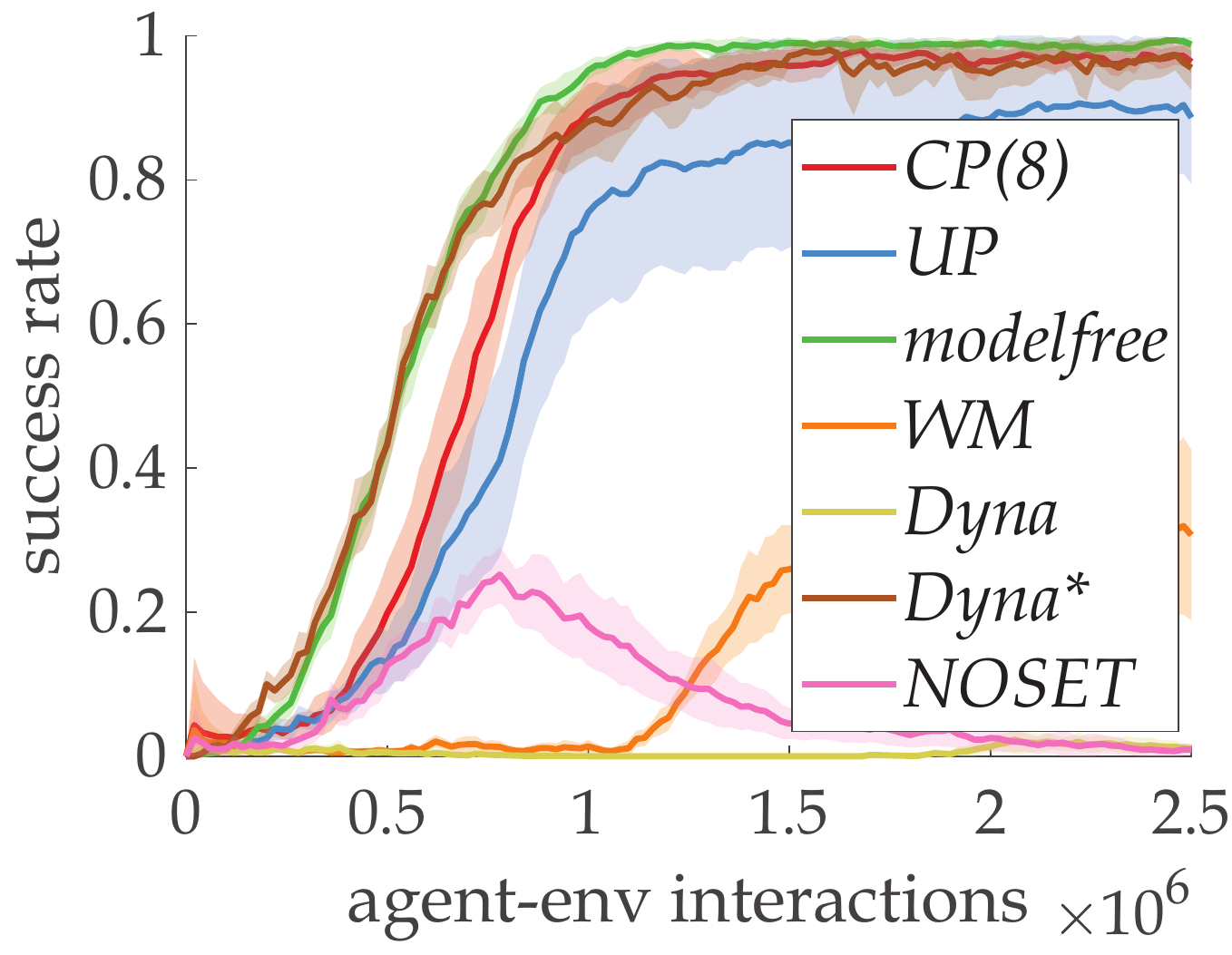}

\caption{\small \textbf{In-distribution task performance}: the $x$-axis shows the training progress ($2.5\times10^{6}$ agent-environment interactions). The $y$-axis values are generated by agent snapshots at times corresponding to the $x$-axis values. CP, UP, model-free and Dyna* agents all learn to solve the in-distribution tasks quickly. All error bars are obtained from $20$ independent runs.}
\label{fig:in_dist}
\end{SCfigure}

\begin{figure*}[htbp]
\centering
\subfloat[OOD, difficulty $0.25$]{
\captionsetup{justification = centering}
\includegraphics[width=0.242\textwidth]{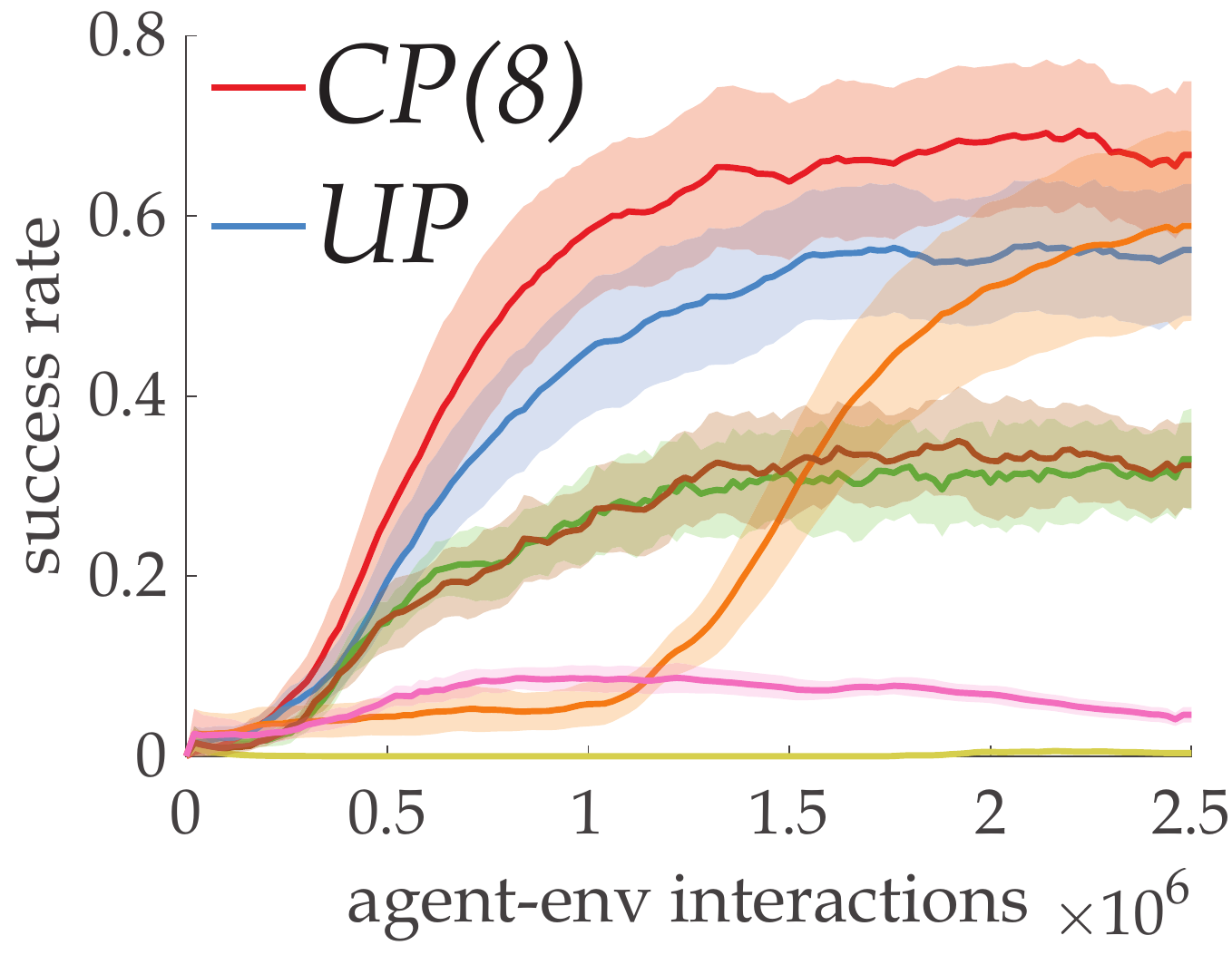}}
\hfill
\subfloat[OOD, difficulty $0.35$]{
\captionsetup{justification = centering}
\includegraphics[width=0.242\textwidth]{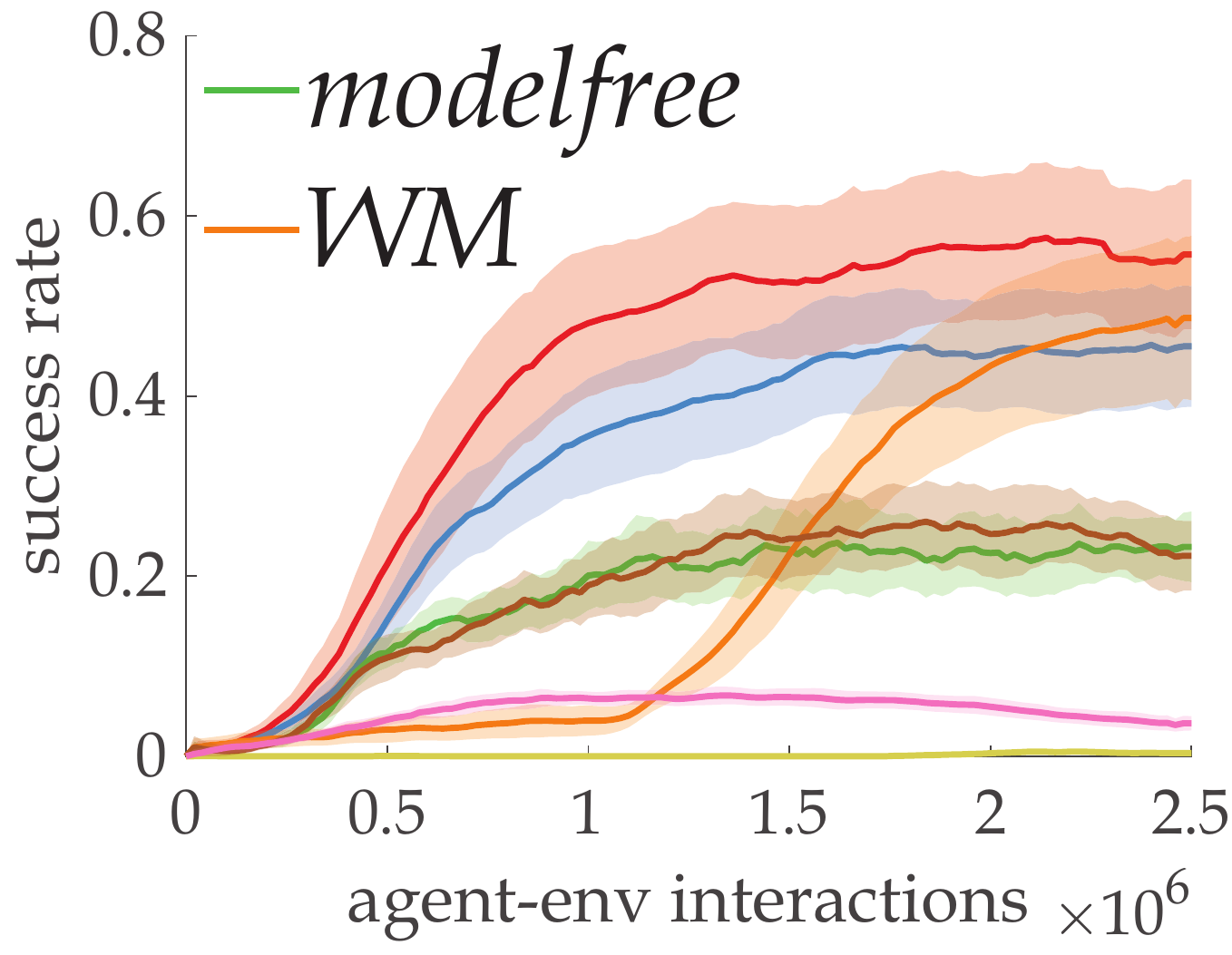}}
\hfill
\subfloat[OOD, difficulty $0.45$]{
\captionsetup{justification = centering}
\includegraphics[width=0.242\textwidth]{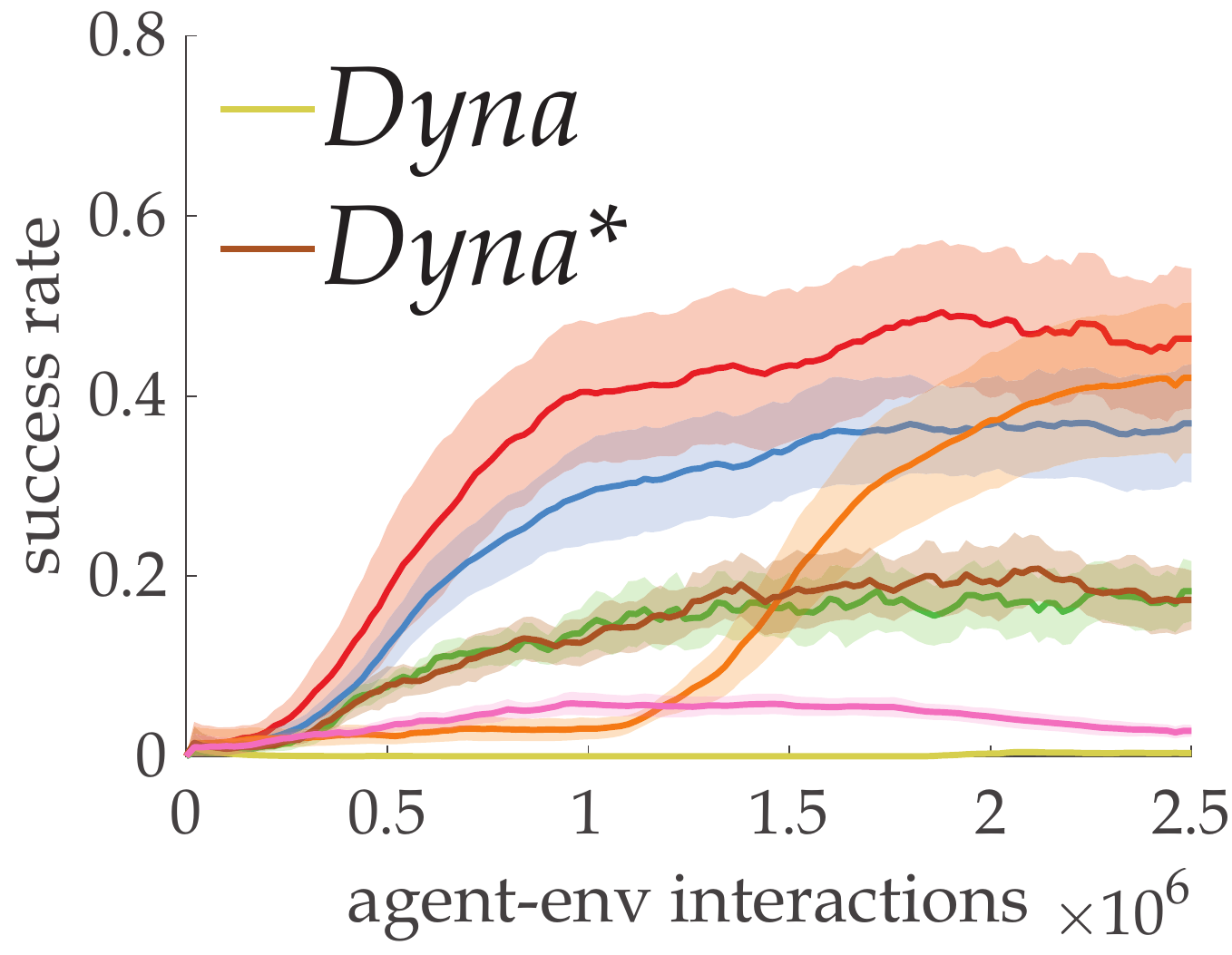}}
\hfill
\subfloat[OOD, difficulty $0.55$]{
\captionsetup{justification = centering}
\includegraphics[width=0.242\textwidth]{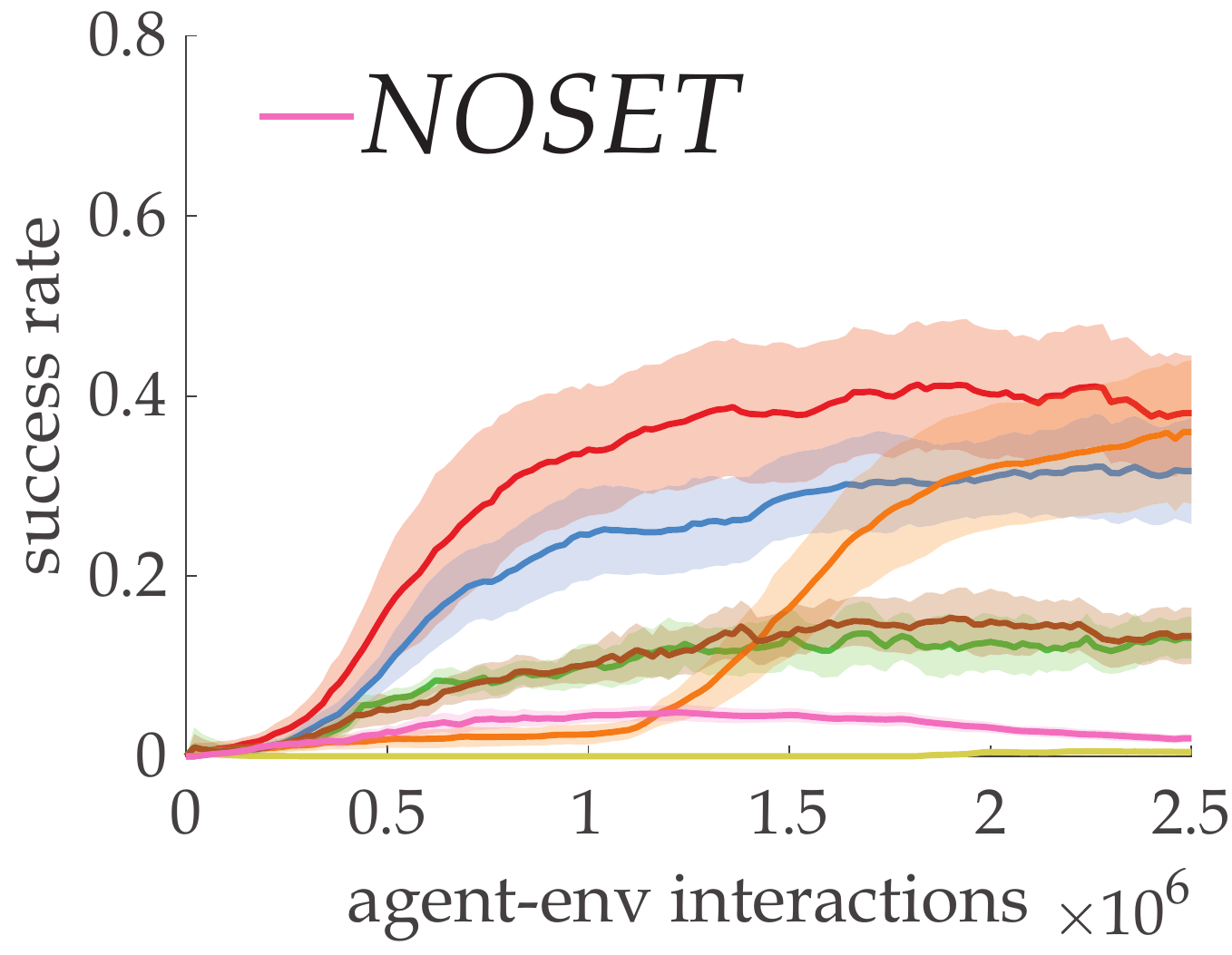}}

\caption{\small \textbf{OOD performance under a gradient of difficulty.} The figures show a consistent pattern: the MPC-based end-to-end agent equipped with a bottleneck (CP) performs the best. All error bars are obtained from $20$ independent runs.}
\label{fig:comparison_OOD}
\end{figure*}

In Figure \ref{fig:in_dist}, we present the in-distribution evaluation curves for the different agents. For UP, CP and the corresponding model-free baselines, the performance curves show no significant difference, which demonstrates that these agents are effective in learning to solve the in-distribution tasks. During the ``warm-up'' period of the WM baseline, the model learns a representation that captures the underlying dynamics. After the warm-up, the encoder and the model parameters are fixed and only the value estimator learns to predict the state-action values based on the given representation. The increase in performance is not only delayed due to the warm-up phase (during which rewards are not taken into account) but also harmed, presumably because the value estimator has no ability to shape the representation to better suit its needs. The Dyna baseline performs badly while the Dyna* baselines perform relatively well. This is likely due to the delusional transitions generated by the model at the early stages of training, from which the value estimator never recovers. However, the Dyna* baseline does not achieve satisfactory OOD performance (Figure~\ref{fig:comparison_OOD}), presumably because its planning only focuses on observed data, and hence only improves the in-distribution performance, due to insufficiently strong generalization.
The NOSET baseline performs very badly even in-distribution, per Figure \ref{fig:in_dist}. In the Appendix, we show that the NOSET baseline seems only able to perform well in a more classical, static RL setting, which may indicate that it relies on memorization. We provide more results regarding the model accuracy in the Appendix.

\subsubsection{OOD Task-Solving Performance}

The OOD evaluation focuses on testing the agents' performance in a set of environments forming a gradient of task difficulty. In Figure \ref{fig:comparison_OOD}, we present the performance error bars of the compared methods under different OOD difficulty levels. CP(8), CP with bottleneck size $n=8$, shows a clear performance advantage over UP, validating the OOD generalization capability. The Dyna* baseline, essentially the performance upper bound of Dyna-based planning methods, shows no significant performance gain in OOD tests compared to model-free methods. WM may have the potential to reach similar performance as CP, yet it needs to warm up the encoder with a large portion of the agent-environment interaction budget, if no free unsupervised phase is provided. We dive into this matter in the Appendix.

\subsubsection{Ablation}

\begin{figure*}[htbp]
\subfloat[\small \textbf{Bottleneck benefits OOD capability}: noplan(8) and noplan(UP) correspond to the CP(8) and UP variants with planning disabled during OOD tests. Comparing \textit{noplan} against \textit{modelfree}, we see that planning during training is beneficial for both value estimation and representation learning.]{
\captionsetup{justification = centering}
\includegraphics[width=0.45\textwidth]{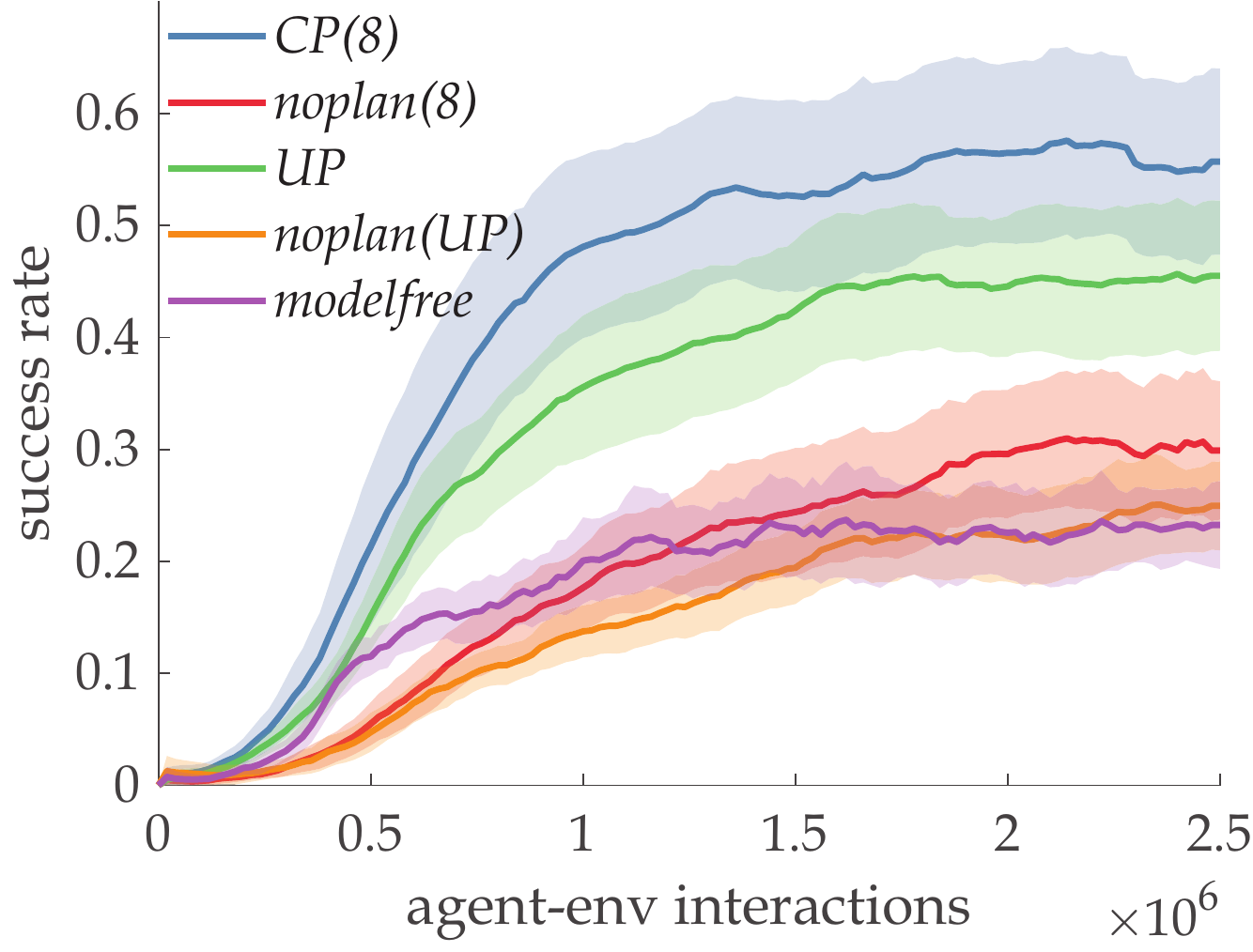}}
\hfill
\subfloat[\small \textbf{Value estimators do not generalize well in our OOD tests}: random heuristic significantly outperforms best-first heuristic OOD.]{
\captionsetup{justification = centering}
\includegraphics[width=0.45\textwidth]{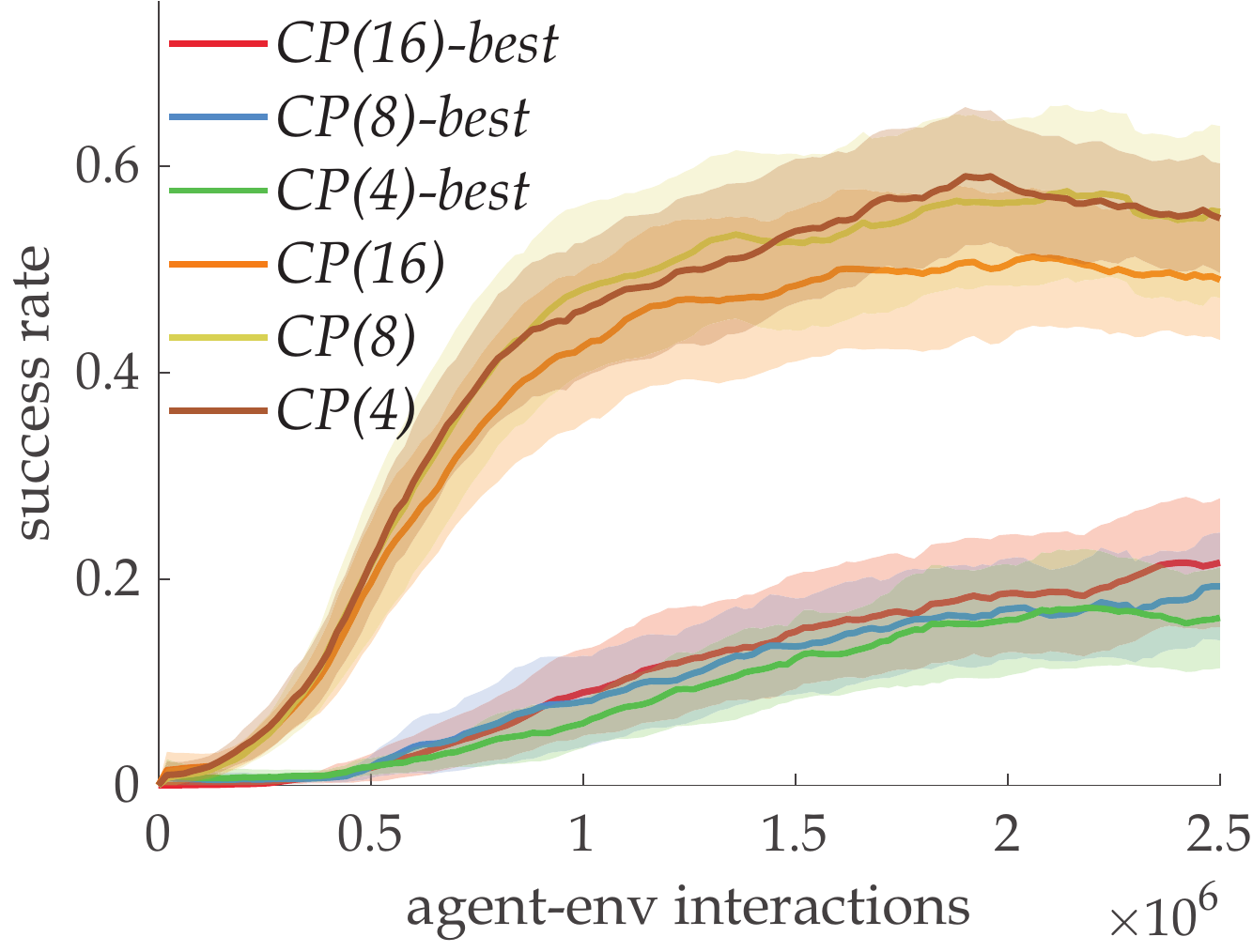}}

\caption{\small \textbf{Key ablation results}: With diff $0.35$, each error bar is obtained from $20$ independent runs.}
\label{fig:OOD35}
\end{figure*}

We validate design choices with ablation. Figure \ref{fig:OOD35} visualizes two of these experiments. For more ablation results, which include validation of the effectiveness of different model choices, and further quantitative measurements, \eg{} of OOD ability as a function of behavior optimality and model accuracy, please check the Appendix.

\subsection{Summary of Experimental Results}

With the scope limited to our experiments, the results allow us to draw these conclusions: 
\begin{itemize}[leftmargin=*]
    \item Set-based representations enable at least in-distribution generalization across different environment instances in our non-static setting, where the agents are forced to discover dynamics that are preserved across environments;
    \item Model-free methods seem to face more difficulties in solving our OOD evaluation tasks which preserved the same environment dynamics to the corresponding in-distribution training settings; 
    \item MPC exhibits better performance than Dyna in the tested OOD generalization settings;
    \item Online joint training of the representation with all the relevant signals could bring benefits to RL, as suggested in \cite{jaderberg2016unreal}. Please check Appendix \ref{sec:more_exp} for more discussions of this matter;
    \item In accordance with our intuition, transition models with bottlenecks tend to learn dynamics better in our tests. This is likely for they prioritize learning the relevant aspects, while models without bottleneck may have to waste capacity on irrelevance;
    \item
    From further experiments provided in the Appendix \ref{sec:more_exp}, we observe that bottleneck-equipped agents may also be less affected by larger environmental scales, possibly due to their prioritized learning of interesting entities.
\end{itemize}

\section{Conclusion \& Limitations}

We introduced a conscious bottleneck mechanism into MBRL, facilitated by set-based representations, end-to-end learning and tree search MPC. In the non-static RL settings, the bottleneck allows selecting the relevant objects for planning and hence enables significant OOD performance.

One limitation of our work is the experimental focus on only Minigrid environments, due to the need to validate carefully our approach. For future works, we would also like to extend these ideas to temporally extended models, which could simplify the planning task, and are also better suited as a conceptual model of C1. Finally, we note that the architectures we use are involved and can require careful tuning for new types of environments.

\section*{Acknowledgements}
Mingde is grateful for the financial support from the Fonds de Recherche du Qu\'ebec - Nature et Technologies (FRQNT). Yoshua acknowledges the financial support from Samsung Electronics and IBM.

We acknowledge the computational power provided by Compute Canada. We are also thankful for the helpful discussions with \href{https://scholar.google.com/citations?user=GwdsMdAAAAAJ&hl=en}{Xiru Zhu} (about the design of the environment generation procedure), \href{https://dyth.github.io/}{David Yu-Tung Hui} (about the bag-of-word representations, insights on BabyAI as well as about the writing of the introduction section), \href{https://www.linkedin.com/in/min-lin-08a3a422/?originalSubdomain=sg}{Min Lin} (about the design of the dynamics model as well as the early stage brainstorming) and \href{https://ianporada.github.io/}{Ian Porada} (for consistently supporting the student authors).

\clearpage
\bibliographystyle{abbrv}
\bibliography{references}

\newpage

\begin{appendices}

\section{Architecture Details}

\subsection{Birdseye View of Overall Design}
\begin{figure*}[htbp]
\centering
\captionsetup{justification = centering}
\includegraphics[width=0.99\textwidth]{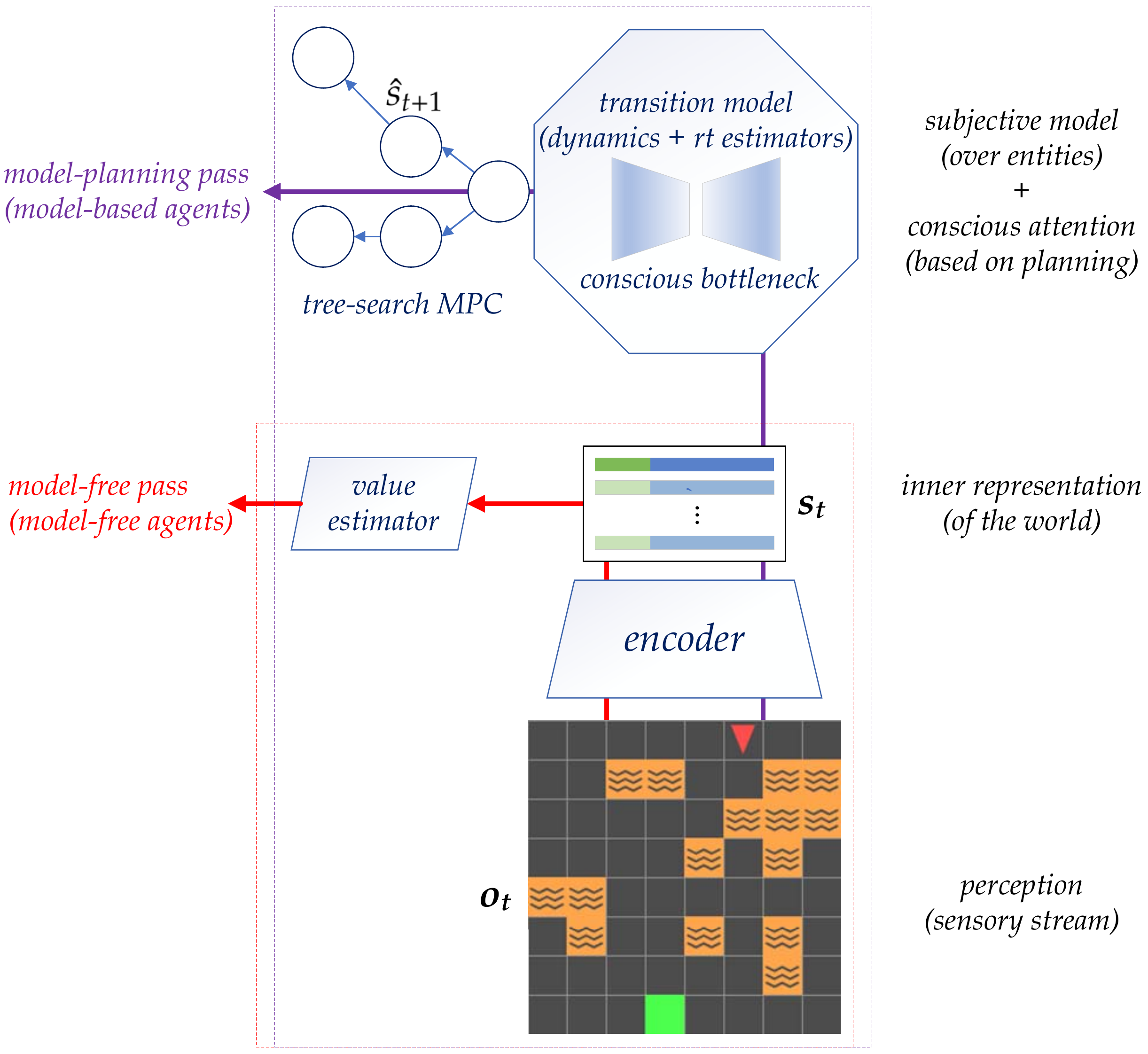}
\caption{\small Overall organization of the proposed components for the CP agent. The transition model includes the reward-termination estimator, the dynamics estimator and the optionally the conscious bottleneck. Drawing similarity to the human mind, the $3$-layered design corresponds naturally to human perception, inner representation and the conscious planning models.}
\label{fig:birdseye}
\end{figure*}

We present the organization of the components for the proposed CP agent in Fig \ref{fig:birdseye}. For the model-free baseline agent, we contributed the design of the state set encoder and the set-based value estimator. For the model-based agent, we additionally devised the design of two transition models, one with the conscious bottleneck and another without.

\subsection{Action-Conditioned Transformer Layer}
\begin{figure*}[htbp]
\centering
\captionsetup{justification = centering}
\includegraphics[width=0.95\textwidth]{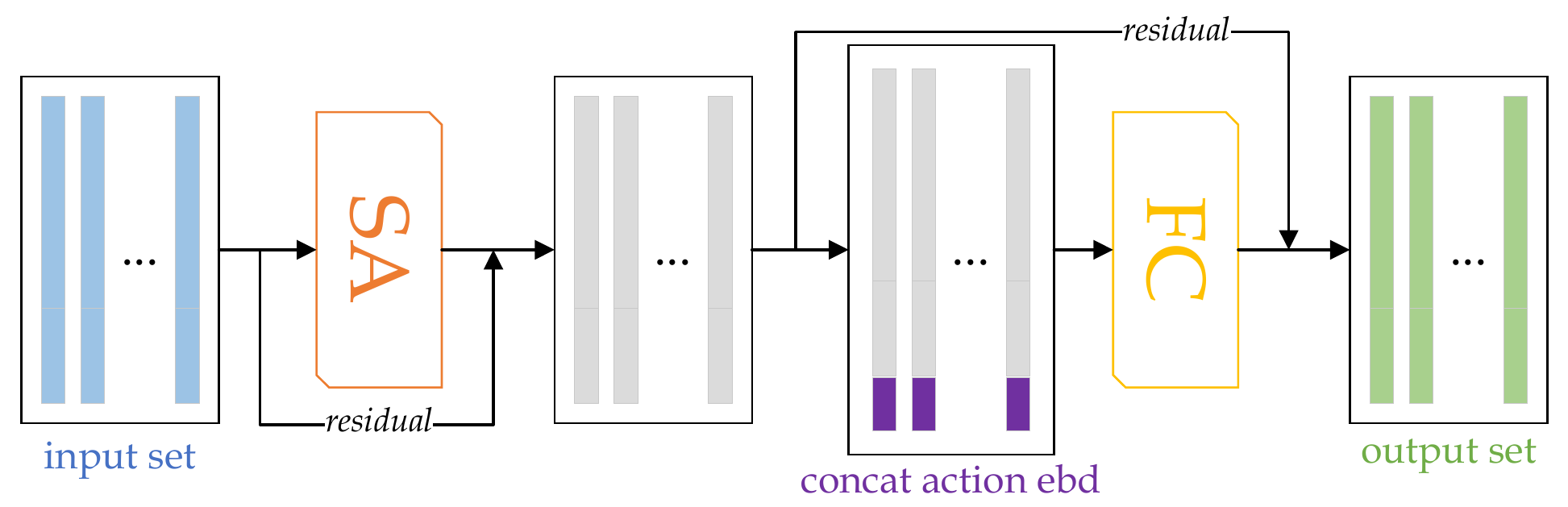}
\caption{\small The computational flow of the action-conditioned transformer layer: compared to the classical transformer layers, we concatenate additionally the action embedding to the end of every intermediate object embeddings in the FC pass. The FC pass facilitates $X' = X + f(cat[X,a])$, where $X$ is the set of objects input to the FC part of the action-conditioned transformer layer, $cat([X,a]])$ is the concatenation of action embedding $\bm{a}$ to every object embedding in $X$ and $X'$ is the output set. Note that $f$ downscales the dimensionality of its input to match $X$.}
\label{fig:layer_transformer_conditioned}
\end{figure*}

A classical transformer layer consists of two consecutive sub-layers, the multi-head SA and the fully connected, each containing a residual pass. Similar to the processing of the positional embedding, we first embed the discrete actions into a vector and then concatenate it to every intermediate object output by the SA sub-layer. This way, each transformer layer becomes action-conditioned. An illustration of the component is provided as Figure \ref{fig:layer_transformer_conditioned}.

\subsection{Bottleneck Dynamics}

The architecture for the bottleneck dynamics (the dynamics operator that simulates $\hat{c}_{t+1}$ from $c_t$, $a_t$) is a stack of action-conditioned transformer layers.

\subsection{Reward-Termination Estimator}

\begin{figure*}[htbp]
\centering
\captionsetup{justification = centering}
\includegraphics[width=0.95\textwidth]{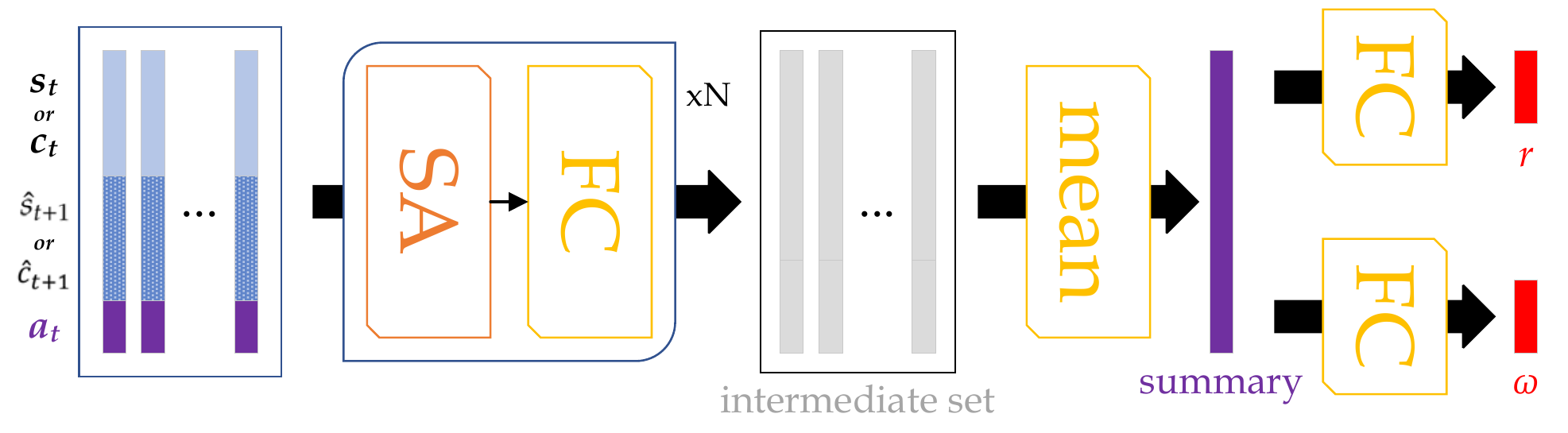}
\caption{\small Design of the reward-termination estimator: the state / bottleneck set, the imagined state / bottleneck set as well as the embedding of the action are aligned and concatenated to predict the two outputs. When there is a conscious bottleneck, $c_t$ comes from the \compression{}, $\hat{c}_{t+1}$ is the output of rolling $c_t$ into the dynamics model with $a_t$; When there is not, $\hat{s}_{t+1}$ comes from the forward simulation of the model. With deterministic, it is sufficient to predict the reward and termination with only $s_t$ and $a_t$. This design would be compatible if the dynamics simulation could handle stochastic dynamics.}
\label{fig:rt_estimator}
\end{figure*}

In the experiments, we wanted functional architectures with minimal sizes for all the components. Thus, globally for the set-input architectures, we have limited the depth of the transformer layers to be $N=1$ wherever possible. The FC components are MLPs with $1$-hidden layer of width $64$. Exceptionally, we find that the effectiveness of the value estimator needs to be guaranteed with at least $3$-transformer layers. For the distributional output, while the value estimator has an output of $4$ atoms, the reward estimator has only $2$.

\subsection{Bottleneck \titlecap{\compressor{}}}
\begin{figure}[H]
\centering
\captionsetup{justification = centering}
\includegraphics[width=0.95\textwidth]{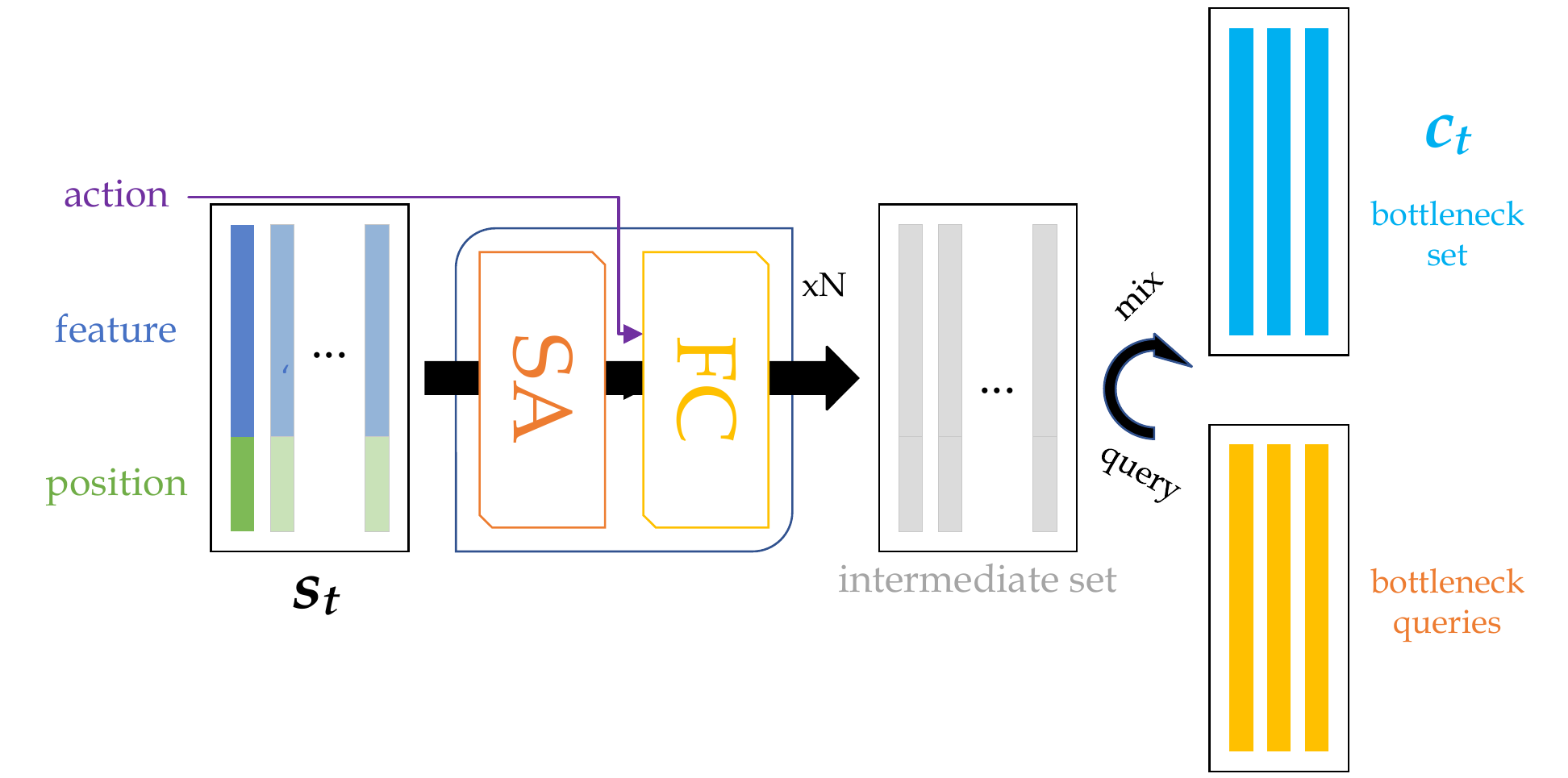}
\caption{\small Design of the Bottleneck Compressor: the bottleneck set $c_t$ is obtained by querying the whole set $s_t$ with a learned query set of size $k$, using semi-hard multi-head attention. The \compression{} is conditioned on the chosen action. Please refer to Section \ref{sec:attention} for more details of the query operation.}
\label{fig:compressor}
\end{figure}

\subsection{Bottleneck \titlecap{\decompressor{}}}
\begin{figure}[H]
\centering
\captionsetup{justification = centering}
\includegraphics[width=0.95\textwidth]{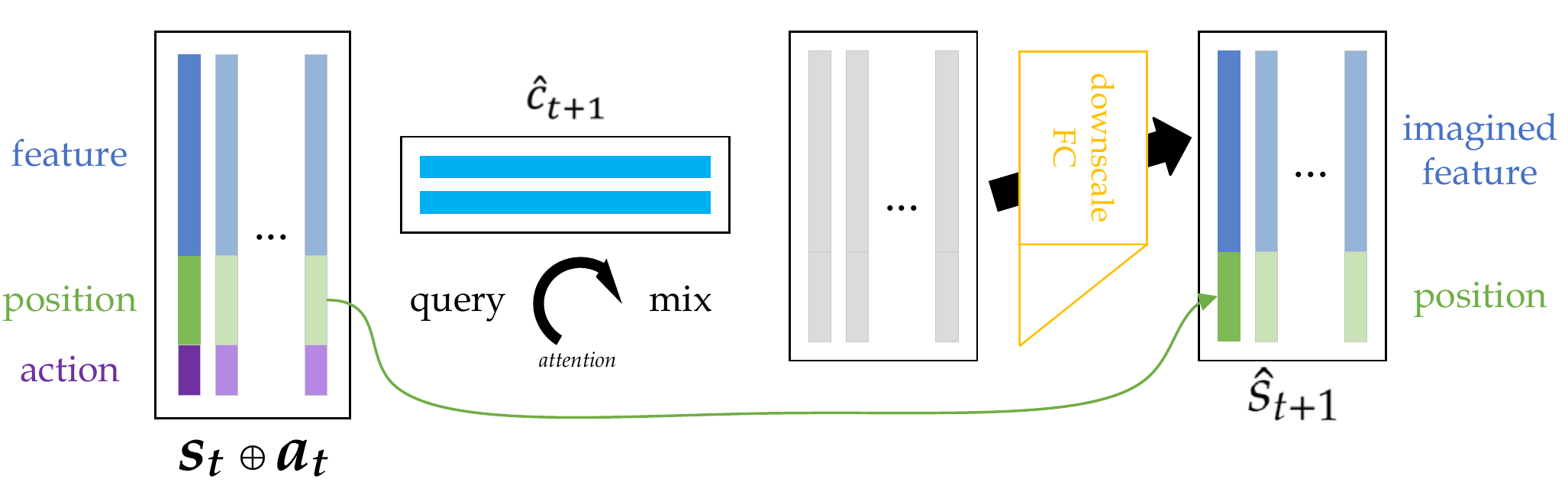}
\caption{\small Design of the Bottleneck \titlecap{\decompressor{}}: $\hat{s}_{t+1}$ is generated by using the action-augmented $s_t$ to query the imagined bottleneck set $\hat{c}_{t+1}$. Note that there is the similar operation of downscaling objects to features and copying the positional tails. Please refer to Section \ref{sec:attention} for more details of the query operation.}
\label{fig:decompressor}
\end{figure}


\section{Prerequisites}

Here, we introduce some prerequisites for better understanding of the used operations.

\subsection{Attention} 
\label{sec:attention}

One of the most important permutation invariant operations on sets of objects is the \textit{attention querying}, which leads to the variants of attention mechanisms \cite{bahdanau2014neural}. Here, we revisit a generic set query procedure:

For an object to \textit{query} another set of objects, the following steps are taken:
\begin{enumerate}[leftmargin=*]
    \item The object is transformed into a query vector. This is generally done via linear transformations. 
    \item The set of objects is independently transformed into two other sets of the same cardinality, named the key set and the value set, respectively. 
    \item The query vector now compares itself with each key vector in the key set according to some similarity function, \eg{} scaled dot product, and obtain a vector of un-scaled ``attention weights'' which is later normalized into a vector with unit $L_1$ norm.
    \item The value vectors are weighted by the normalized attention weight vector and combined (typically by linear transformations), yielding the output vector; Querying a set with another set is no different from independently applying the described procedure multiple times. The number of outputs always matches the size of the query set.
\end{enumerate}

Using a set to query itself using the above procedure yield the so-called ``self-attention''. Using multiple groups of linear transformations and computing the final output from the ensemble of query results is called ``multi-head attention''. If we erase the lowest attention weights and keep only the top-$k$ ones before the $L_1$ re-normalization, the resulting method is called ``semi-hard'' attention: for the top-$k$ matches, the attention is soft while for the bad matches, the attention is hard.

\subsection{Distributional Outputs} 
\label{sec:distoutputs}

In this paper, we adopt distributional outputs for the designs of the value and reward estimators. In a nutshell, a distributional output converts a scalar prediction problem with a $1$-dimensional output to a predicting a distribution, which is later converted to a scalar by a weighted sum corresponding to the support. This greatly alleviates the problem introduced by the difference in the magnitude of outputs. Please check \cite{bellemare2017distributional} for more details and \cite{hessel2017rainbow} for a representative use case.

\section{Experiment Insights}

\paragraph{Integer Observations}
For MiniGrid worlds, the observations are consisted of integers encoding the object and the status of the grids. We found that for the UP models with these integer observations, the transformer layers are not sufficiently capable to capture the dynamics. Such problem can be resolved after increasing the depth of the FC layer depth by another hidden layer. This is one of the reasons why we prioritized on using CP models for the observation-level learning of Dyna, \ie{} CP models can handle integer features without deepening.

Similarly, we have tested the effect of increasing the depth of the linear transformations in SA layers. We did not observe significance in the enhancement of the performance, in terms of model learning or RL performance.

\paragraph{Addressing Memorization with Noisy Shift}
We discovered a generic trick to enforce better generalization based on our state-set encoding: if we use fixed integer-based positional tails which correspond to the absolute coordinates of the objects, we can add a global noise to all the $x$ and $y$ components in a set whenever one is encoded. By doing so, the coordinate systems would be randomly shifted every time the agent updates itself. Such shifts would render the agent unable to memorize based on absolute positions. This trick could potentially enhance the agents' understanding of the dynamics even if in a classical static RL setting, under which the environments are fixed.

\section{Experiment Configurations}
The source code for this work is implemented with TensorFlow 2.x and open-source at \url{https://github.com/PwnerHarry/CP}.

Multi-Processing: we implement a multi-process configuration similar to that of Ape-X \cite{horgan2018distributed}, where $8$ explorers collect and sends batches of $64$ training transitions to the central buffer, with which the trainer trains. A pause signal is introduced when the trainer cannot consume fast enough \st{} the uni-process and the multi-process implementation have approximately the same performance, excluding the wall time.

Feature Extractor: We used the Bag-Of-Word (BOW) encoder suggested in \cite{hui2020babyai}. Since the experiments employ a fully-observable setting, we did not use frame stack. In gym-MiniGrid-BabyAI environments, a grid is represented by three integers, and three trainable embeddings are created for the BOW representation. For each object (grid), each integer feature would be first independently transformed into embeddings, which is then mean-pooled to produce the final feature. The three embeddings are learnable and linear (with biases).

Stop criterion: Each runs stops after $2.5 \times 10^{6}$ agent-environment interactions.

Replay Buffer: We used prioritized replay buffer of size $10^{6}$, the same as in \cite{hessel2017rainbow}. We do not use the weights on the model updates, only the TD updates.

Optimization: We have used Adam \cite{kingma2014adam} with learning rate $2.5 \times 10^{-4}$ and epsilon $1.5\times {10}^{-4}$. The learning rate is the same as in \cite{mnih2015human}. Our tests show that using $6.25 \times 10^{-5}$, as suggested in \cite{hessel2017rainbow}, would be too slow. The batch size is the same for both value estimator training and model training, $64$. The training frequency is the same as in \cite{hessel2017rainbow}: every $4$ agent-environment interactions.

$\gamma$: Same as in \cite{hessel2017rainbow}. $0.99$.

Transformers: For the SA sublayers, we have used $8$ heads globally. For the FC sublayers, we have used 2-layer MLP with $64$ hidden units globally. All the transformer related components have only $1$ transformer layer except for that of the value estimator, which has $3$ transformer layers before the pooling. We found that the shallower value estimators exhibit unstable training behaviors when used in the non-static settings.

Set Representation: The length of an object in the state set has length $32$, where the feature is of length $24$ and the positional embedding has length $8$. Note that the length of objects must be dividable by the number of heads in the attentions. The positional embeddings are trainable however their initial values are constructed by the absolute $xy$ coordinates from each corner of the gridworld ($4 \times 2 = 8$). We found that without such initialization the positional embedding would collapse.

Action Embedding: Actions are embedded as one-hot vectors with length 8.

Planning steps: for each planning session, the maximum number of simulations based on the learned transition model is 5.

Exploration: $\epsilon$ takes value from a linear schedule that decreases from $0.95$ to $0.01$ in the course of $10^{6}$ agent-environment interactions, same as in \cite{hessel2017rainbow}. For evaluation, $\epsilon$ is fixed to be $10^{-3}$.

Distributional Outputs: We have used distributional outputs \cite{bellemare2017distributional} for the reward and value estimators. $2$ atoms for reward estimation (mapping the interval of $[0, 1]$) and $4$ atoms for value estimation (mapping the interval of $[0, 1]$).

Regularization: We find that layer norm is crucial to guarantee the reproducibity of the performance with set-representations. We apply layer normalization \cite{ba2016layer} in the sub-layers of transformers as well as at the end of the encoder and model dynamics outputs. This applies for the NOSET baseline as well.

Modelfree baseline: We did not use the full Rainbow agent \cite{hessel2017rainbow} as the baseline for that we want to keep our agent as minimalist as possible. The agent does not need the dueling head and the noisy net components to perform well, according to our preliminary ablation tests.

\section{More Experimental Analyses}
\label{sec:more_exp}

\subsection{In-Distribution Model Accuracy}
We intend to demonstrate how well the bottleneck set captures the underlying dynamics of the environments. For each transition, we split the grid points into two partitions: one containing all relevant objects that changed during the transition or have an impact on reward or termination, while the other contains the remaining grid points. As a result, the dynamics error is split into into two terms which correspond to the accuracy of the model simulating the relevant and irrelevant objects respectively.

Acknowledging the differences in the norm of the learned latent representations, we use the element-wise mean of $L_1$ (absolute value) difference between $\hat{s}_{t+1}$ and $s_{t+1}$ but normalize this distance by the element-wise mean $L_1$ norm of $s_{t+1}$, as a metric of model accuracy, which we name the \textit{relative L1}. This metric shows the degree of deviation in dynamics learning: the lower it is, the more consistent are the learned and observed dynamics.

\begin{figure*}[htbp]

\subfloat[Split of Relative L1 error]{
\captionsetup{justification = centering}
\includegraphics[width=0.48\textwidth]{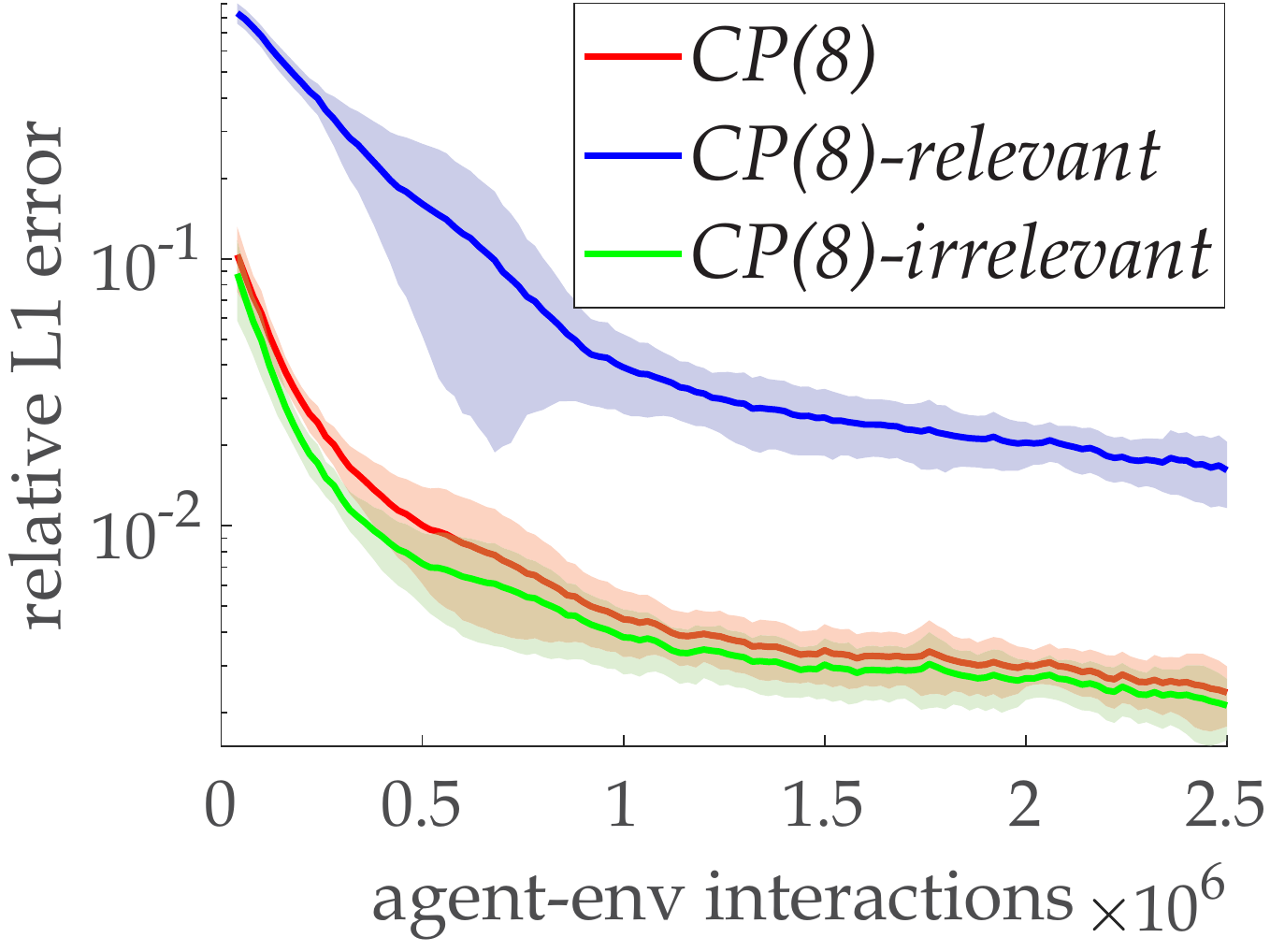}}
\hfill
\subfloat[Comparison of Relative L1 error]{
\captionsetup{justification = centering}
\includegraphics[width=0.48\textwidth]{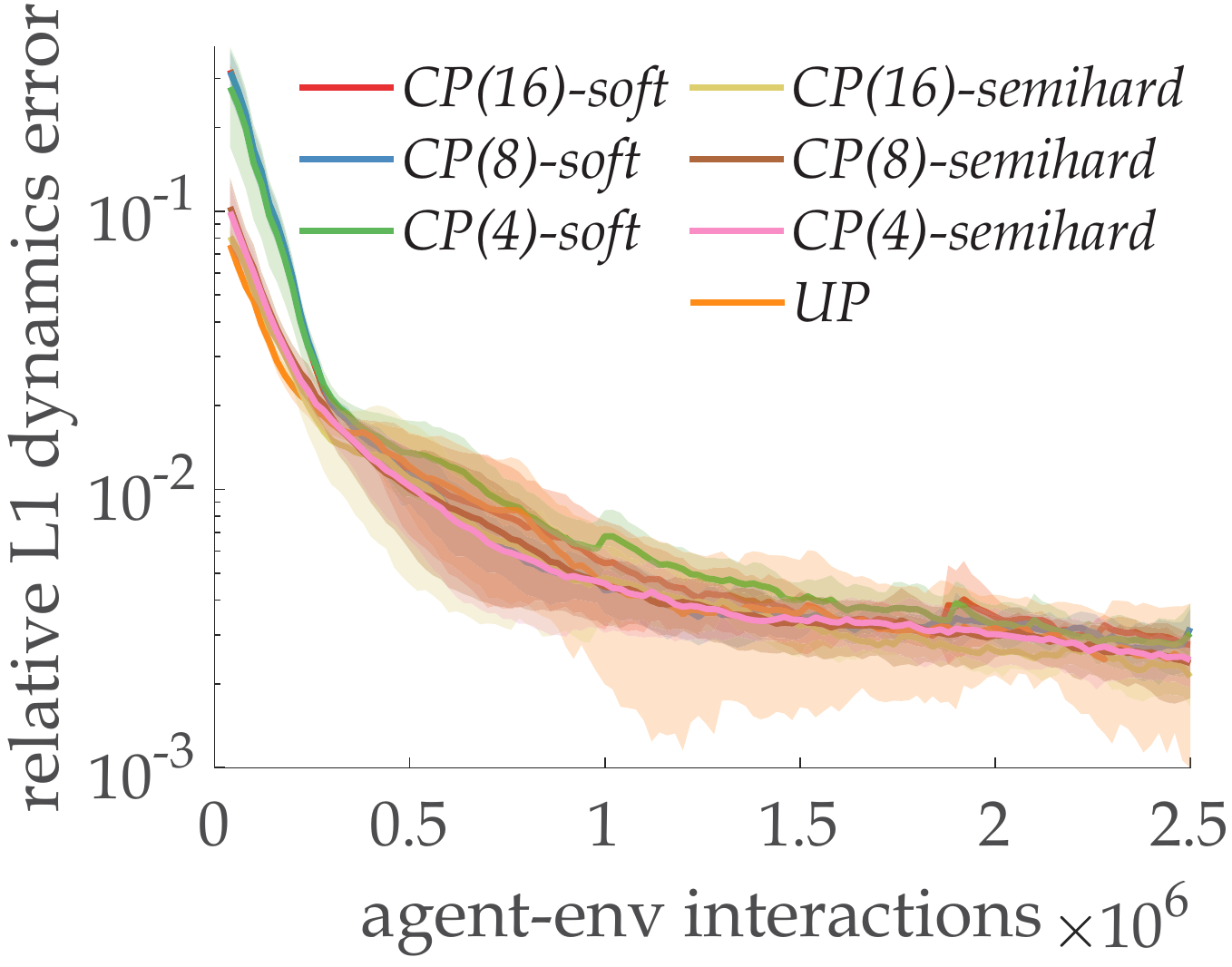}}

\caption{\small Curves showing in-distribution evaluation: Each band shows the mean curve (bold) and the standard deviation interval (shaded) obtained from $20$ independent seed runs. a) Partitioning of the relative L1 dynamics prediction errors into that of the relevant objects and the irrelevants: The difference in the errors shows that the bottleneck learns to ignore the irrelevance while prioritizing on the relevant parts of the state; b) Comparison of the overall relative L1 errors (not partitioned). For CP variants, the numbers in the parentheses correspond to the bottleneck sizes and the suffixes the types of attention for the bottleneck \compression{}. Semi-hard attention learns more quickly than soft attention at early stages but they both converge to similar accuracy levels. This is likely due to the fact that semi-hard attention is forced to pick few objects and thus to ignore irrelevant objects even at early stages of training.}
\label{fig:in_dist_acc}
\end{figure*}

Figure \ref{fig:in_dist_acc} (a) presents the \textit{relative L1} error of the a CP configuration during the in-distribution learning. With the help of the bottleneck, the error for the irrelevant parts converge very quickly while the model focuses on learning the relevant changes in the dynamics. Additionally, we provide the model accuracy curves of the WM and Dyna baselines in the Appendix.

For reward and termination estimations, our results show no significant difference in estimation accuracy with different bottleneck sizes. However, they do seem to have significant impact on the dynamics learning. In Figure \ref{fig:in_dist_acc} (b), we present the convergence of the relative dynamics accuracy of different CP and UP agents. CP agents learn as fast as UP, which indicates low overhead for learning the \compression{} and \decompression{}.

\subsection{More Ablation Results}
\begin{figure*}[htbp]

\subfloat[\small Attention Type: semi-hard attention outperforms better when used in bottleneck \compression{}]{
\captionsetup{justification = centering}
\includegraphics[width=0.45\textwidth]{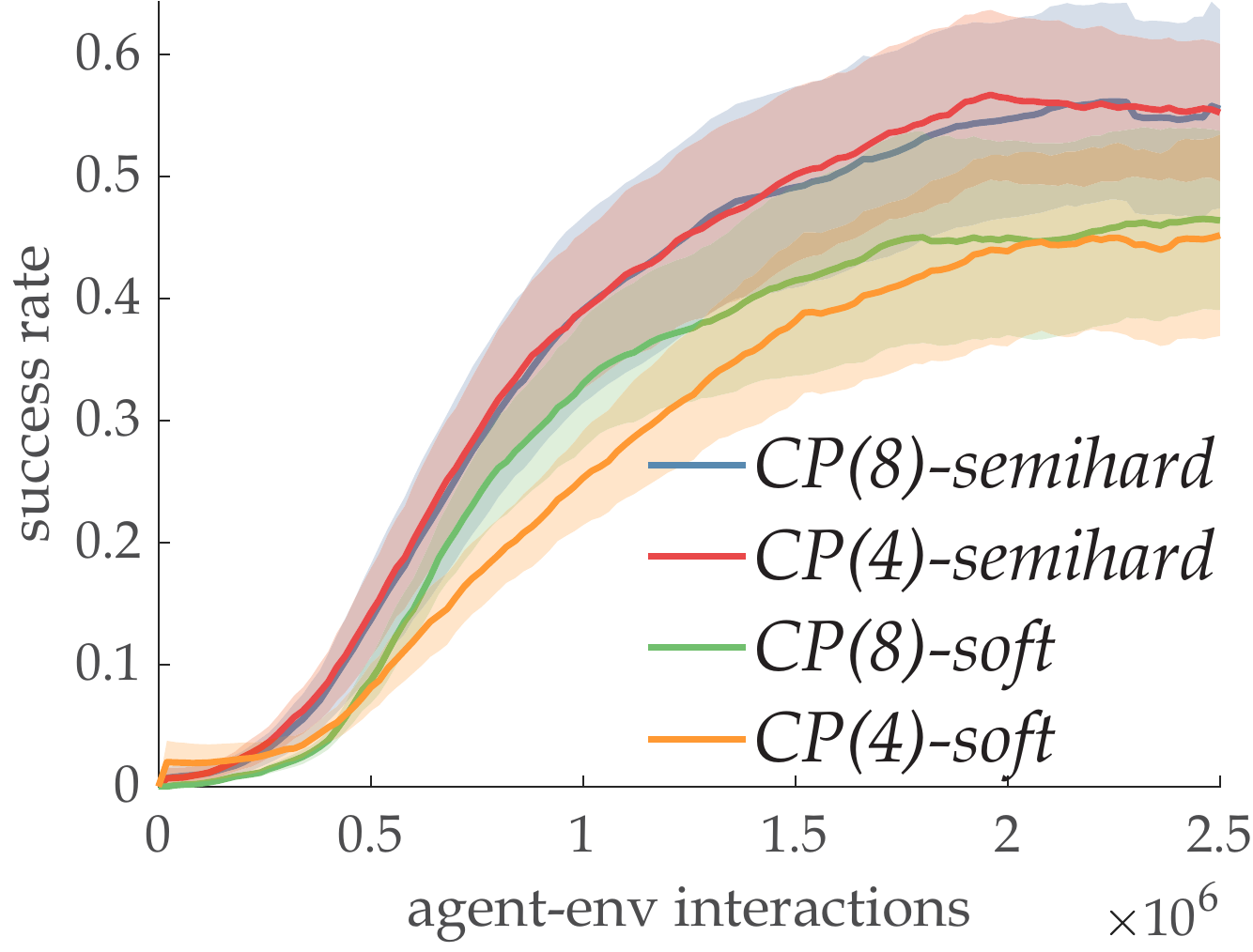}}
\hfill
\subfloat[\small Bottleneck Size: bottleneck sizes $4$ and $8$ perform similarly the best within $\{2, 4, 8, 16\}$. Also, the performance with bottlenecks is consistently better than that without (UP), showing the bottlenecks' effectiveness for OOD generalization]{
\captionsetup{justification = centering}
\includegraphics[width=0.45\textwidth]{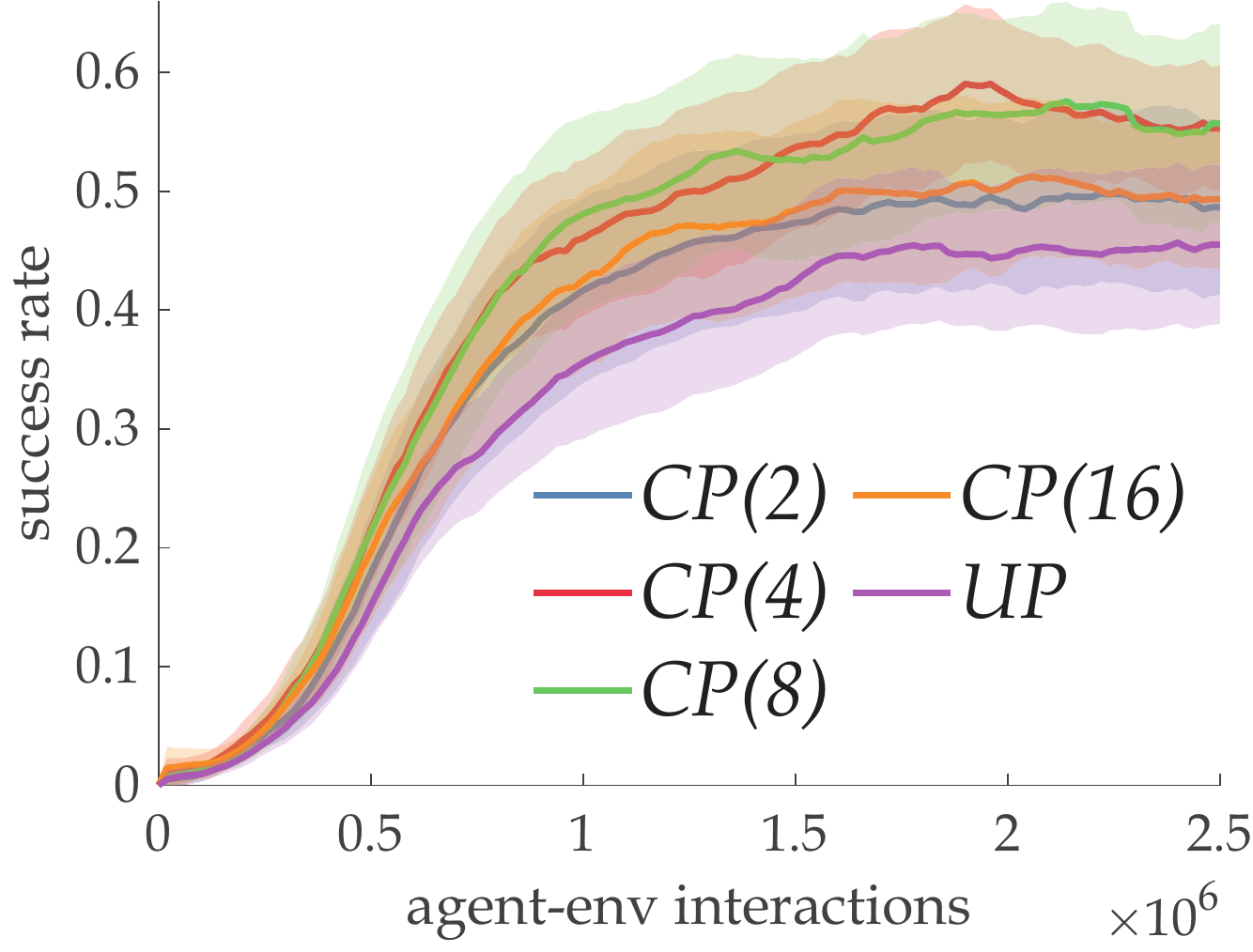}}

\subfloat[\small Action Quality: we record if the actions taken by the methods are optimal. For in-distribution evaluation, the methods both perform well. Interestingly, the model-free agent performs superior possibly due to its simple value-based greedy policy. However in OOD evaluation, only the CP agent with the random heuristic shows neither significant deterioration nor signs of overfit in the action qualities.]{
\captionsetup{justification = centering}
\includegraphics[width=0.45\textwidth]{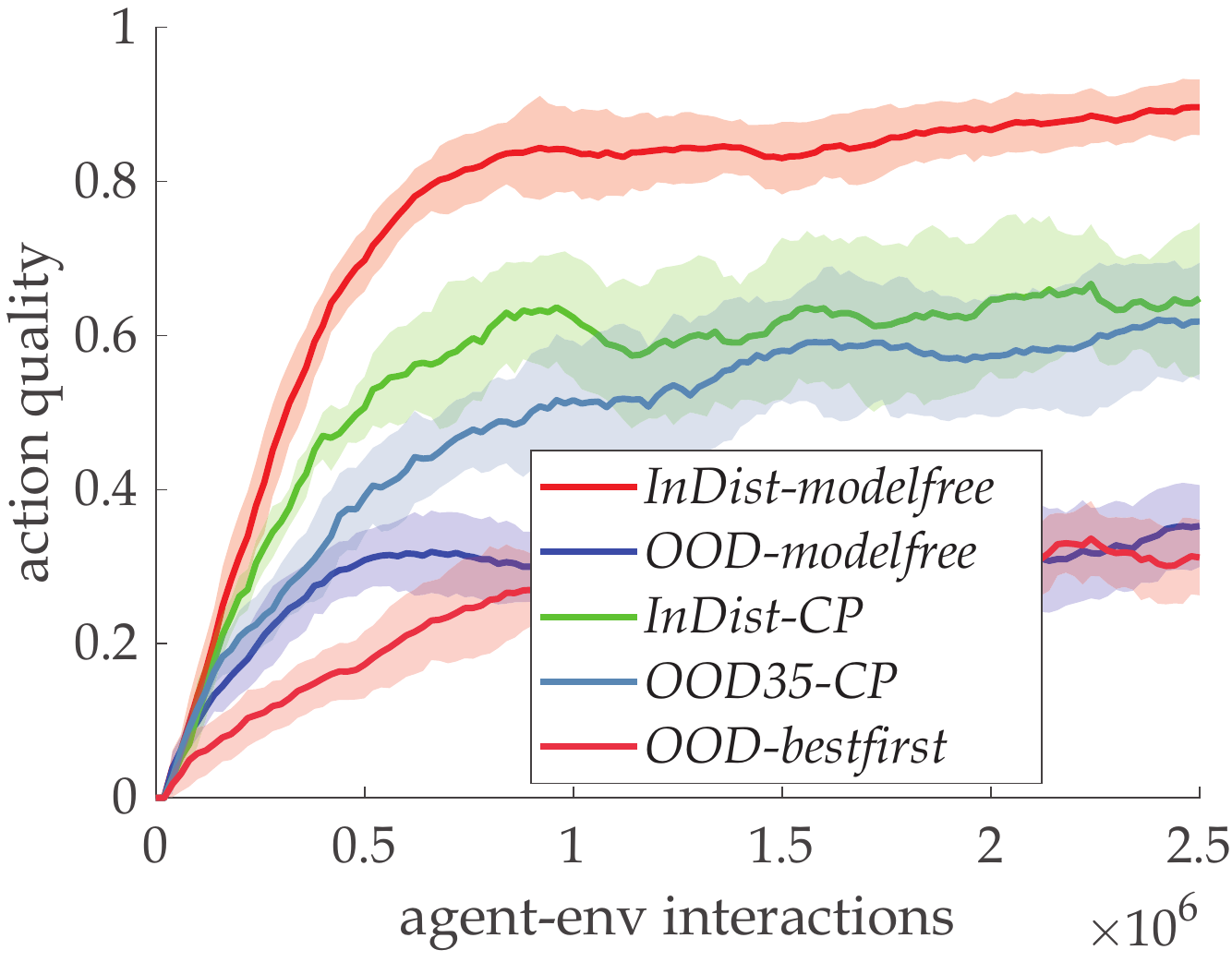}}
\hfill
\subfloat[\small Tree Search Dynamics Accuracy: the curves show the cumulative L1 error of the chosen trajectory during tree search. These are obtained by comparing the imagined states simulated through multi-step planning with the help of a perfect environment model. The curves show no signs of overfit as the cumulative trajectorial dynamics accuracy during OOD evaluation is growing over time.]{
\captionsetup{justification = centering}
\includegraphics[width=0.45\textwidth]{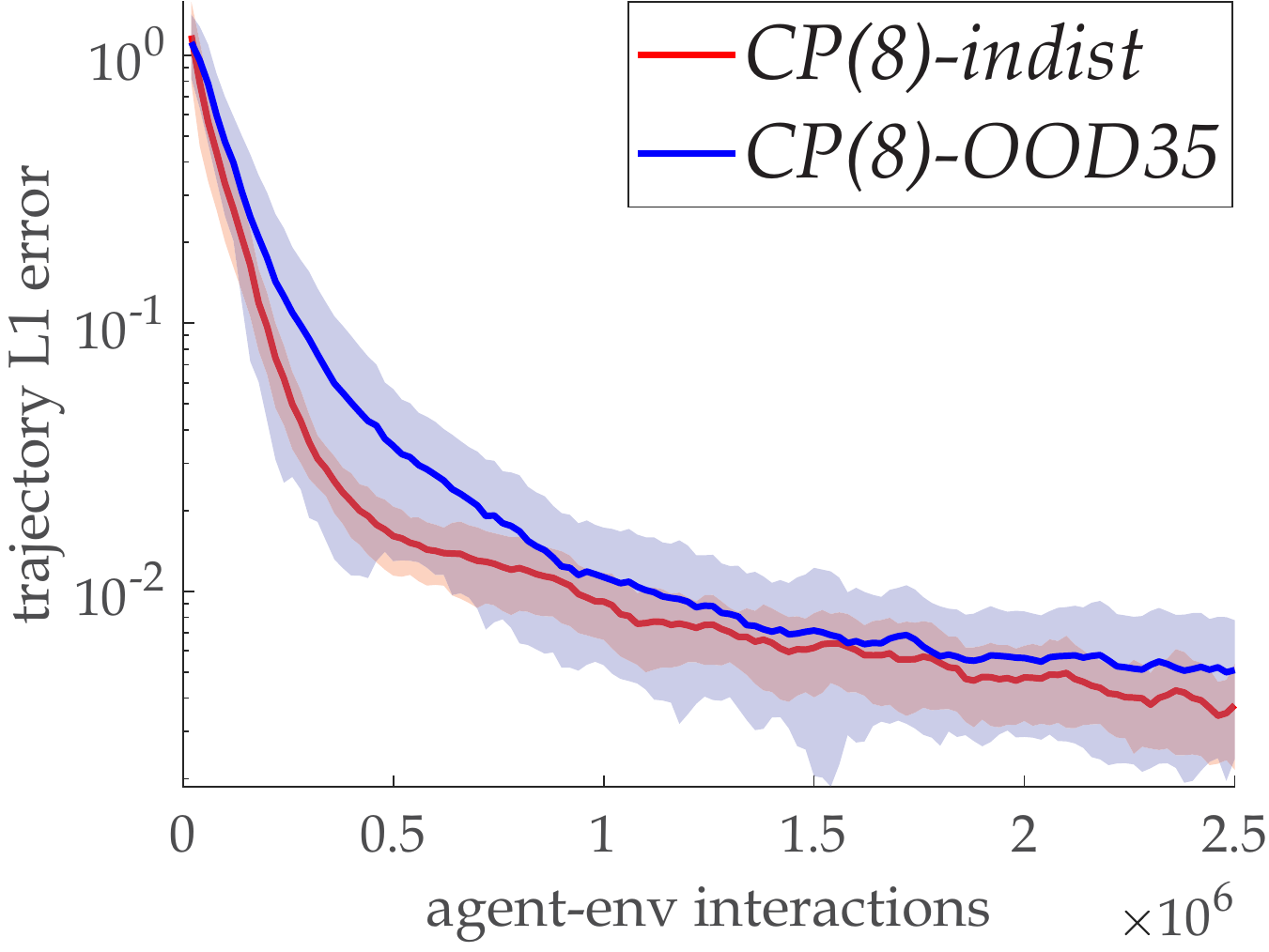}}

\caption{\small Ablation results with difficulty $0.35$: each band is consisted of the mean curve and the standard deviation interval shades obtained from $20$ independent seed runs.}
\label{fig:OOD35_more}
\end{figure*}

Figure \ref{fig:OOD35_more} visualizes more experiments which highlight the effectiveness of the bottleneck's contribution towards OOD generalization.

\subsection{Planning Steps}
\begin{figure*}[htbp]
\centering
\captionsetup{justification = centering}
\includegraphics[width=0.45\textwidth]{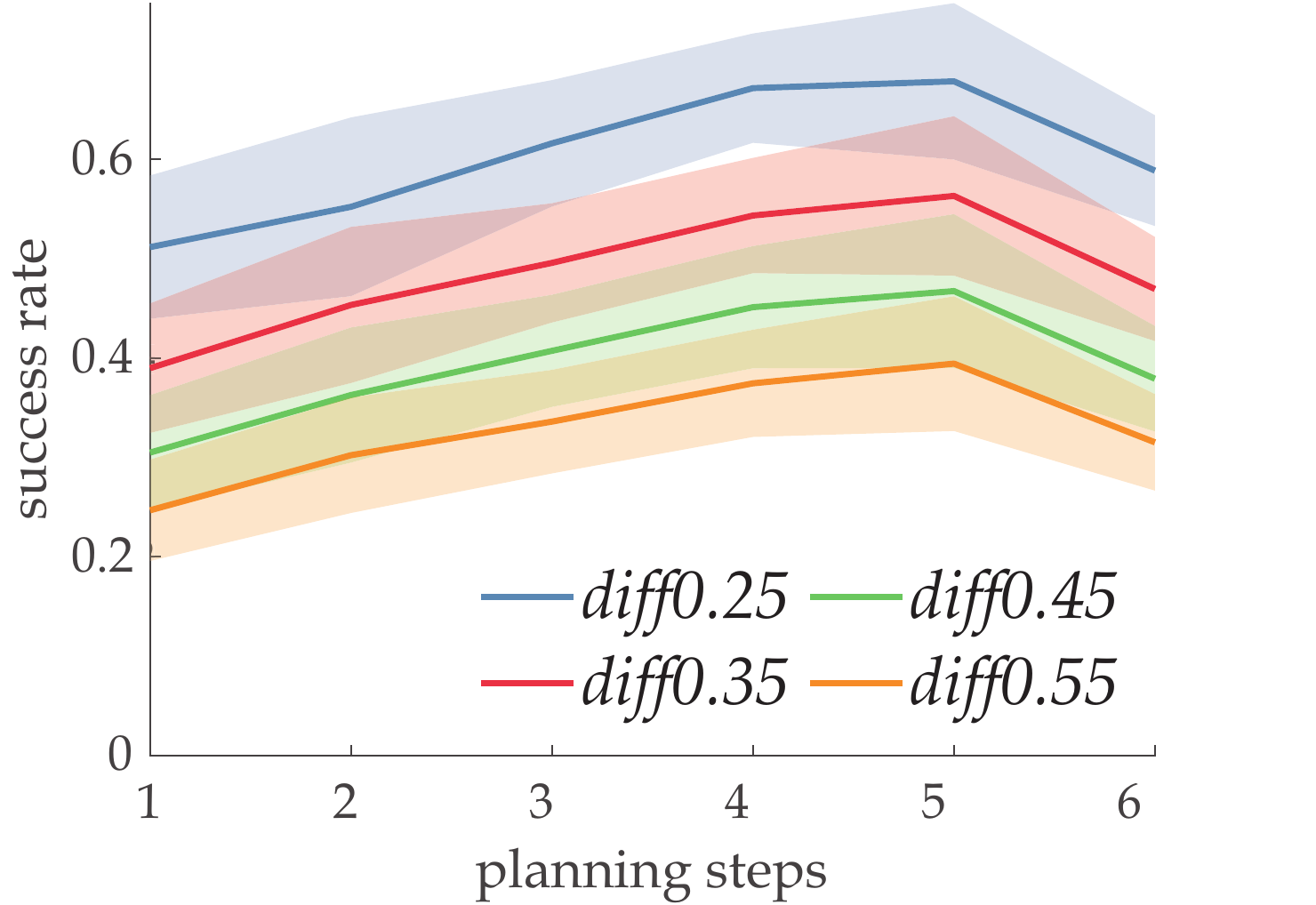}
\caption{\small Success rate of CP(8) agent under OOD difficulty $0.35$. Note that for each agent variant, the planning steps used in training and OOD evaluation are the same.}
\label{fig:nshaped_steps}
\end{figure*}

Intuitively we know there should be a good value for the planning step hyperparameter. If the planning steps are too few, then the planning would have little gain over model-free methods. While if the planning steps are too many, we suffer from cumulative planning errors and potentially prohibitive wall time. We tried different number of planning steps for $8$-picks semi-hard CP. Note that the planning steps during training and OOD evaluation are equivalent. Such particular choice is to make sure that the planning during evaluation would be carried out to the same extent during training. The results visualized in \ref{fig:nshaped_steps} suggested that $5$ planning steps achieves the best performance in OOD with difficulty 0.35.

\subsection{Action Regularization}
\begin{figure*}[htbp]
\centering
\captionsetup{justification = centering}
\includegraphics[width=0.45\textwidth]{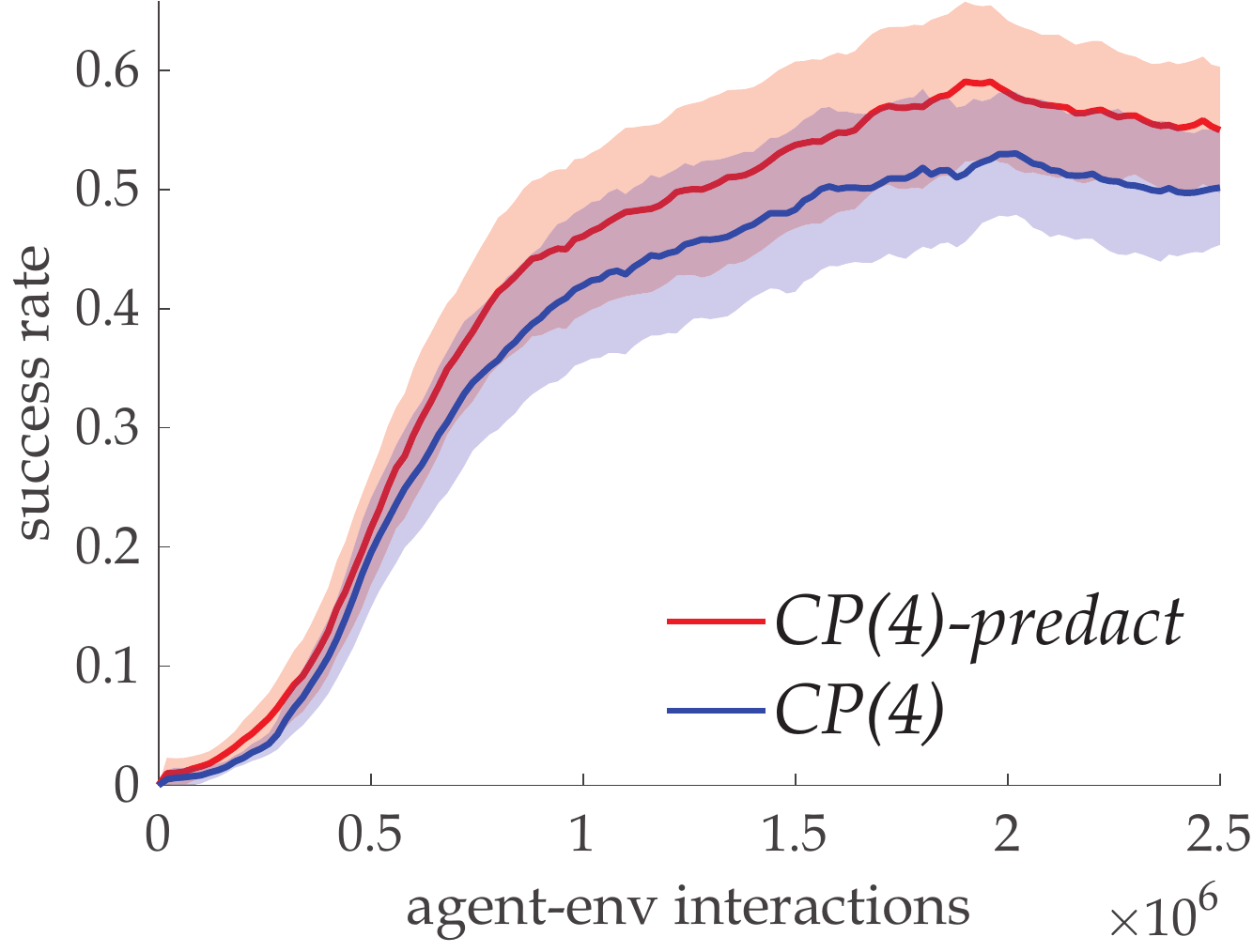}
\caption{\small Impact on the success rate of CP(4) agents under OOD evaluation with difficulty $0.35$ by the action regularization loss in the bottleneck. The ``predact'' configuration is by default enabled in the main manuscript, \ie{} all the CP results shown except in this figure has action regularization enabled. Each point of the band correspond to the mean and standard deviation of the success rate of OOD evaluation during the last $5\times 10^{5}$M agent-environment interactions (last 20\% training stage).}
\label{fig:predact}
\end{figure*}

We applied an additional regulatory loss that predicts the action $a_t$ with $c_t$ and $\hat{c}_{t+1}$ as input, resembling the essence of an inverse model \cite{conant1970every}. The loss is a unscaled categorical cross-entropy, like that of the termination prediction. This additional signal is shown in experiments to produce better OOD results, especially when the bottleneck is small, as visualized in Figure \ref{fig:predact}.

\subsection{Potential of WM Baseline}
\begin{figure*}[htbp]
\centering

\subfloat[OOD $0.25$]{
\captionsetup{justification = centering}
\includegraphics[width=0.242\textwidth]{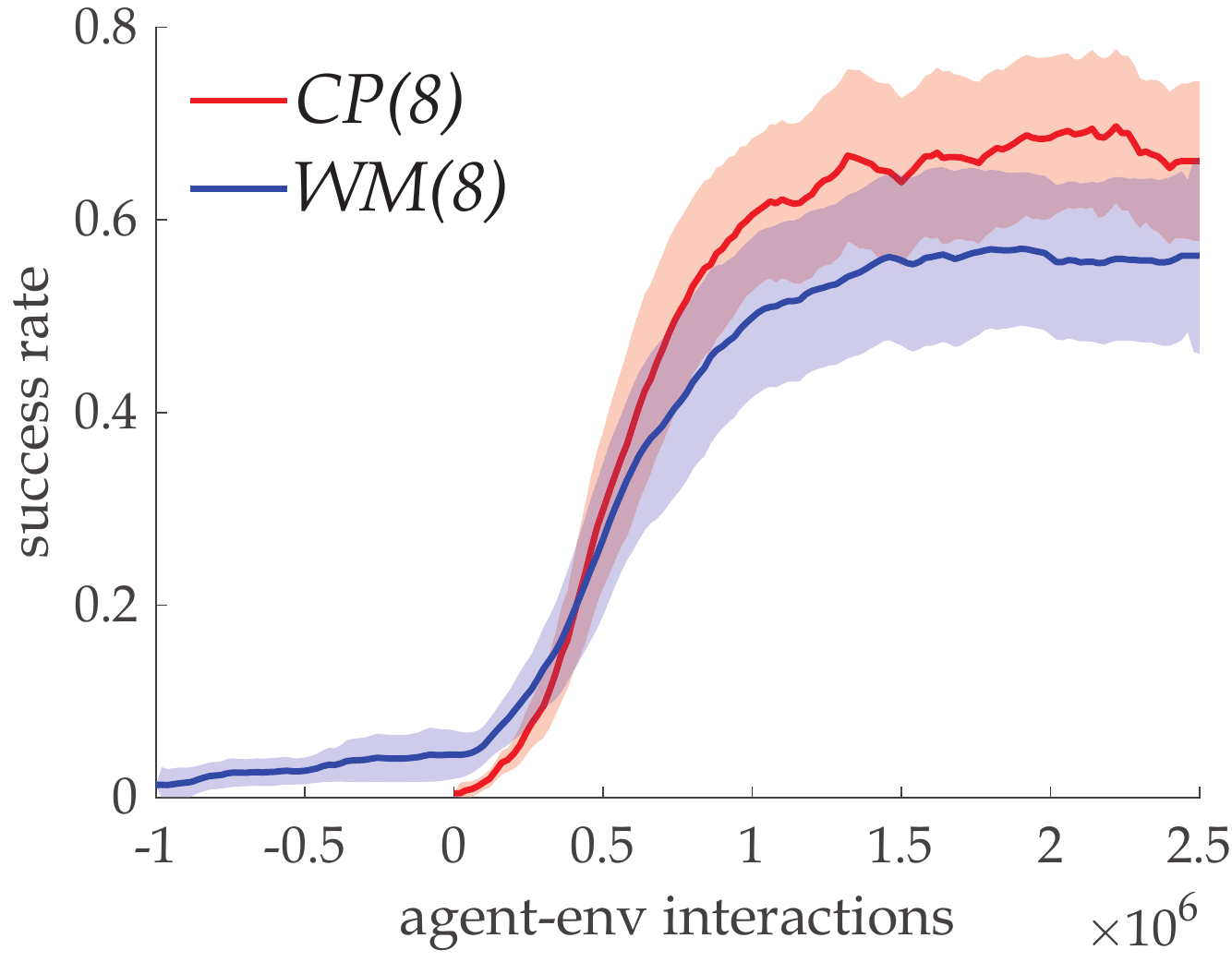}}
\hfill
\subfloat[OOD $0.35$]{
\captionsetup{justification = centering}
\includegraphics[width=0.242\textwidth]{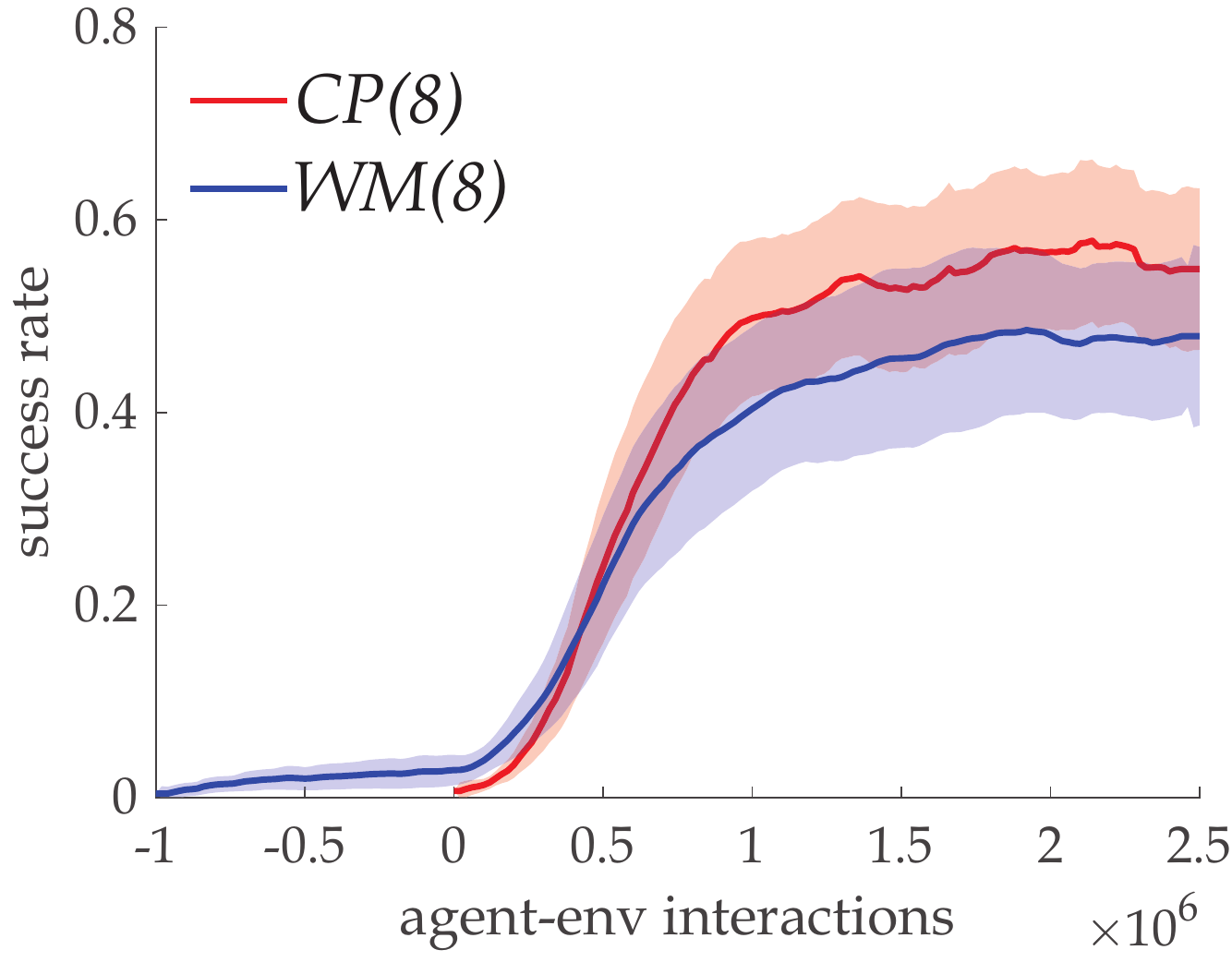}}
\hfill
\subfloat[OOD $0.45$]{
\captionsetup{justification = centering}
\includegraphics[width=0.242\textwidth]{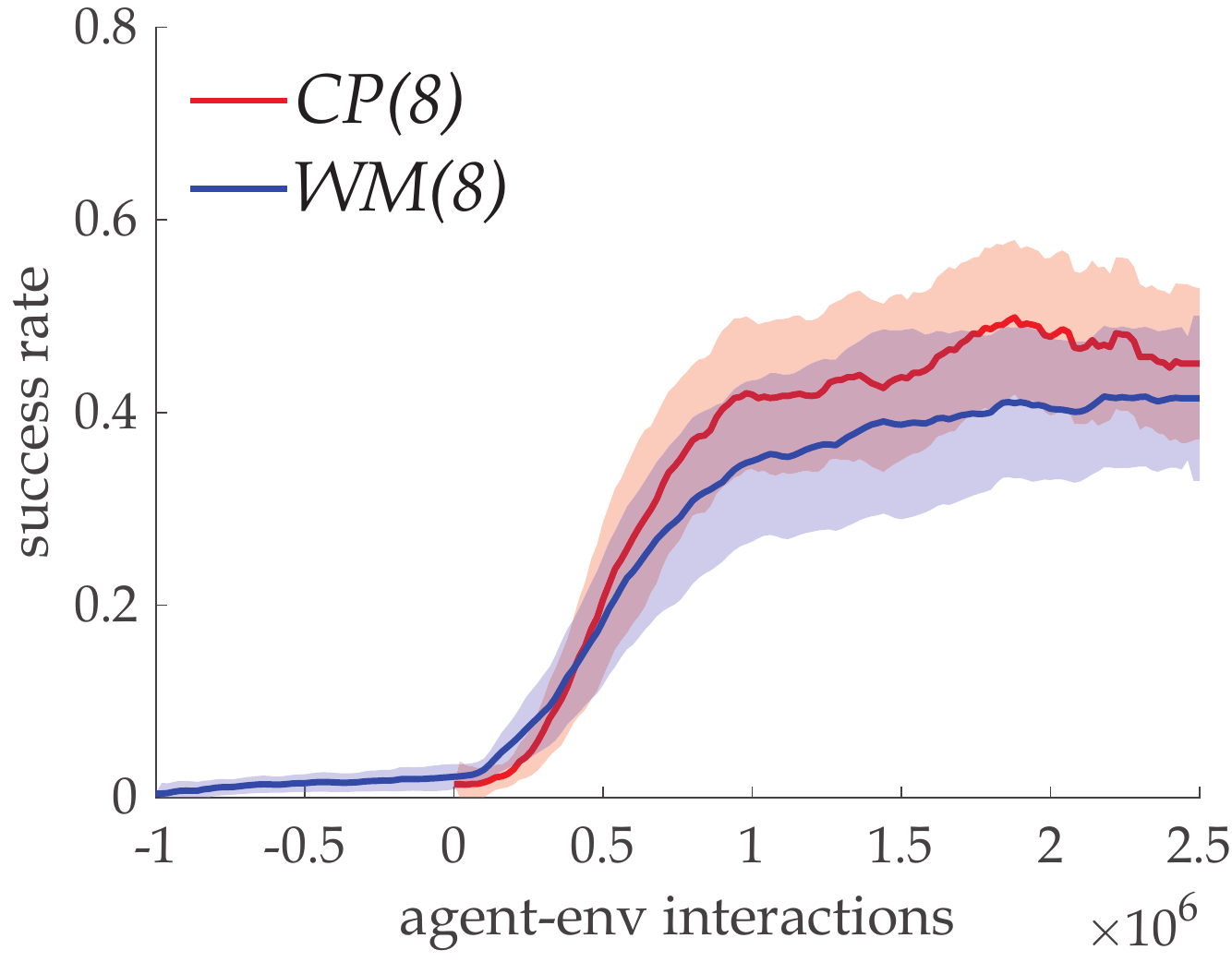}}
\hfill
\subfloat[OOD $0.55$]{
\captionsetup{justification = centering}
\includegraphics[width=0.242\textwidth]{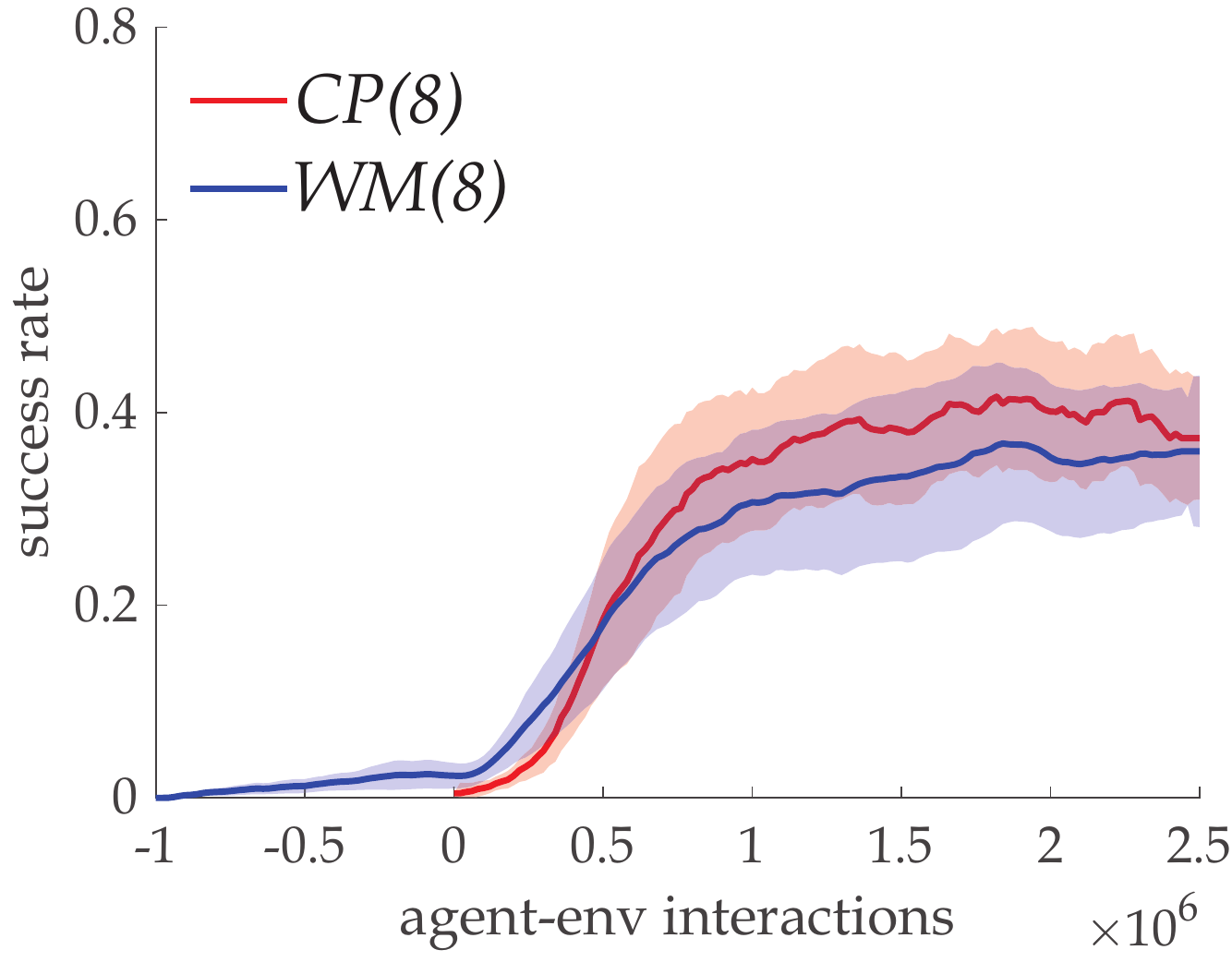}}

\caption{\small \textbf{OOD performance comparing CP and WM under a spectrum of difficulty.} WM(8) is the WM baseline which uses the same architecture as CP(8), for fair comparison. The WM(8) results are shifted for a free unsupervised world model learning phase of $10^6$ steps. All error bars are obtained from $20$ independent runs.}
\label{fig:free_unsupervised}
\end{figure*}

In case the readers are curious about how the WM baseline would evolve after the $2.5\times 10^{6}$ steps cutoff, we provide an additional set of experiments featuring a free unsupervised learning phase of $10^{6}$ agent-environment interactions. As illustrated in Figure \ref{fig:free_unsupervised}, observations suggest that WM baseline could not achieve similar performance as that of CP due to that the representation is not jointly shaped for value estimation. The results show promise of the methodology of representation learning with joint signals. However, this is not to say that an unsupervised learning of a world model is not beneficial in general, just limited to this case and this planning methodology.

\subsection{Tests on Different World Sizes}
\begin{figure*}[htbp]
\centering

\subfloat[OOD $0.25$]{
\captionsetup{justification = centering}
\includegraphics[width=0.242\textwidth]{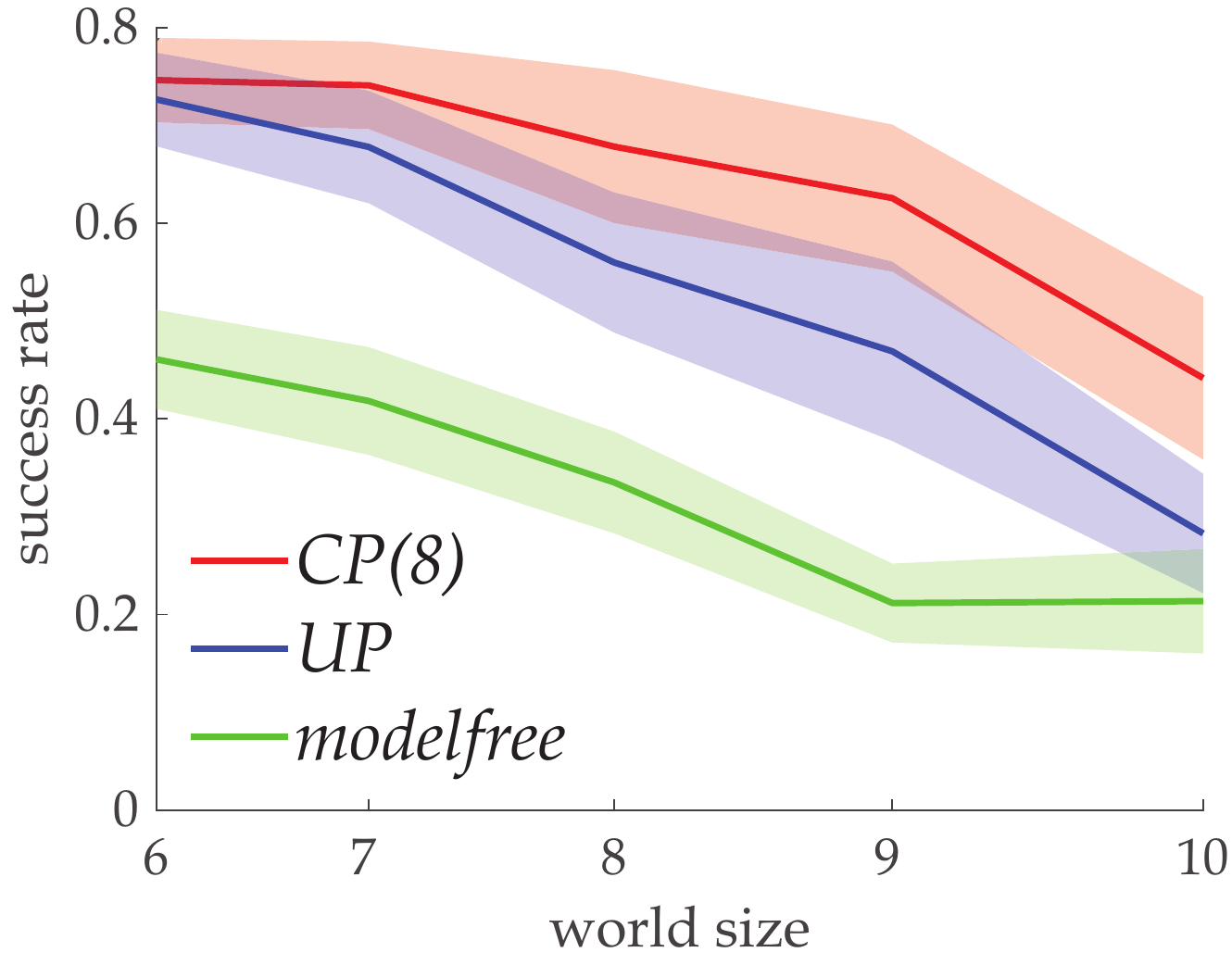}}
\hfill
\subfloat[OOD $0.35$]{
\captionsetup{justification = centering}
\includegraphics[width=0.242\textwidth]{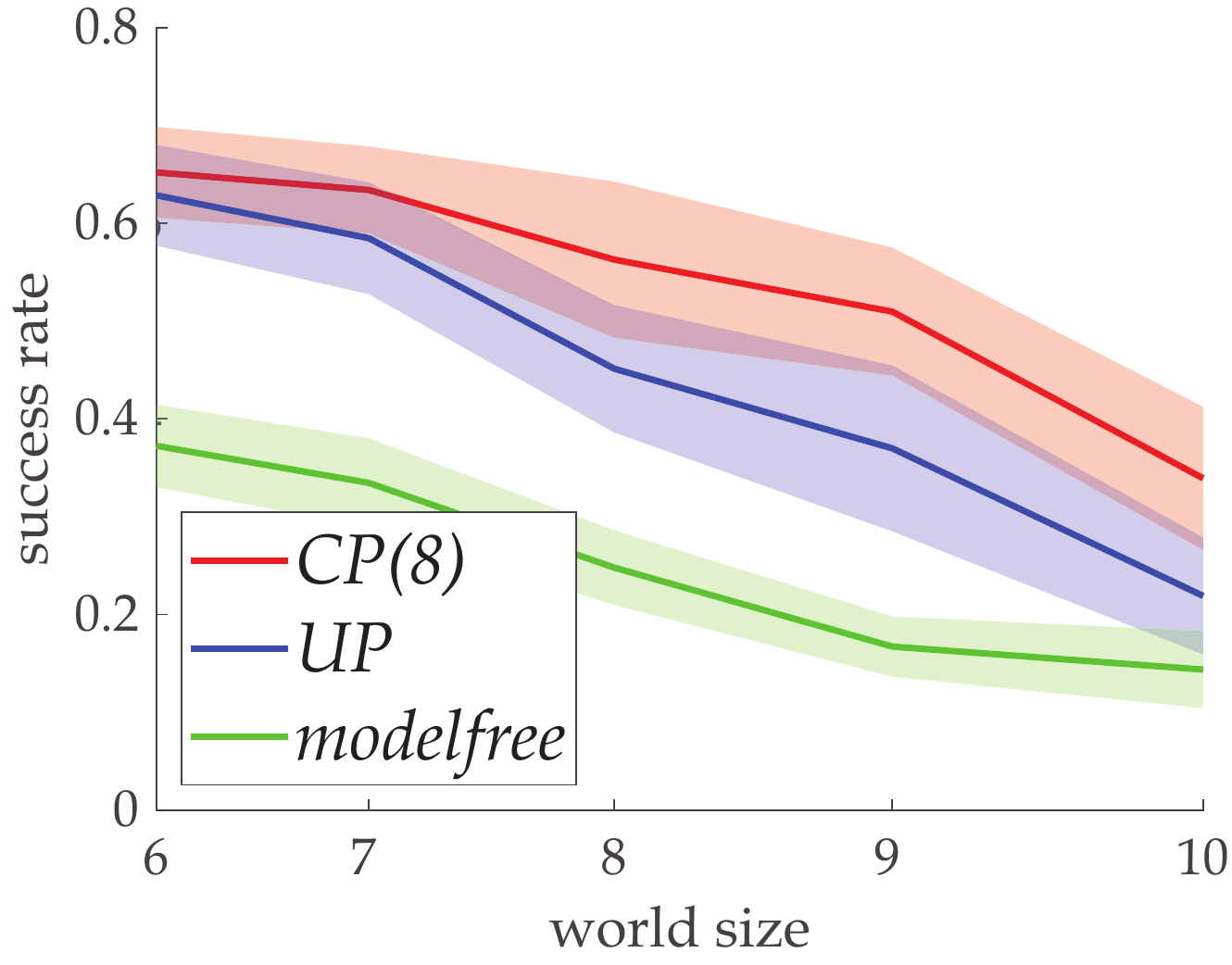}}
\hfill
\subfloat[OOD $0.45$]{
\captionsetup{justification = centering}
\includegraphics[width=0.242\textwidth]{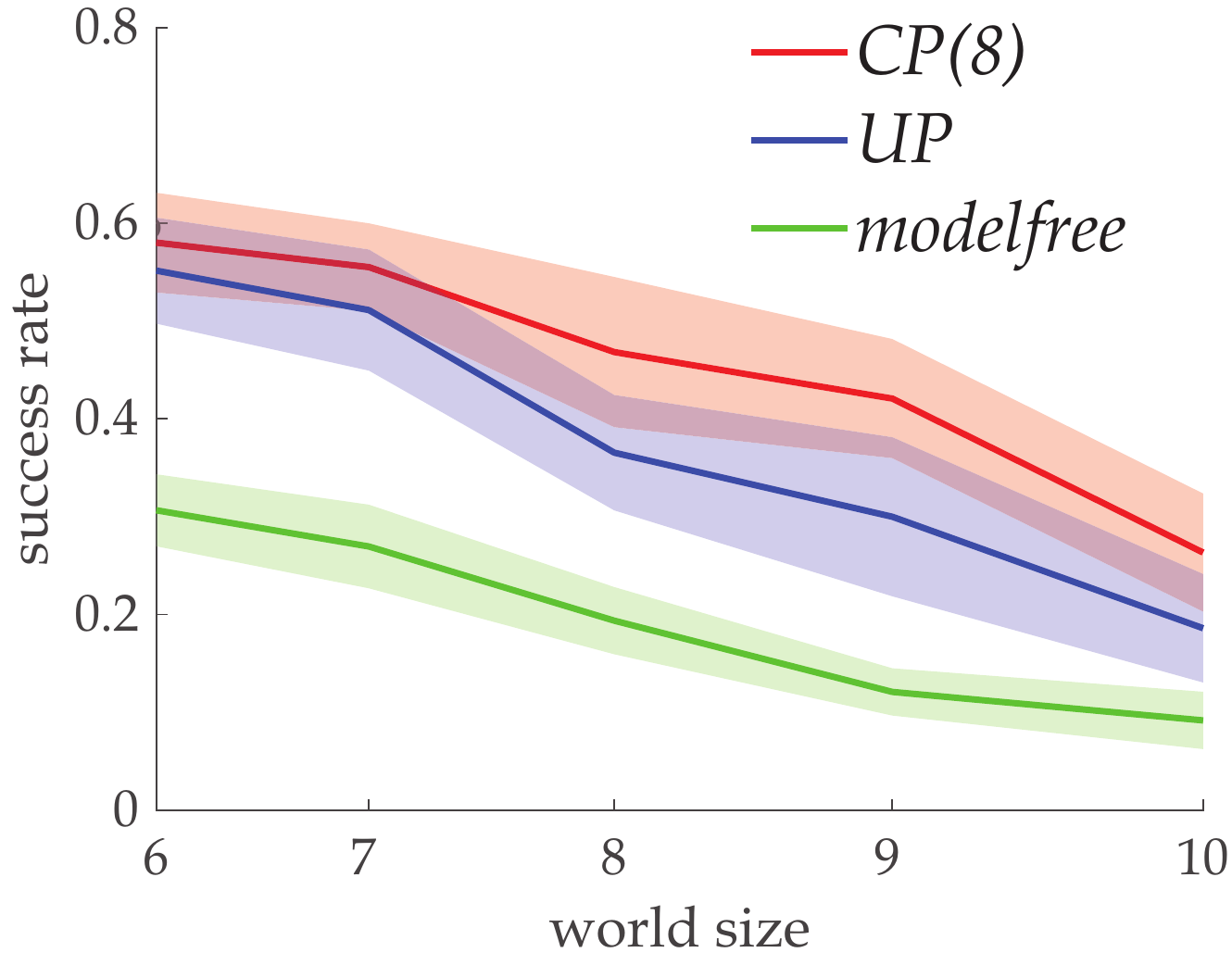}}
\hfill
\subfloat[OOD $0.55$]{
\captionsetup{justification = centering}
\includegraphics[width=0.242\textwidth]{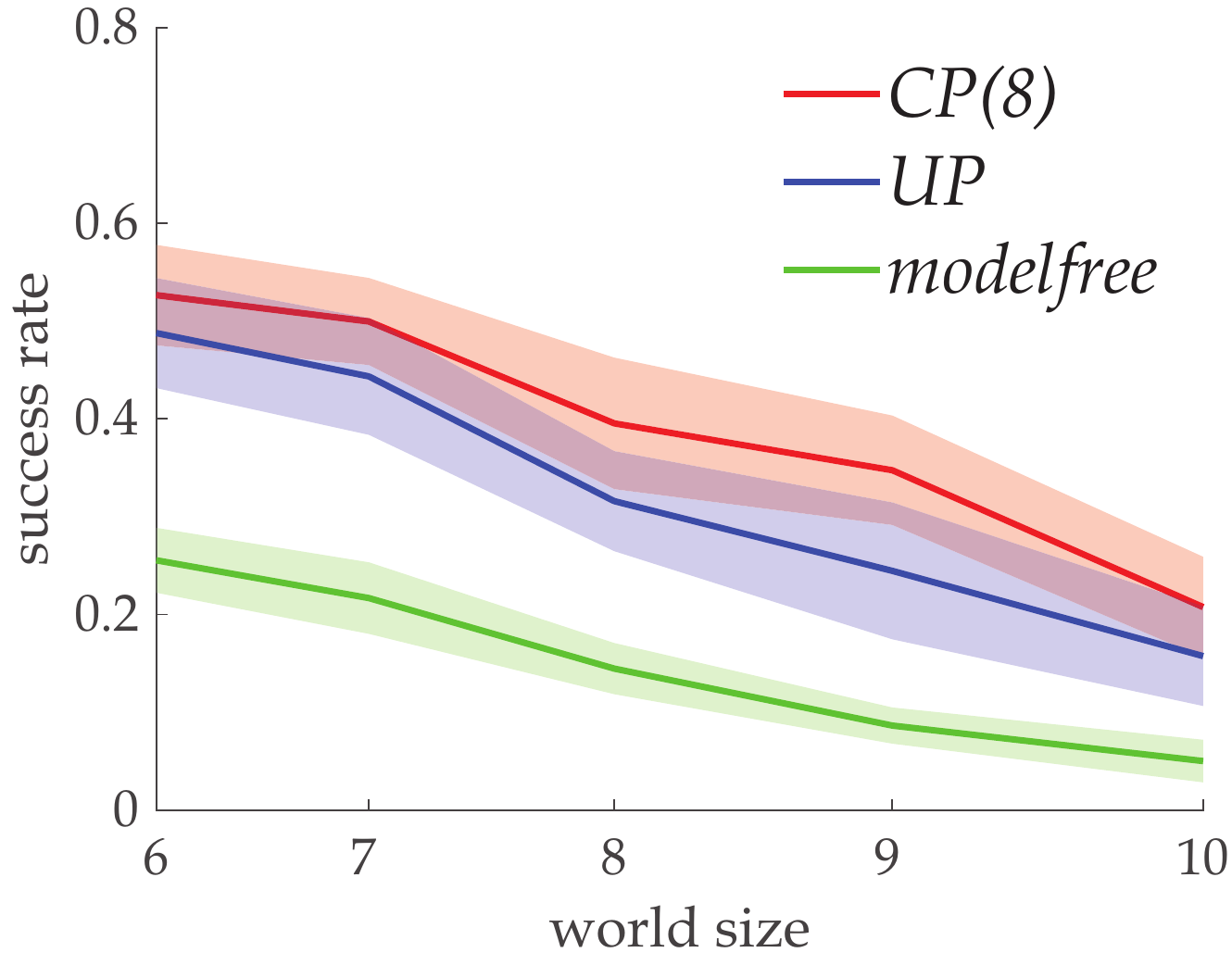}}

\caption{\small \textbf{OOD performance under a spectrum of difficulty and world sizes.} The $x$-axes are ticked with \#grids in each gridworld size, representing the number of entities for in the state set, thus non-uniform. The smaller the world sizes, the better and the closer the performance of the three methods are. The fact that the CP(8) performance deteriorates slower than UP suggests that the bottleneck may contribute to more scalable performance in tasks with larger amount of entities. All error bars are obtained from $20$ independent runs.}
\label{fig:comparison_worldsizes}
\end{figure*}

To inspect the scalability of the proposed method, we compare the methods CP(8), UP and model-free in a gradient of gridworld sizes. The results are presented in Figure \ref{fig:comparison_worldsizes}.

\subsection{Different Tasks}
To test the applicability of CP on more scenarios, we craft some additional sets of experiments with MiniGrid to test the robustness. For these extra sets of experiments, we prioritize on presenting the comparison of the CP, UP and modelfree agents' performance.

\subsection{Alternative Dynamics}
First, we want to see if the experimental conclusions would still hold on a task with different action dynamics. Thus, we modify the original task in the main manuscript by a new set of Turn-And-Forward dynamics: the action space is re-resigned to include $4$ composite actions which first turns to some directions (forward, left, right or back based on the current facing direction) and then move forward if possible (if not stepping out of the world).

\begin{figure*}[htbp]
\centering

\subfloat[OOD $0.25$]{
\captionsetup{justification = centering}
\includegraphics[width=0.242\textwidth]{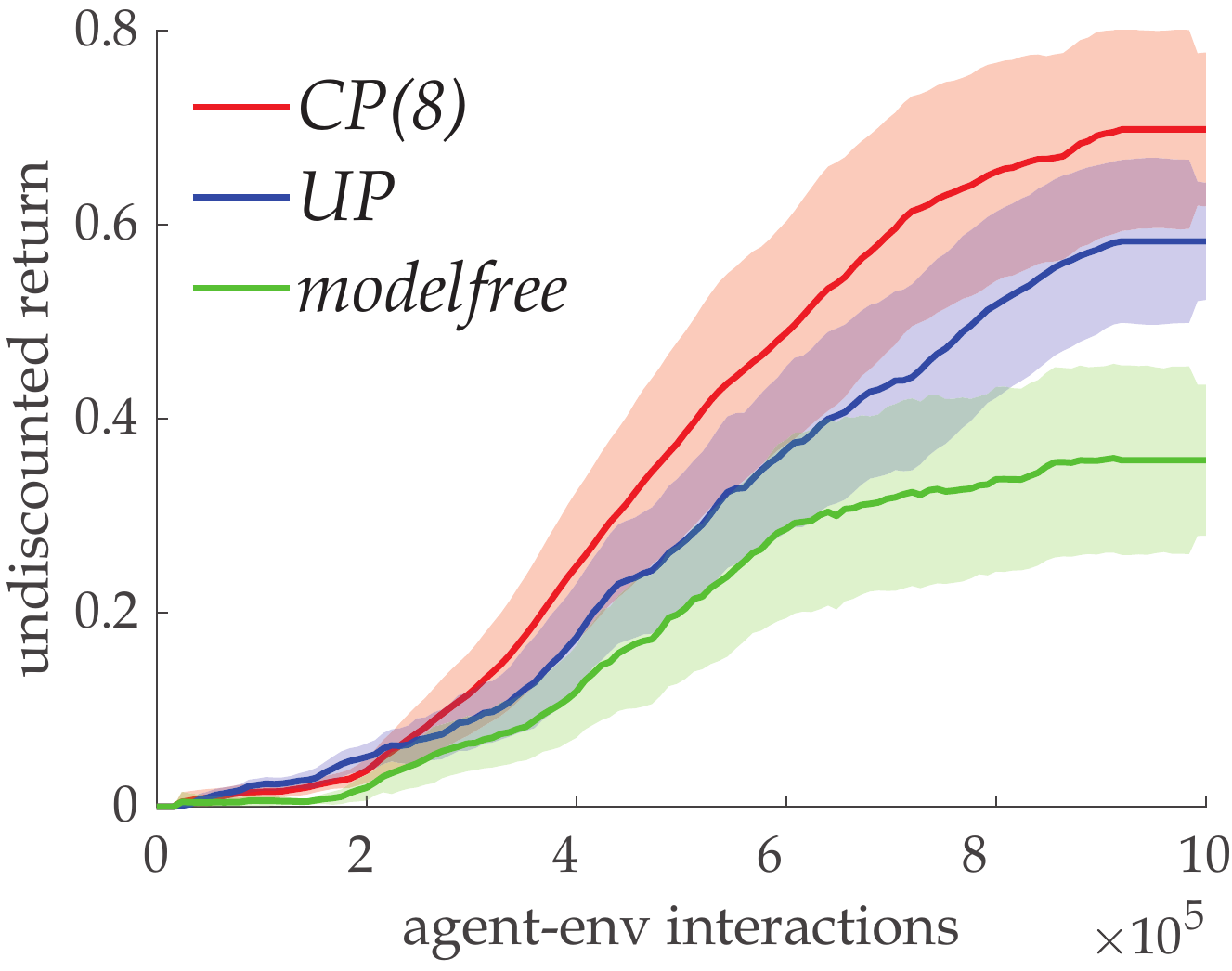}}
\hfill
\subfloat[OOD $0.35$]{
\captionsetup{justification = centering}
\includegraphics[width=0.242\textwidth]{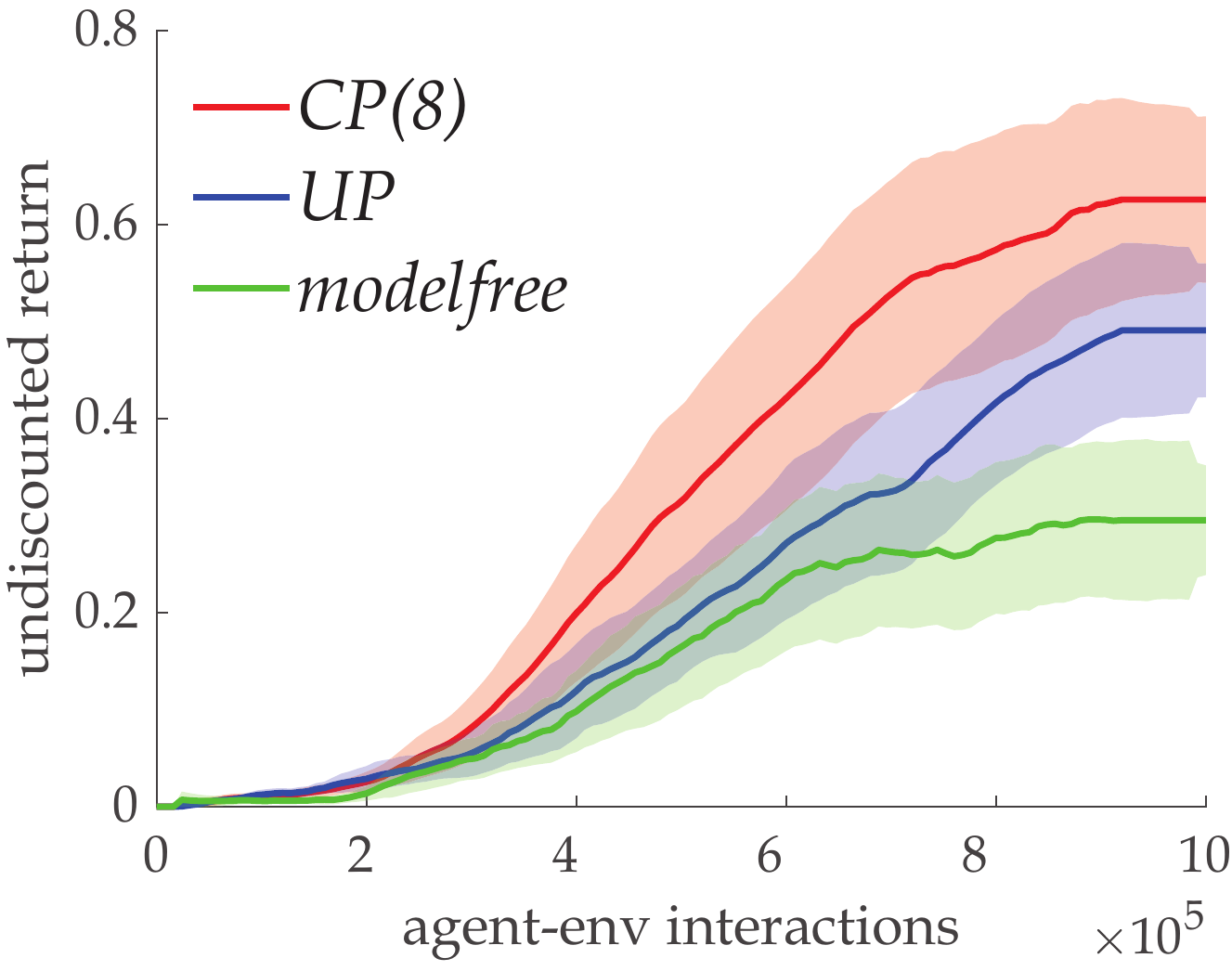}}
\hfill
\subfloat[OOD $0.45$]{
\captionsetup{justification = centering}
\includegraphics[width=0.242\textwidth]{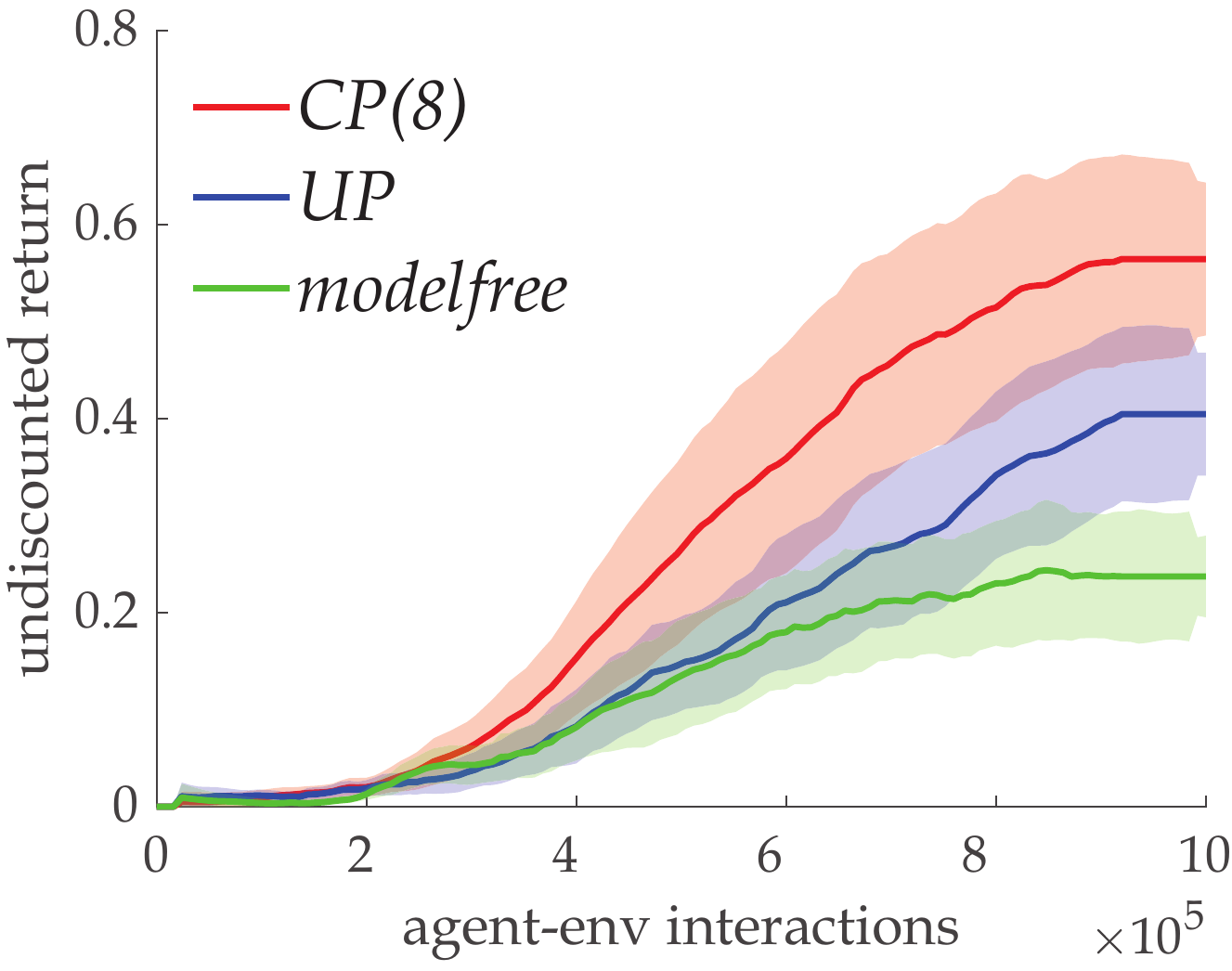}}
\hfill
\subfloat[OOD $0.55$]{
\captionsetup{justification = centering}
\includegraphics[width=0.242\textwidth]{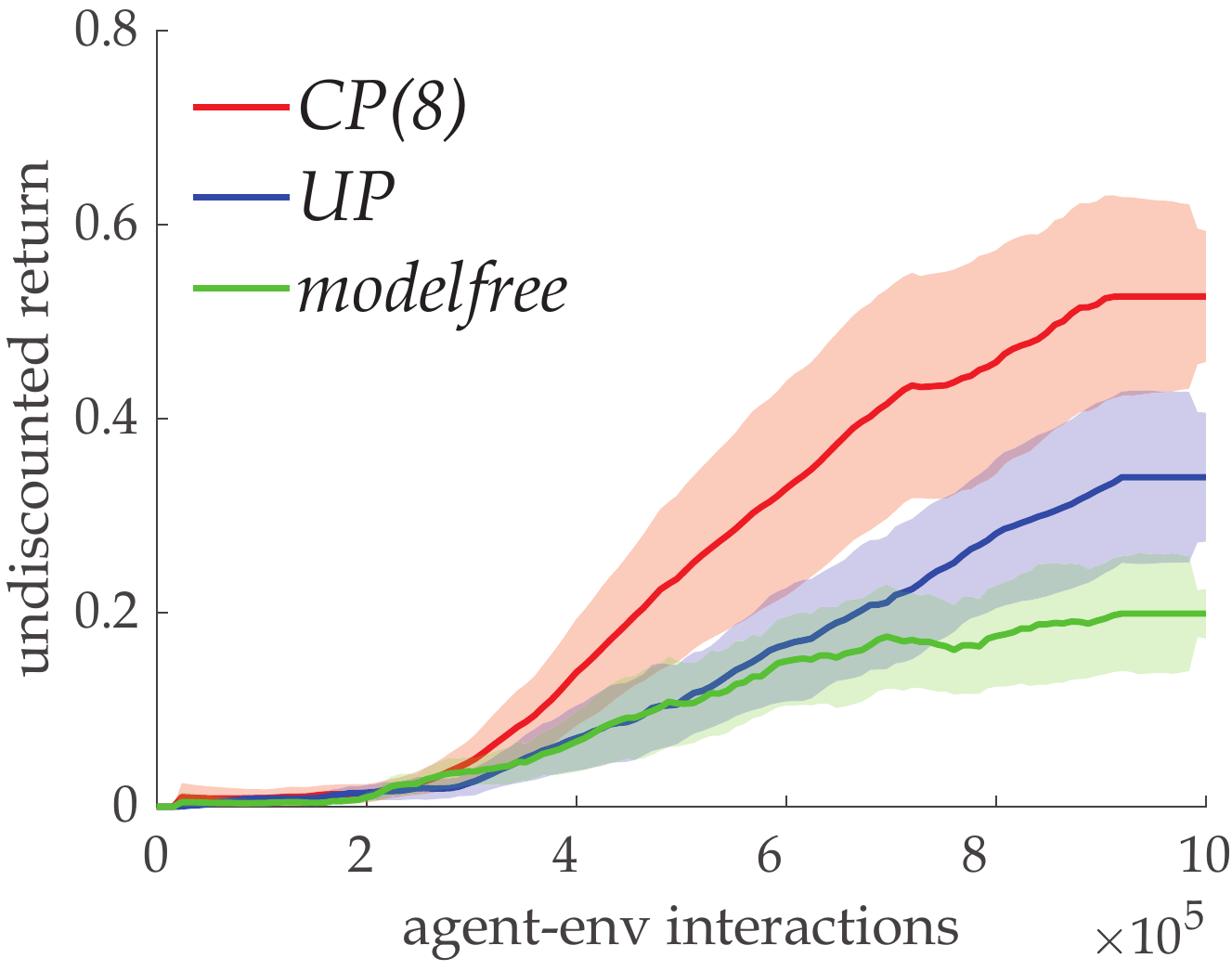}}

\caption{\small \textbf{OOD performance under a gradient of difficulty in Turn-and-Forward tasks.} All error bars are obtained from $20$ independent runs.}
\label{fig:comparison_v3}
\end{figure*}

Intuitively, this set of new dynamics can be seen as a composition of the original and hence is easier to solve and more effective in terms of planning, \ie{} the same number of planning steps would lead deeper into the future. In Figure \ref{fig:comparison_v3}, we observe that all three methods are performing better compared to the original tasks and the experimental conclusions are re-validated.

\subsubsection{Cluttering Effect}
For the second set, upon the turn-and-forward environments, we add randomly changing colors to every grid so that a cluttering effect is posed to hinder the agents from understanding the object interactions as well as disturbing the bottleneck \compression{}. Specifically, the distracting colors are sampled uniformly randomly from $6$ possibilities for each grid of each observation. We add one additional baseline named CP(8)+, which denotes a CP(8) agent with noise injection at the input of the dynamics model. Specifically, we sample an $8$-dimensional $\bm{0}$-mean identify variance Gaussian noise that is replicated and concatenated to every bottleneck object.

\begin{figure*}[htbp]
\centering

\subfloat[OOD $0.25$]{
\captionsetup{justification = centering}
\includegraphics[width=0.242\textwidth]{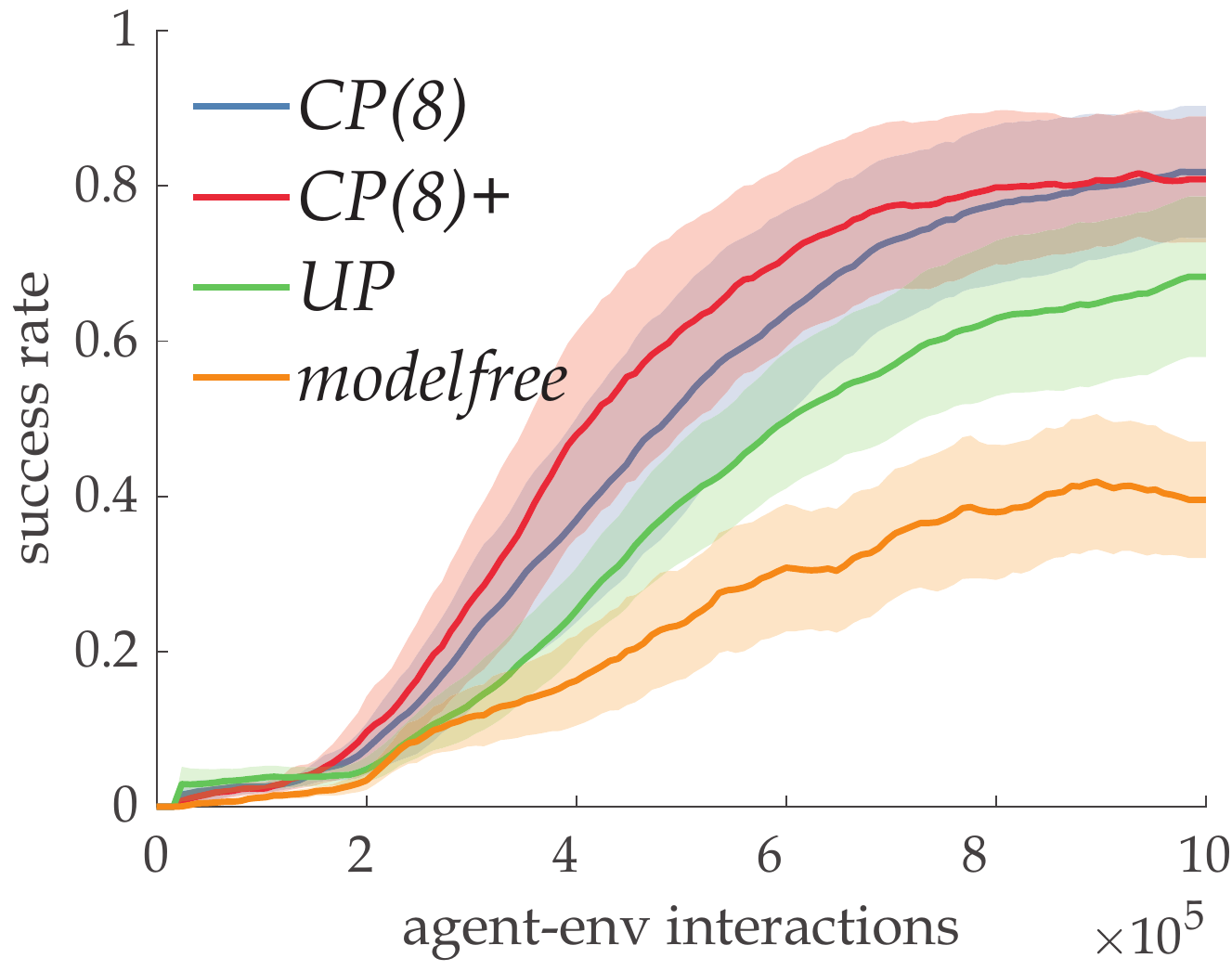}}
\hfill
\subfloat[OOD $0.35$]{
\captionsetup{justification = centering}
\includegraphics[width=0.242\textwidth]{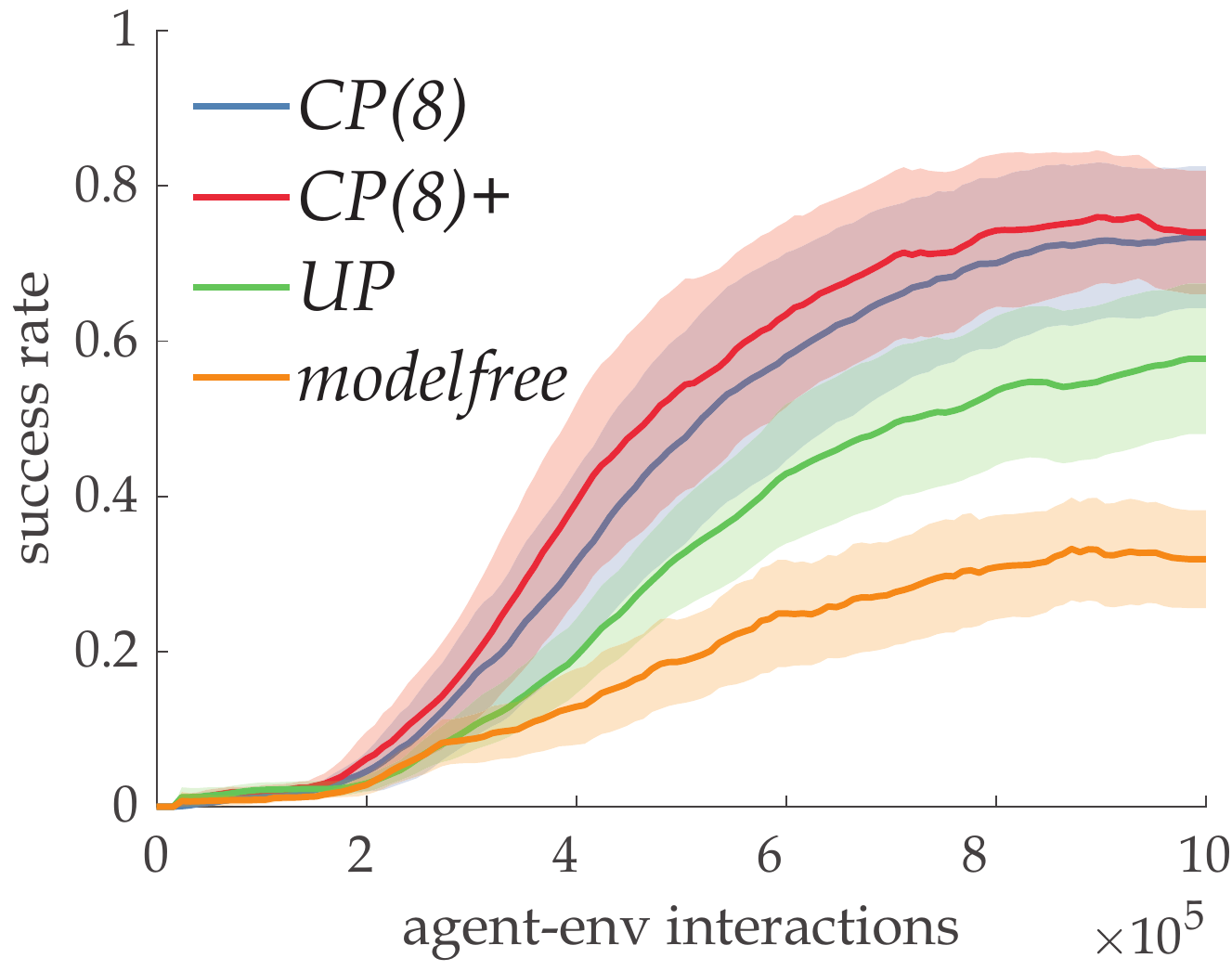}}
\hfill
\subfloat[OOD $0.45$]{
\captionsetup{justification = centering}
\includegraphics[width=0.242\textwidth]{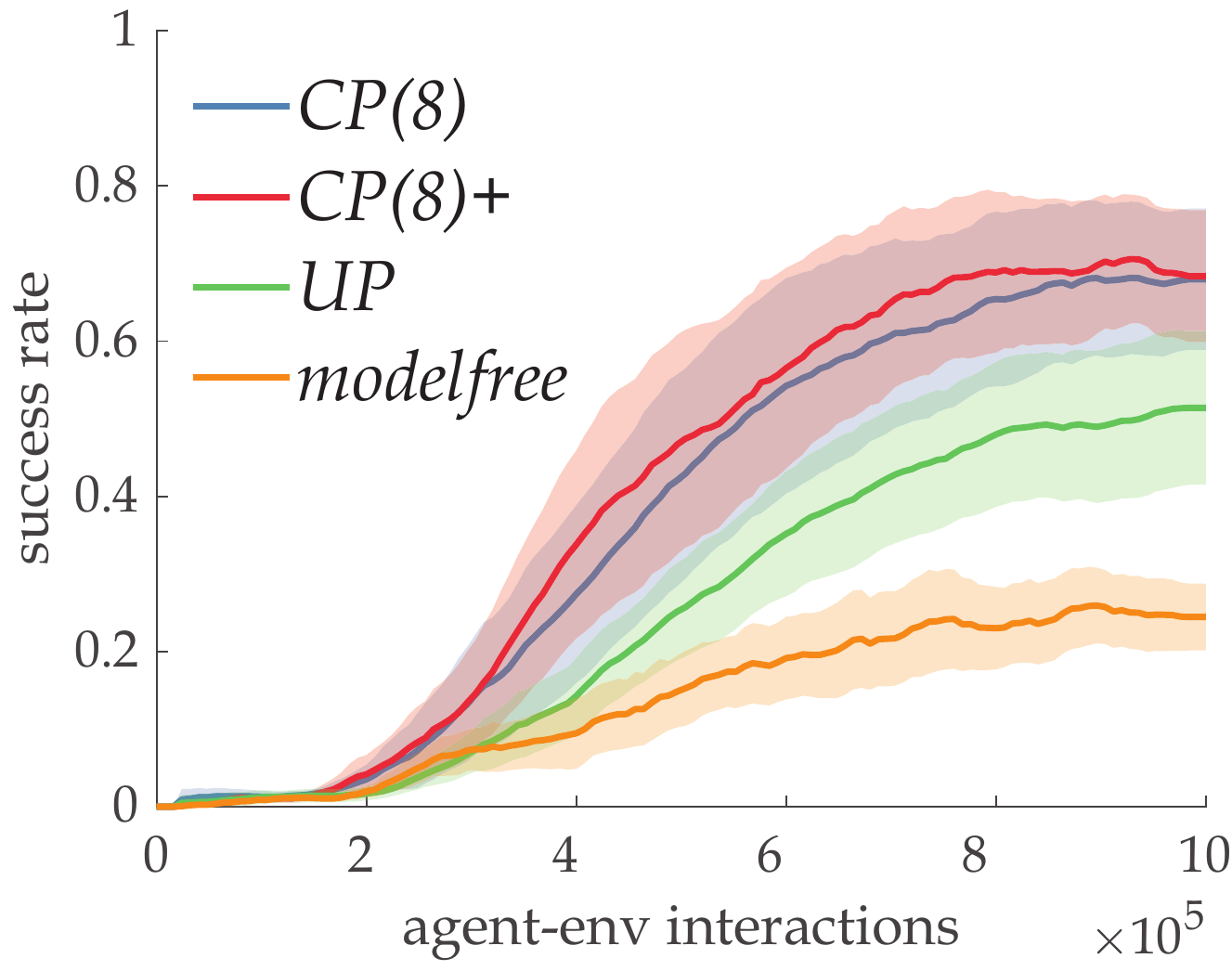}}
\hfill
\subfloat[OOD $0.55$]{
\captionsetup{justification = centering}
\includegraphics[width=0.242\textwidth]{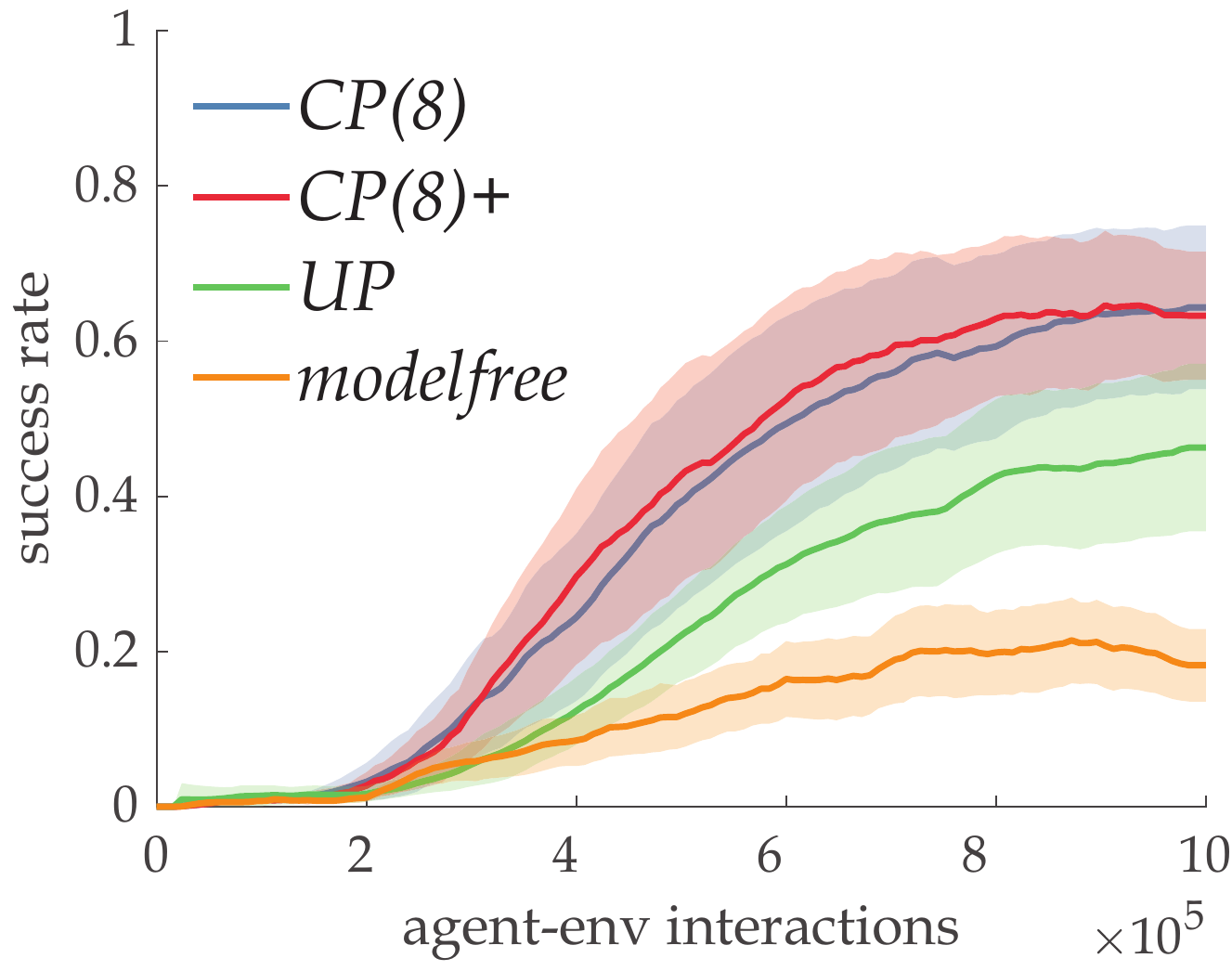}}

\caption{\small \textbf{OOD performance under a gradient of difficulty in Turn-and-Forward tasks with color distractions.} All error bars are obtained from $20$ independent runs.}
\label{fig:comparison_distractv3}
\end{figure*}

From Figure \ref{fig:comparison_distractv3}, we observe that CP(8) still performs better than UP therefore re-validating the effectiveness of the bottleneck mechanism. It is likely that our state set encoder to learn to ignore the distractions and thus make the bottleneck selector to be able to direct attention to the relevant objects as what we have done for the tasks with the old dynamics. Furthermore, CP(8)+ seems to achieve similar performance as CP(8) but converges faster. This experimental observation suggests that noisy inputs may be beneficial to the learning behavior of the dynamics model for a noisy environment. Yet across the paper, we have tried not to use any other additional means to enhance the performance of our agents in order to isolate impact of unwanted components.

\subsubsection{Key-Chest Unlocking}
We built the final set of experiments built upon the logic of MiniGrid’s MiniGrid-Unlock-v0 and Turn-and-Forward dynamics. The agent needs to navigate the gridworld while avoiding obstacles (0 reward, end of episode) to get a key first (+0.5 reward) and then unlock a chest (+0.5 reward, end of episode, considered as success) to finish the task.

\begin{figure*}[htbp]
\centering

\subfloat[OOD $0.25$]{
\captionsetup{justification = centering}
\includegraphics[width=0.242\textwidth]{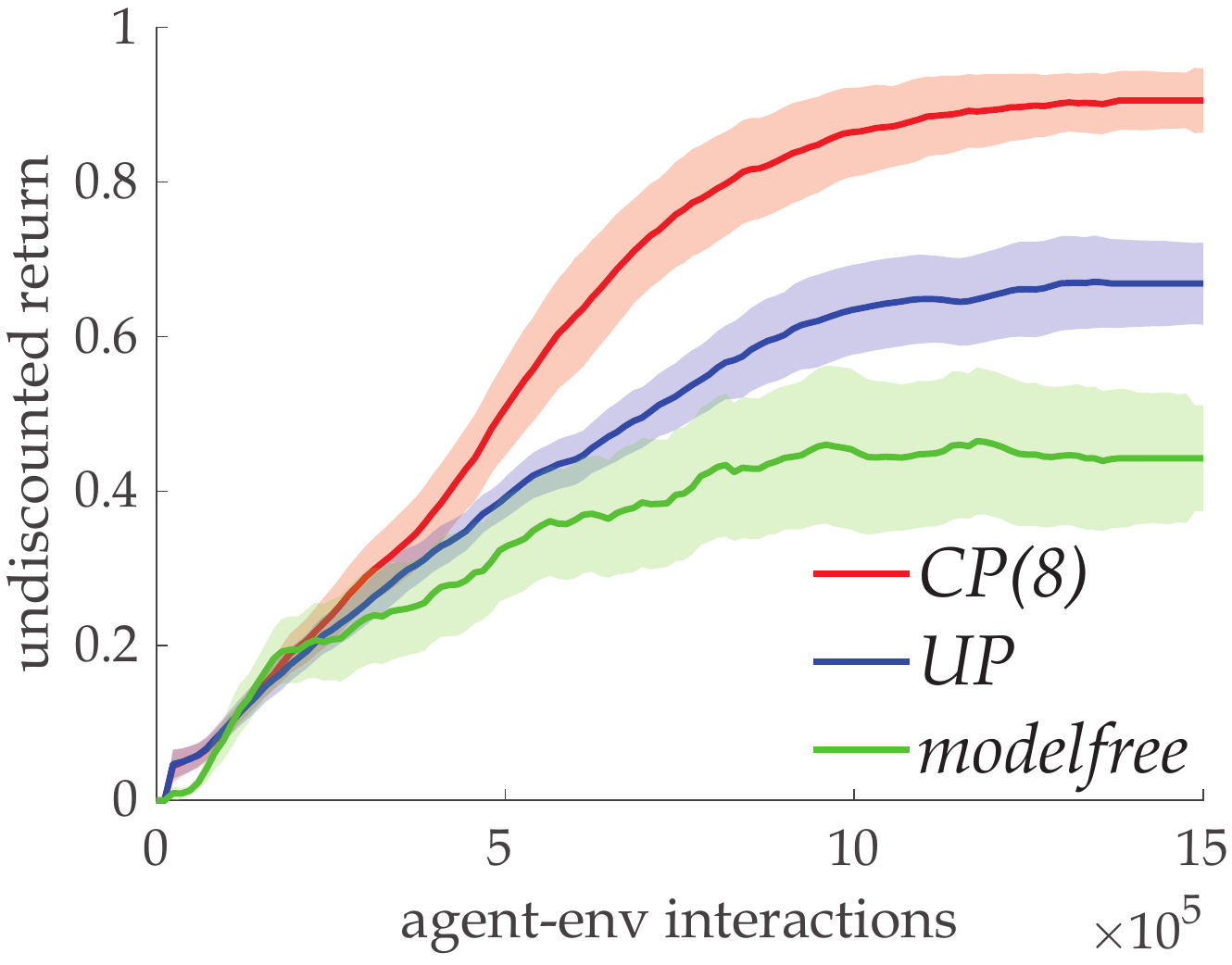}}
\hfill
\subfloat[OOD $0.35$]{
\captionsetup{justification = centering}
\includegraphics[width=0.242\textwidth]{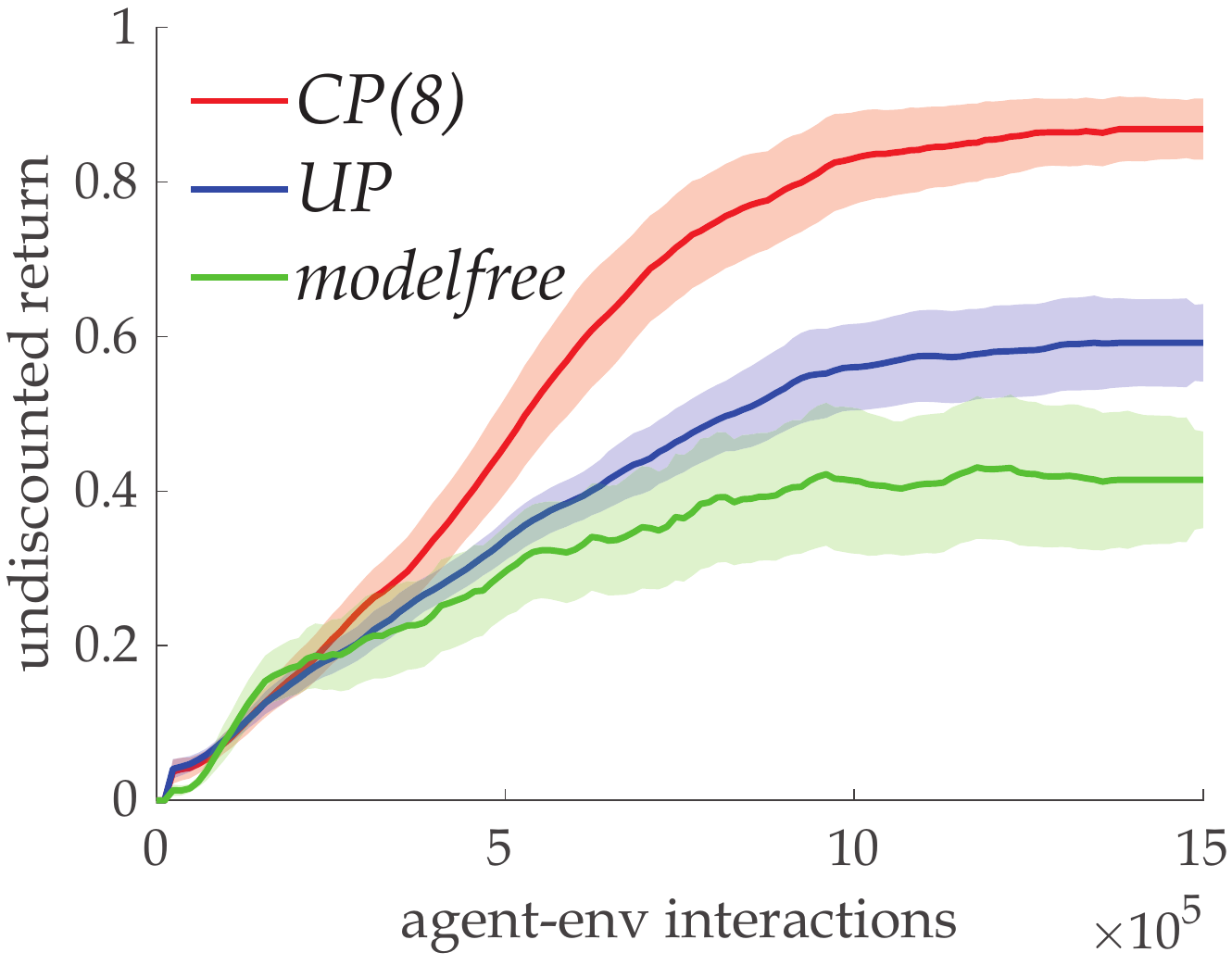}}
\hfill
\subfloat[OOD $0.45$]{
\captionsetup{justification = centering}
\includegraphics[width=0.242\textwidth]{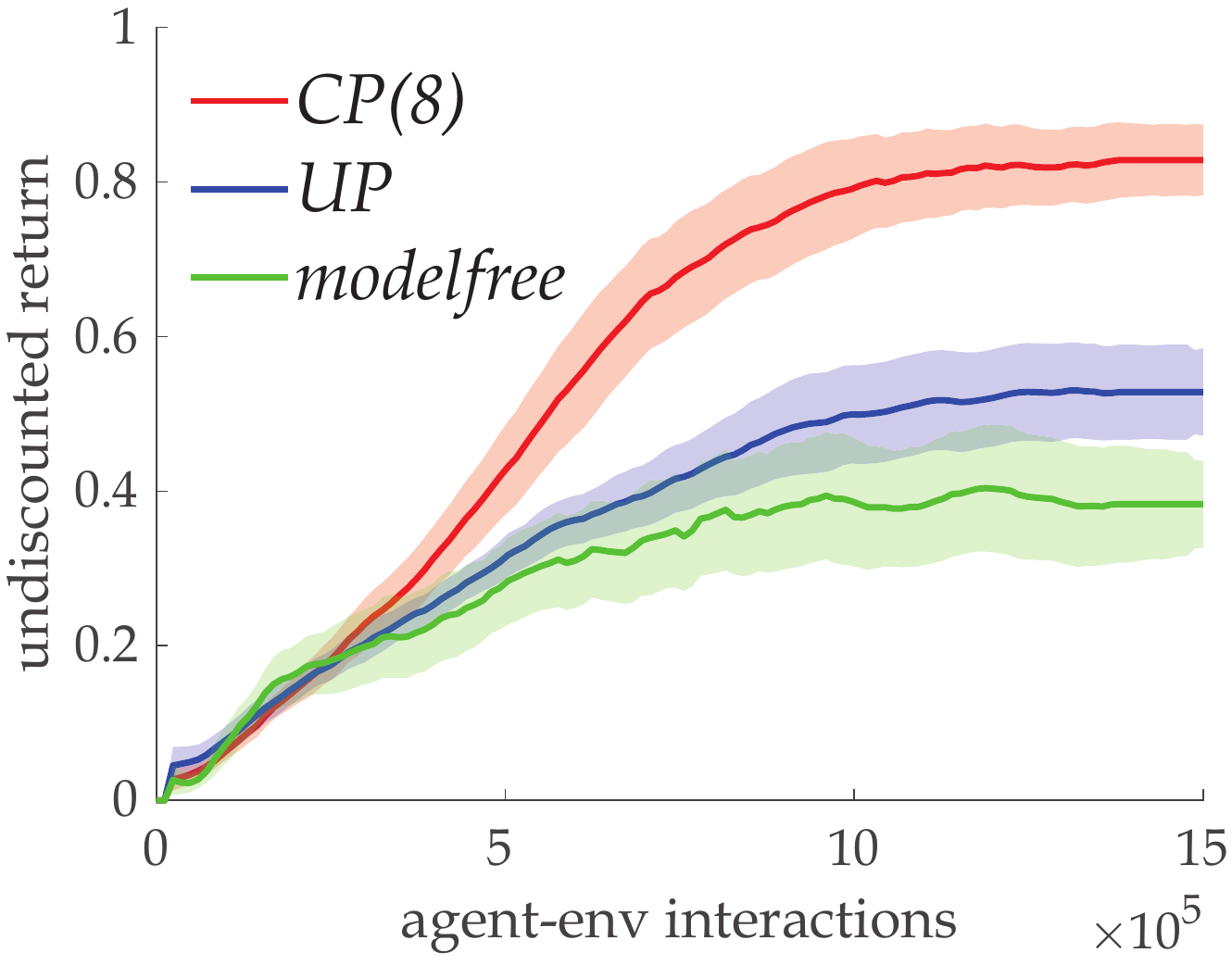}}
\hfill
\subfloat[OOD $0.55$]{
\captionsetup{justification = centering}
\includegraphics[width=0.242\textwidth]{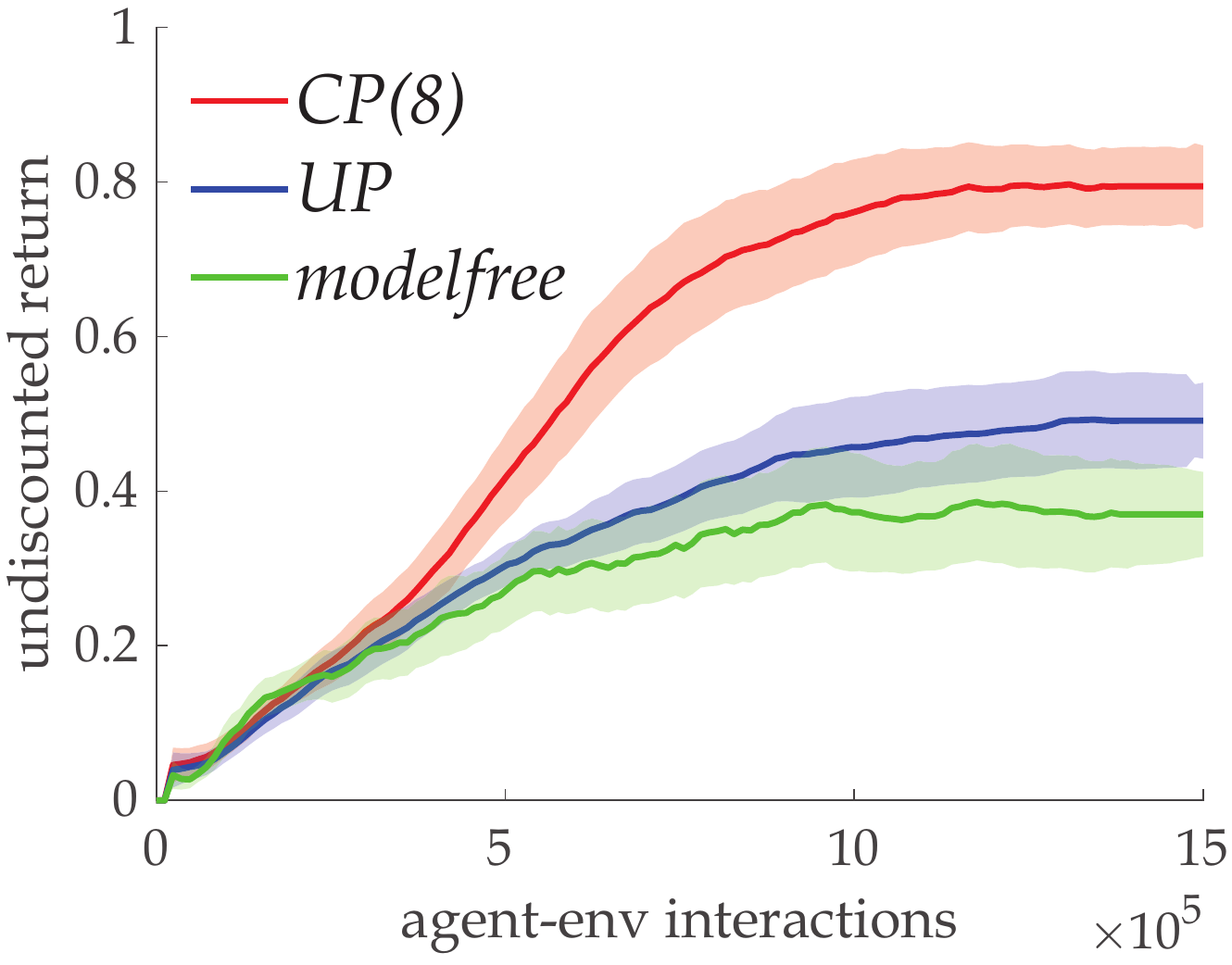}}

\caption{\small \textbf{OOD performance under a gradient of difficulty in Key-Chest Unlock Task.} The $y$-axes values are undiscounted cumulative episodic return. All error bars are obtained from $20$ independent runs.}
\label{fig:comparison_kdv3}
\end{figure*}

In this task in particular, the bottleneck is tested against a conditional selection: the reward and termination feedbacks on the chest is conditioned on if the key is still not acquired. Hence, in the bottleneck, when the agent plans to step into the chest, the bottleneck should additionally select the key to predict well the outcome. From Figure \ref{fig:comparison_kdv3}, the results of the three compared methods are more differentiable than the original turn-or-forward setting and the pattern remains consistent across all settings, the bottleneck-equipped CP agent performs better than UP and modelfree.

\subsection{Learning Capable Representation with Non-Conflicting Joint Signals}
We have gathered more empirical evidence regarding the non-conflicting training of the state-representation based on the signals. In terms of model learning accuracy, according to our results related to WM baseline, removing the value estimation signal would result in poorer representation but not lower accuracy when predicting other relevant signals; Removing termination signal would not impact the convergence of reward prediction accuracy or that of the state prediction however the RL performance is hindered. Removing the reward signal or the next state prediction signal leads to total collapse of the tree-search based behavior policy however the convergence of the remaining model training signals is not affected much. With these, we would like to suggest that we have, at least in this task setting, learned a set-based representation capable of predicting all interesting quantities. 

\section{Visualization of Selection}
\begin{figure*}[htbp]

\subfloat[Turn Left]{
\captionsetup{justification = centering}
\includegraphics[width=0.32\textwidth]{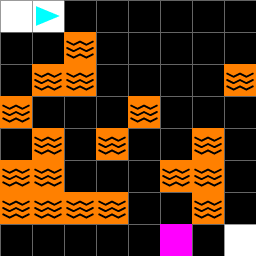}}
\hfill
\subfloat[Turn Right]{
\captionsetup{justification = centering}
\includegraphics[width=0.32\textwidth]{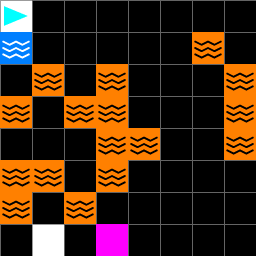}}
\hfill
\subfloat[Step Forward]{
\captionsetup{justification = centering}
\includegraphics[width=0.32\textwidth]{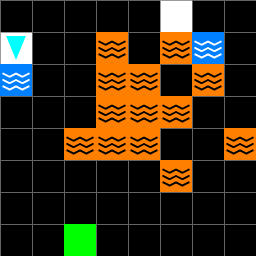}}

\caption{\small Visualization of the bottleneck selection given the observation and specific actions. These figures are extracted from a fully trained CP(2) agent under OOD evaluation. The bottleneck is set to very small for clearer visualization purposes. We invert the color of the selected objects by the best performing head, \ie{} the head that covers the most relevant objects, though the selection quality would be sufficiently justified if \textit{all} the heads could \textit{cumulatively} cover all the interested objects. The grids of the selection would be at least $2$ but at most $4$ due to the design. }
\label{fig:visualize_att}
\end{figure*}

We present some visualization of the object selection during the planning steps in Figure \ref{fig:visualize_att}. In (a), with the intention of turning left, the agent takes into the bottleneck the location of itself within the grid (visualized as the teal triangle with white surroundings, color-inverted from red-black); For (b), the agent additionally pays attention to the lava grid on its right while trying to turn right. In (a) and (b), the goal square (pink, color-inverted from green) is also paid attention but we cannot interpret such behavior. Finally in (c), we can see that the agent takes consideration into the grid (the blue lava grid, color-inverted from orange) that it is facing before taking a step-forward action. Though these visualization provides an intuitive understanding to the agents' behavior, they do not serve statistical purposes.

We additionally have collected the coverage ratio of all the relevant objects by the selection phase in all the in-distribution and OOD evaluation cases along the process of learning. The collected data on bottleneck sizes $4$, $8$ and $16$ indicate that the coverage is almost perfect very early on during training. We do not provide these curves because the convergence to $100\%$ is so fast that the curves would all coincide with line $y=1$, with some minor fluctuations of the standard deviation shades.

\section{Details of Compared Baseline Methods}
\subsection{Staged Training (World Models)}
The agents with World Model (WM) trained in stages share the same architectures as their CP or UP counterparts. The main difference is that the WM agents adopt a $2$-staged training strategy: In the first $10^{6}$ agent-environment interactions, only the model is trained and therefore the representation is only shaped by the model learning. In the first stage the agent relies on a uniformly random policy. After $10^{6}$ interactions, the agent freezes its encoder as well as the model to carry out value estimator learning. Note that the agent carries out tree-search MPC with the frozen model in the second stage. Compared to CP or UP, the exploration scheme is delayed but unchanged. Also, the training configurations do not change.

\subsection{Dyna}
The Dyna agents share the model-free part of the architecture as CP or UP. The models that Dyna baselines learn are powered by our action-conditioned set-to-set architecture on an observation-level. The training timings for both the model and the value estimator are not changed, though they do not jointly shape the representation and are used very differently compared to CP or UP. In our implementation, the generation of imagined transitions is carried out by dedicated processes. These processes generate transitions and send them to a dedicated global replay buffer of size $1024$. The small size is to ensure that the delusional transitions would be washed out soon after the model is effective. The TD learning of the value estimator samples a double-sized mini-batch, half from the buffer of real transitions and half imagined. While the model training uses only the true transitions, with unit-sized batches. Since our model is not generative, we rely on free-of-budget model-free agents to collect true $\langle s, a \rangle$ pairs from the environment and then complete the missing parts of the transitions (reward, termination and next observation) using the model (for the Dyna baseline with true dynamics, we just collect the whole transition exclude the model). This way, the transitions would follow the state-action occupancy jointly defined by the MDP dynamics and the policy. The approach is a compromise to implement a correctly performing Dyna agent with a non-generative model.

\subsection{NOSET}
\begin{figure*}[htbp]
\centering
\captionsetup{justification = centering}
\includegraphics[width=0.45\textwidth]{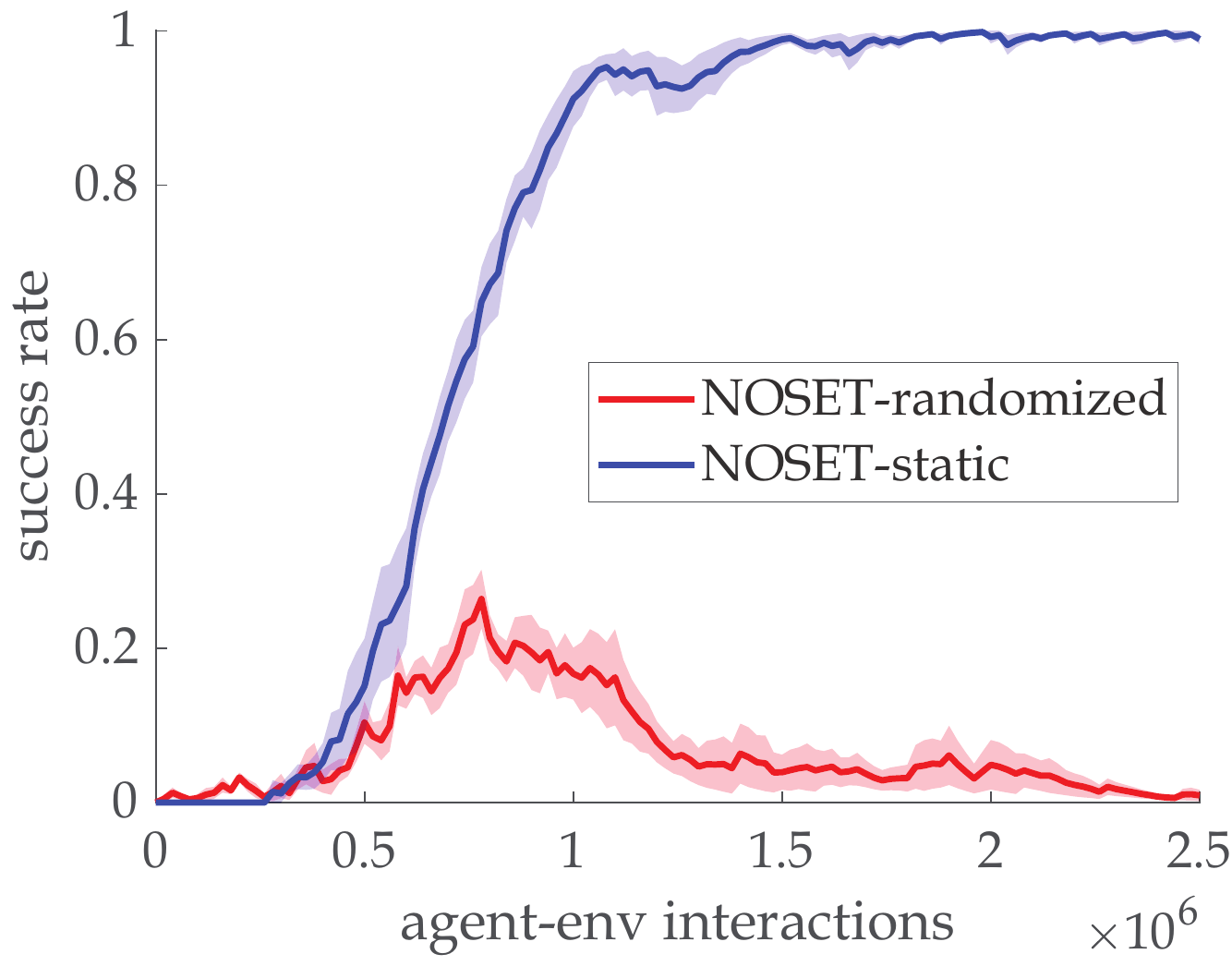}
\caption{\small NOSET baseline performance on randomized and static random environments. Each band is consisted of the mean curve and the standard deviation interval shades obtained from $20$ independent seed runs.}
\label{fig:noset_static}
\end{figure*}

The NOSET baselines embraces traditional vectorized representations. We use the same encoder but instead of transforming the feature map into a set, we flatten it and the linearly project it to some specific dimensionality ($256$). This vector would be treated as $s_t$, the same as the most existing DRL practices. Since all set-based operations would be now obsolete for the vectorized representation, they are substituted with $3$-layered FCs with hidden width $512$. The $2$-layered dynamics model employ a residual connection with the expectation that the model might learn incremental changes in the dynamics. In our experiments with randomly generated environments for each episode, the NOSET baseline performs miserably. However, if we instead randomly generate an environment at the start of the run and use the same one for the whole run, \ie{} adopt the more classical RL setting, we find that the NOSET baseline is able to perform effectively, as shown in Figure \ref{fig:noset_static}.

The dimension of the state representation, the widths and the depths of the FC layers are obtained through coarse grid tuning of the exponents of $2$. We find that architectures exceeding the chosen size are hardly superior in terms of performance.

\section{Tree Search MPC}
The agent (re-)plans at every timestep using the learned model in the hidden state level. The in-distribution planning strategy is a best-first search MPC heuristic. While the OOD planning heuristic is random search. Note that no matter which heuristic is used, the chosen action is always backtracked by the trajectory with the most return.

\begin{algorithm*}[htbp]
\caption{Prioritized Tree-Search MPC}
\label{alg:bfsearch}
\KwIn{$s_0$ (current state), $\scriptA$ (action set), $\scriptM$ (model), $\scriptQ$ (value estimator), $\gamma$ (discount)}
\KwOut{$a^*$ (action to be taken)}

$q = \text{queue}()$; $q_T = \text{queue}()$ \textcolor{darkgreen}{//$q_T$ for terminal nodes}\\

$n_u = \text{NODE}(s_0, \text{root}=\text{True})$ \textcolor{darkgreen}{//$n_u$ denotes a node with branches unprocessed nor in $q$}\\

\While{True}{
    \If{$n_u.\omega$}{
        $q_T.\text{add}(\langle n_u, n_u.\sigma \rangle)$ \textcolor{darkgreen}{//identified as a terminal state. $n_u$ is added to $q_T$ using bisection, together with the discounted sum of the simulated rewards along the way $n_u.\sigma$}
    }
    \Else{
        \lFor{$a \in \scriptA$}{
            $q.\text{add}(\langle n_u, a, n_u.\sigma +  \gamma ^{n_u.\text{depth}} \cdot Q(n_u.s,a) \rangle)$ \textcolor{darkgreen}{//bisect \wrt{} priority}
        }
    }
    \lIf{$\text{isempty}(q)$}{\textbf{break} \textcolor{darkgreen}{//tree depleted}}
    $n_c,a_c,v_e=q.\text{pop}()$ \textcolor{darkgreen}{//get branch with highest priority; for in-distribution setting, priority is the estimated value of the leaf trajectory}\\
    \lIf{budget depleted}{\textbf{break} \textcolor{darkgreen}{//termination criterion met}}
    $\hat{s}, \hat{r}, \hat{\omega} = \scriptM(n_c.s, a_c)$ \textcolor{darkgreen}{//simulate the chosen branch}\\
    $n_u = \text{NODE}(\hat{s}, \text{parent}=n_c)$\\
    \llIf{$n_c.\text{depth}>0$}{$n_u.a_b=n_c.a_b$}
    \lElse{$n_u.a_b=a_c$ \textcolor{darkgreen}{//descendants trace root action}}
    $n_u.\omega=\hat{\omega}$; $n_u.\sigma=n_c.\sigma + \gamma^{n_c.\text{depth}} \cdot \hat{t}$
}
$n_c,a_c,v_e=q.\text{pop}(\text{`highest value'})$ \textcolor{darkgreen}{//get branch with highest \textbf{value} within the expandables}\\


$n^* = n_c$;\\

\If{$\neg\text{isempty}(q_T)$}{

$n_T = q_T.\text{pop}(\text{`highest value'})$ \textcolor{darkgreen}{//get node with highest \textbf{value} within simulated terminal states}\\

\llIf{$n_T.\text{value} \geq v_e \vee \text{isempty}(q)$}{$n^* = n_T$}
} 

\llIf{$\text{isroot}(n^*)$}{$a^*=a_c$}
\lElse{$a^*=n^*.a_b$} 
\end{algorithm*}

We present the pseudocode of the tree search MPC in Algorithm \ref{alg:bfsearch}. Additionally, we provide an example showing how the best-first heuristic works in an assumed decision time with $\gamma=1$ and $|\scriptA{}| = 3$ and maximum planning steps $3$.

\begin{figure*}[htbp]
\centering
\captionsetup{justification = centering}
\includegraphics[width=0.99\textwidth]{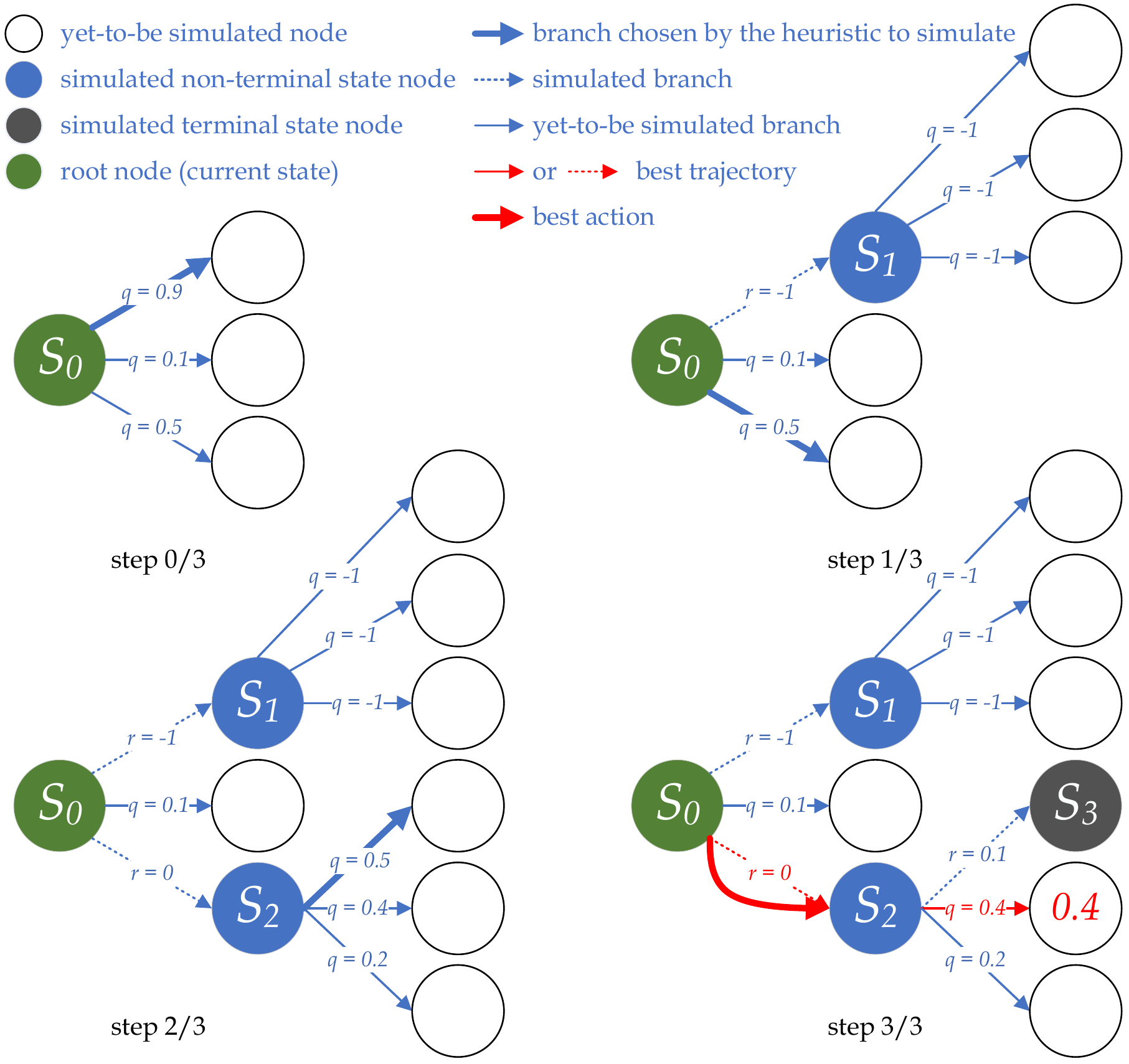}

\caption{\small Example of the Best-First Heuristic: Step 0 / 3) Start of planning, with the root node and three branches. The branch $\langle s_0, a_0 \rangle$ is chosen due to the best-first heuristic. If we employ the random search heuristic, like what we do in OOD evaluation, a random branch would be chosen; Step 1 / 3) We expand the chosen branch, popped out of the priority queue. A new node is constructed, together with its out-reaching branches, which are added to the queue. Now the queue has $5$ branches in it. The heuristic marks $\langle s_0, a_2 \rangle$ to be the next simulated branch; Step 2 / 3) Simulation of $\langle s_0, a_2 \rangle$ is finished and $\langle s_2, a_0 \rangle$ is marked; Step 3 / 3) Node S3 is imagined via $\langle s_2, a_0 \rangle$ but it is estimated to be a terminal state. Now, the tree search budget is depleted. We locate the root node branch $\langle s_0, a_2 \rangle$ which leads to the trajectory with the most promising return $0.4$.}
\label{fig:bfsearch}
\end{figure*}

\section{Failed Experiments}
We list here some of our failed trials along our way of exploring the topic of this work.
\subsection{Straight-Through Hard Subset Selection with Gumbel}
We initially tried to use Gumbel subset selection \cite{xie2019differentiable} to implement a hard selection based bottleneck but to no avail. We expect the model to pick the right objects by generating a binary mask and then use the masked objects as the bottleneck set. This two-staged design would align more with the consciousness theories and would yield clearer interpretability. However, it suffers from an implicit chicken-and-egg problem that we have not successfully addressed: to learn how to pick, the model should first understand the dynamics. Yet if the model does not pick the right objects frequently enough, the dynamics would never be understood. Our proposed semihard / soft approaches address such problem by essentially making the two staged selection and simulation as a whole for the gradient-based optimization.

\section{More Discussions on Limitations \& Future Directions}
This paper serves as a proof-of-concept of an interesting research direction: System-2 DRL. It is healthy to point out the limitations of this work as well as some interesting future research directions:
\begin{itemize}[leftmargin=*]
\item
This paper does not solve the ``planning horizon dilemma'', a fundamental issue of error accumulation of tree search expansion using imperfect models \cite{janner2019trust}. We strongly believe that incorporating temporal abstraction of actions, \eg{} options or subjective time models \cite{zakharov2020episodic} would gracefully address such problem. Promising as this is, introducing temporal abstraction to model-based RL is non-trivial and requires considerate investigation.
\item
Constant replanning may be prohibitive in reaction-demanding environments, especially when equipped with a computationally expensive set-based transition model. A planning strategy could be devised to control when or where for the agent to carry out planning, through the means of capturing uncertainty.
\item
The CP model cannot yet learn stochastic dynamics. The difficulty lies in the design of a compatible end-to-end trainable set-to-set machinery. We would like to address this in future.
\end{itemize}

\section{Potential Negative Societal Impacts}
We do not anticipate potential negative societal impacts since this paper is fundamental research regarding reinforcement learning methodology.

\end{appendices}

\end{document}